\documentclass[preprint,review,10pt,authoryear]{elsarticle}
\usepackage[cmex10]{amsmath}
\usepackage{algorithm}
\usepackage{algpseudocode}
\usepackage{algorithmicx}
\usepackage{url}
\usepackage{footnote}
\usepackage{epstopdf}
\usepackage{epsfig}
\usepackage{changes}
\usepackage{graphicx}
\usepackage{multirow}
\usepackage{array}
\usepackage{amsfonts,epsfig,array,graphicx,amssymb,epstopdf}
\usepackage{caption}
\usepackage{subcaption}
\usepackage[]{natbib}
\usepackage{cite}

\begin{document}

\begin{frontmatter}

\title{Estimation of Linear Motion in Dense Crowd Videos using Langevin Model}

%% Group authors per affiliation:  \fnref{myfootnote}
\author{$^*$Shreetam Behera$^a$,
         Debi Prosad Dogra$^a$, Malay Kumar Bandyopadhyay$^b$  and Partha Pratim Roy$^c$} 
\address{School of Electrical Science,$^a$\\Indian Institute of Technology Bhubaneswar, Bhubaneswar-752050, India$^a$\\School of Basic Sciences,$^b$\\Indian Institute of Technology Bhubaneswar, Bhubaneswar-752050, India$^b$ \\Department of Computer Science and Engineering$^c$,\\Indian Institute of Technology Roorkee, Roorkee-247667, India$^c$\\
Email:\ sb46@iitbbs.ac.in$^a$, dpdogra@iitbbs.ac.in$^a$, malay@iitbbs.ac.in$^b$, proy.fcs@iitr.ac.in$^c$}

\cortext[mycorrespondingauthor]{Corresponding author. Tel.: +91 9861350475} 
\ead{sb46@iitbbs.ac.in}

\begin{abstract}
Crowd gatherings at social and cultural events are increasing in leaps and bounds with the increase in population. Surveillance through computer vision and expert decision making systems can help to understand the crowd phenomena at large gatherings. Understanding crowd phenomena can be helpful in early identification of unwanted incidents and their prevention. Motion flow is one of the important crowd phenomena that can be instrumental in describing the crowd behavior. Flows can be useful in understanding instabilities in the crowd. However, extracting motion flows is a challenging task due to randomness in crowd movement and limitations of the sensing device. Moreover, low-level features such as optical flow can be misleading if the randomness is high. In this paper, we propose a new model based on Langevin equation to analyze the linear dominant flows in videos of densely crowded scenarios. We assume a force model with three components, namely external force, confinement/drift force, and disturbance force. These forces are found to be sufficient to describe the linear or near-linear motion in dense crowd videos. The method is significantly faster as compared to existing popular crowd segmentation methods. The evaluation of the proposed model has been carried out on publicly available datasets as well as using our dataset. It has been observed that the proposed method is able to estimate and segment the linear flows in the dense crowd with better accuracy as compared to state-of-the-art techniques with substantial decrease in the computational overhead.

\end{abstract}

\begin{keyword}
\texttt{Crowd Flow Segmentation, Crowd Dynamics, Visual Surveillance, Langevin Equation}

\end{keyword}

\end{frontmatter}

%\linenumbers
\section{Introduction}
\label{sec:introduction}
Recent advancement in computer vision-based crowd surveillance has drawn interests of the researchers and law enforcing agencies across the world. Automatic visual surveillance through expert decision making systems often results in efficient crowd monitoring and management with higher accuracy and better information fusion. Moreover, such intelligent systems can reduce human efforts leading to less errors in estimation. Expert systems guided automatic visual surveillance frameworks can promptly indicate unusual behavior or activity in crowd. Therefore, precautionary measures can be taken in order to avoid undesirable incidents. Such systems can also be used to understand human behavior of the people in crowded situations. However, majority of existing systems find it hard to handle dense crowds such as religious festivals, social and political gatherings because of the complexity of the problem in terms of functionality and time \citep*{junior2010crowd}. Computer vision research community tend to adopt machine vision-based algorithms in the above situations \citep*{junior2010crowd, yogameena2017computer}.
\subsection{Related Work}
Flow detection and segmentation are key to develop automatic crowd monitoring systems. Existing research work on crowd flow segmentation are either physics and particle dynamics-based or standard computer vision guided techniques as mentioned in~{\citep*{zhang2018physics}. 
\subsubsection{Physics or Particle Dynamics-based Methods}
In case of physics-based or particle dynamics-based methods,  typical fluid-flow model or freely moving particles on air may not be directly applied on crowd. For example, Langevin theory of Brownian motion may not be directly applied on crowd dynamics. However, such physics-based models can be adopted with context imposed restrictions. For example, correlating the actual forces in dense crowded situations with particle dynamics can be interesting. Such models have started to emerge off late~\citep*{zhang2018physics}. Ali et al.~\citep*{ali2007lagrangian}~have used Lagrange particle dynamics to segment high density crowd flows. The same framework has also been used for detecting instabilities in crowd flows. Zhang et al. in~\citep*{zhong2008crowd} have used Markov Random Field to define crowd energy upon which wavelet analysis has been performed to detect abnormal behaviors. However, the method is not independent of background and it is sensitive to video shaking. The authors in~\citep*{ali2008floor} have used a scene-structure-based force model to detect individuals in high-density crowd by analyzing its static, dynamic, and boundary floor fields. The algorithm is highly computation intensive. Mehran et al.~\citep*{mehran2010streakline} have used streaklines for crowd flow analysis. They have used social force graph technique and streaklines to analyze the flow. The authors of \citep*{ji2017anomaly} have proposed a method based on social force model to detect crowd anomaly at pixel and block levels. In~\citep*{wu2017bilinear}, the authors perform analysis of the crowd based on a bilinear interaction of curl and divergence of the flows. In~\citep*{ullah2017density}, a density independent hydrodynamics model (DIHM) for coherency detection in crowded scenes, has been proposed. The method has the capability to handle changing density over time. The method doesn't perform finer-level crowd flow segmentation. A spatio-temporal driving force model has been proposed in~\citep*{li2010group} to perform group segmentation in crowded scenarios. However, the model is not view variant and it needs to learn for different views with different parameter settings. In \citep*{chen2011adaptive}, the authors have presented an adaptive human motion analysis and prediction
method for understanding the motion patterns in crowds. Solmaz et al. in~\citep*{solmaz2012identifying} have proposed a method to identify multiple crowd behaviors through stability analysis for dynamical systems avoiding object detection, tracking, or training. Their method cannot capture the randomness in a crowd. In~\citep*{lin2016diffusion}, coherent regions in a crowded scene is detected based on thermal diffusion process and time-series clustering. The coherency is lost as the method merges both the motion and non-motion regions together over time. The agent-based method proposed in \citep*{kountouriotis2014agent} can model crowd behavior based on group dynamics and agent-based personality traits. Though their method performs reasonably well in real-time scenario, but its performance ceases with increase in the number of agents. In \citep*{zhou2015learning}, Zhou et al. have proposed a new mixture model of dynamic pedestrian-Agents (MDA) to learn the collective behavioral patterns of pedestrians in crowded scenes. However, it is unclear that how their method can handle varying density. In \citep*{su2013large}, the authors have proposed a spatio-temporal viscous fluid field to recognize the large-scale crowd behavior from appearance and driven factor perspectives. An application of real-time monitoring of crowd density at Puri Rath Yatra, combined with modeling evacuation scenarios using agent-based simulation has been proposed in \citep*{basak2017developing}. The technique is useful in predicting scenarios in emergency situations even though the technique is computationally expensive.

\subsubsection{Computer Vision-based Methods}
\par Conventional computer vision-based methods like optical flow-based methods have been instrumental in flow segmentation. The method proposed in\\~\citep*{cheriyadat2008detecting} finds dominant motions of crowd by clustering low-level feature point tracks in videos. Wu et al.~\citep*{wu2009crowd} have presented a region growing segmentation scheme based on the translational domain for segmenting crowd flows. The method fails if the translational flow is not local. Santoro et al.~\citep*{santoro2010crowd} have used Lucas-Kanade Tracker along with the density-based clustering for analysis of crowd motion. In the last step, a crowd tracker has been applied in each frame of the video. The authors claims their method can detect and track crowd with various shapes. However, the calculations are based on 2D coordinates of the motion point. Thus, the distance calculation between the motion points is not accurate. Moreover, the time complexity increases with the increase in motion points. In~\citep*{wu2009shape}, the authors have proposed a new framework for crowd movement analysis. The crowd flow segmentation is performed using optical flow field. An interpolation method based on Delaunay Triangulation has been used to estimate the smooth optical flow field in a robust way. Motion regions are then clustered and a shape derivative technique is combined with a region growing scheme in order to segment a crowd. However, the method cannot detect all motion regions in a typical crowd. The authors in \citep*{lu2017trajectory} have proposed a trajectory clustering-based method to understand crowd motion patterns. The method in \citep*{nasir2014prediction} aims to generate accurate sequence waypoints for the pedestrian walking path by analyzing videos in closed environments only. The authors in \citep*{anwar2012mining} have developed an anomaly detection method to analyze anomalous events. However, the proposed method works at microscopic level. An Interval-Based Spatio-Temporal Model (IBSTM) have been proposed in \citep*{kardas2017svas} in order to detect untoward events in a video. However, the proposed method is a microscopic event model that cannot deal with macroscopic events such as crowd flow. The method proposed in \citep*{walia2017novel} uses a multi-stage tracker for precise localization of targets. However, this model is a microscopic model aimed at individual humans only. The authors in \citep*{fernandez2012human} have proposed a finite state machines-based technique for human activity monitoring in a closed scene. The method needs to have prior information about source and destination points. The method described in \citep*{zhou2014CC} segments the motion flow in sparse crowds in terms of collectiveness. Fradi et al.~\citep*{fradi2017crowd} have proposed local descriptors which provide semantic information and interactive sparse crowd behaviors. However, it is not clear how the method handles dense crowd.

\par Traditional computer vision algorithms amalgamated with machine learning techniques have also been used for performing crowd flow segmentation in videos. Cao et al.~\citep*{cao2015large} have performed large scale crowd analysis using Convolutional Neural Networks (CNN). The authors have combined CNN guided classification with regression to get accurate results. However, a large database with proper labeling must be available  for such a method to be successful. The authors in \citep*{zhou2016spatial} have proposed a spatio-temporal CNN for crowd anomaly detection. In~\citep*{kruthiventi2015crowd}, crowd flow analysis is performed using Conditional Random Field. However, the method is incapable to handle intersecting flows. Deep learning-based optical flow schemes are also proposed in \citep*{dosovitskiy2015flownet}~and~\citep*{ilg2017flownet} to predict optical flow of consecutive frames based on Convolution Neural Networks. But, they do not address the dynamics of typical crowded scenarios. The methods proposed in \citep*{shao2015deep} and \citep*{long2015fully} are based on deep learning techniques for scene understanding and semantic segmentation. However, these methods are unable to describe the dynamics of the crowd. The authors of~\citep*{chaker2017social} have proposed an unsupervised approach for crowd scene anomaly detection based on the social network model. 
In the paper \citep*{wu2018collective}, collective density clustering is performed for detection of coherent crowd regions. However, the method is dependent on stability and accuracy of the tracking algorithm. 
In \citep*{direkoglu2017abnormal},  the angle difference between optical flow vectors,  has been used as a feature, fed to Support Vector Machine (SVM) for detecting abnormality in crowd. The authors in \citep*{yuan2016congested} have proposed a sparse representation method for crowd anomaly detection. The work presented in~\citep*{chan2008modeling} 
and~\citep*{ma2009motion} are based on dynamic mixture model of textures and expected-maximization (EM) algorithm. Such methods can segment motion in traffic and crowd videos. However, the authors have not provided any evidence on how it addresses crowd in terms of varying density.

\par From the aforementioned work, we have made the following observations:
\begin{itemize}
\item The methods similar to \citep*{santoro2010crowd} are restricted to be applicable for low and medium density crowd. These methods lack robustness in handling densely crowded scenarios.
\item  Though the physics-based \citep*{chaker2017social} and particle-dynamics-based \citep*{ali2007lagrangian} models partially address the issues in densely crowded scenarios, they are complex in functionality and often leads to increased execution time. Moreover, such methods lack simplicity from the point of implementation.
\item  It has also been understood that, none of the existing methods such as \citep*{ali2007lagrangian} or~\citep*{ullah2017density} address the movement as random particles in the fluid. This has been one of the key motivations behind the idea presented in this paper. 
\end{itemize}

\subsection{Contributions}
Following research contributions have been made to mitigate the aforementioned limitations:
\begin{itemize}
\item We propose a fast computational model to understand the dense crowd flow in videos using Langevin theory of Brownian particles in fluid. 

\item Using the aforementioned model, we propose an algorithm that can segment linear and near-linear flows of dense crowd movements in videos with the help of a context adaptive force model.

\end{itemize}

\par The rest of the paper is organized as follows. The foundation of Langevin equation is explained in Section~\ref{sections:BF}. In Section~\ref{sections:proposed}, we explain how Langevin equation can be adopted for designing expert decision making system to understand crowd flow through segmentation. The results are presented in Section~\ref{sections:results} using public datasets as well as using our video dataset. In Section~\ref{sections:Conclusions}, we have concluded the paper with possible future directions.

\section{Background and Foundation}
\label{sections:BF}
Langevin equation is perhaps the simplest way to describe the dynamics of non-equilibrium systems. It is a stochastic differential equation introduced first to describe the motion of a particle in fluid as mentioned in~\citep*{langevin1908theorie,coffey2004langevin}. Since the motion of a particle is random, it cannot be described only using Newton's force. In order to estimate the random and fluctuating motions of the particle, the basic Newtonian force is added with two additional force components: frictional force and random force.
%%---------------------------------------------------------------------
	\begin{figure}[h]      %{0.40\textwidth}
	\centering	
    \includegraphics[scale=0.4,width=0.46\textwidth]{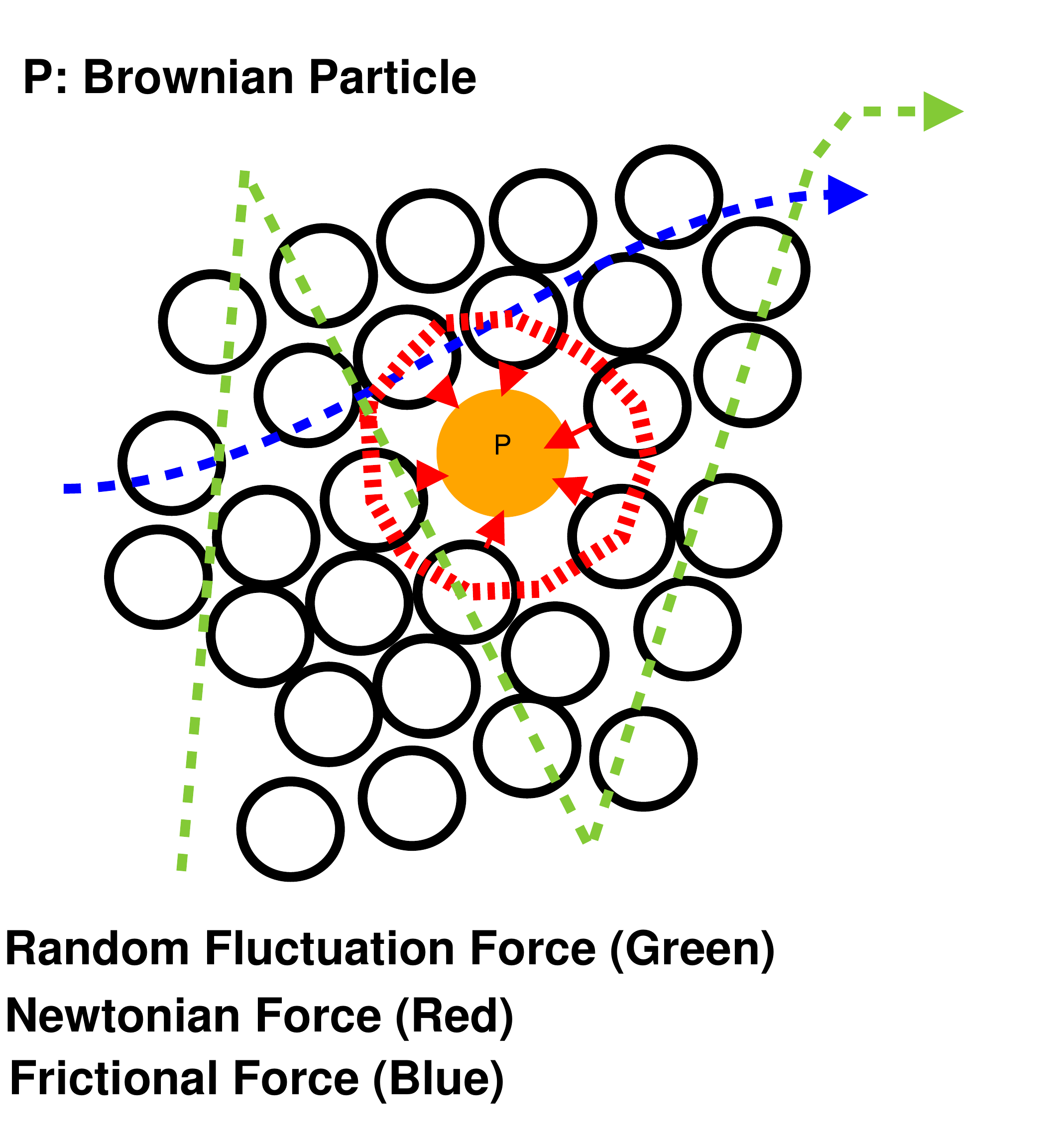}
    \caption{Pictorial representation of interaction of a Brownian particle with different forces in a fluid.}
    % Here the Newtonian force is depicted in dotted red lines which is due to the particles in the surrounding particles}. The blue dotted line represents frictional force ($\gammav(t)$) and the green dotted lines represent the random fluctuation force caused due to random density fluctuations in the fluid($\xi(t)$).}
    \label{fig:LE}
%\end{figure}
\end{figure}
%\hspace{10pt}
	\begin{figure}[h]            %{0.40\textwidth}
%\begin{figure}[H]
	\centering	
	%\hspace{-1 cm}
 \includegraphics[scale=0.58,width=0.58\textwidth]{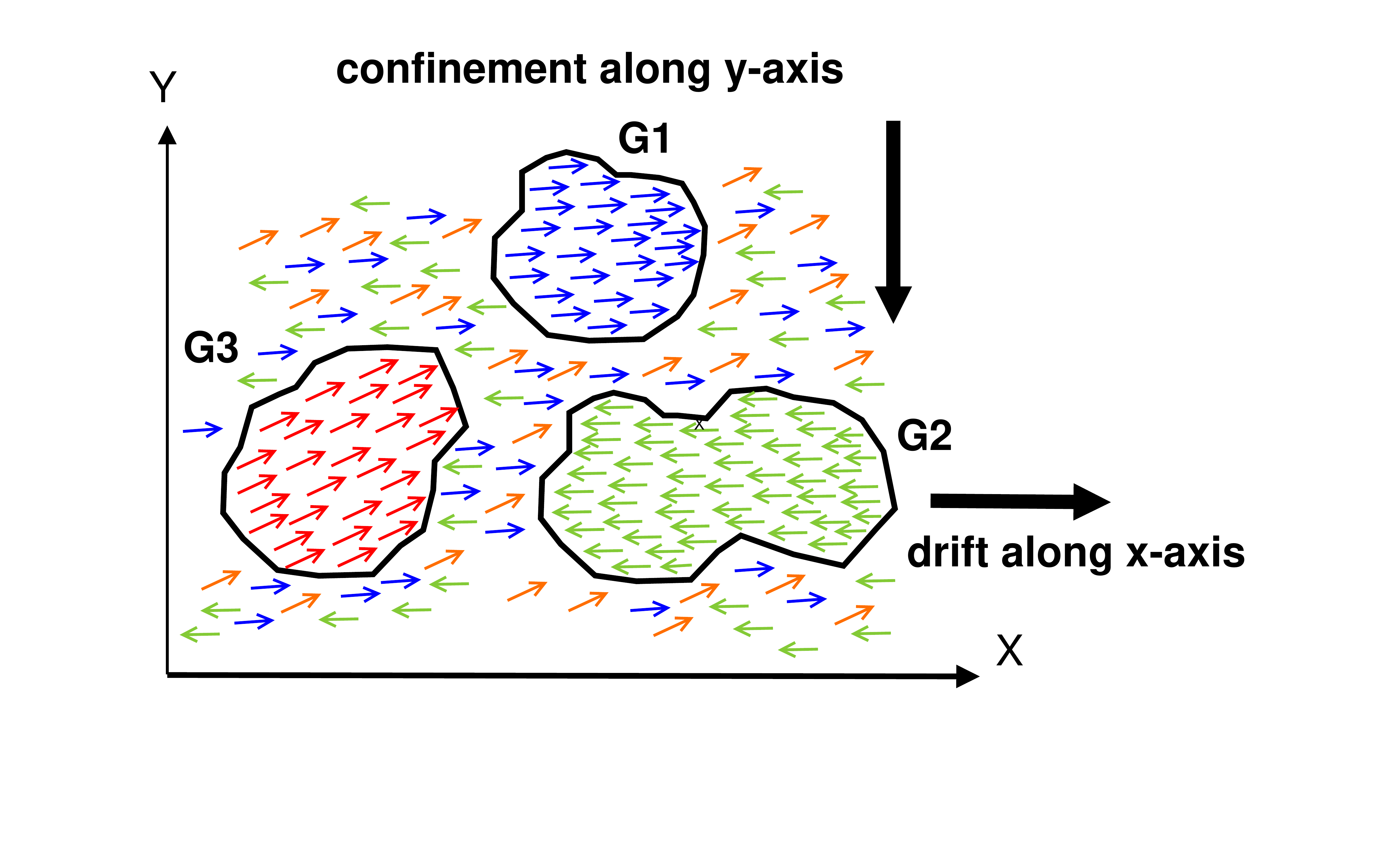}        
    \caption{Representation of particles in motion in 2D space, grouped with similar orientations, experiencing drift and confinement forces along $x$ and $y$-axes, respectively.}
    \label{fig:CG}
 \end{figure}
 
%%--------------------------------------------------------------------

Consider the \texttt{1D} motion of a particle of mass $m$ as shown in Fig.\ref{fig:LE}. According to Newton's second law of motion, the motion of the particle is described using (\ref{Eqn:ForceV}),

\begin{equation}
\label{Eqn:ForceV}
\frac{\text{d}v(t)}{\text{d}t}=F(t)
\end{equation} 
where $m$ is the mass of the particle, $v(t)$ is the velocity of the particle at time $t$ and $F(t)$ is the instantaneous force exerted on the particle at time $t$.

The instantaneous force $F(t)$ as represented in (\ref{Eqn:Langevin1}), acting on a particle, is originated from the impact received from the surrounding fluid molecules. Langevin suggested that the force $F(t)$ can be written as a sum of two components. The first part is an $averaged-out$ component which basically represents the viscous drag, $-\gamma v(t)$, where $\gamma$ is the frictional coefficient. In general, this frictional force is assumed to be proportional to the velocity of the particle. The second component of this instantaneous force $F(t)$ is a rapidly fluctuating  part $\xi(t)$ which arises due to random density fluctuations in the fluid.
\begin{equation}
 m\frac{\text{d}v(t)}{\text{d}t}= -\gamma v(t) + \xi(t) 
\label{Eqn:Langevin1}
\end{equation} 

The random force $\xi(t)$ averages out to be zero over long intervals as mentioned in (\ref{Eqn:disspForce1}). The second moment is actually relating the fluctuating force with the viscous drag or dissipative force $g$, which is related to $\gamma$ as mentioned in (\ref{Eqn:disspForce2}),
\begin{equation}
\label{Eqn:disspForce1}
\langle\xi(t)\rangle_{\xi}=0,~ \langle\xi(t_{1})\xi(t_{2})\rangle_{\xi}=g\delta(t_{1}-t_{2})
\end{equation}
where $\langle\xi(t)\rangle_{\xi}$ represents an average value considered with respect to the distribution of the realizations of the variable $\xi(t)$,

\begin{equation}
\label{Eqn:disspForce2}
\texttt{g}=2 \xi K_{B} T
\end{equation}
and $g$ is the measure of strength of the fluctuation force, $K_{B}$ is Boltzmann's constant, $T$ is the temperature, and $\delta$ is the delta function.

\section{Proposed Crowd Flow Segmentation Method}
\label{sections:proposed}
The proposed crowd segmentation method using Langevin equation is discussed here. For a given video sequence, over a window of size $W$, the keypoints are extracted and propagated to the proposed model as illustrated in Fig.\ref{fig:BD}. Inside a typical window, this partial flow information is passed on to the proposed model where flow segmentation is carried out over the remaining frames of the window. Windowing ensures re-initialization of the process at regular intervals, which tracks the flow changes in the frames in temporal domain.
\begin{figure}
%\vspace{-10pt}
	\centering	
    \includegraphics[scale=0.68,width=0.74\textwidth]{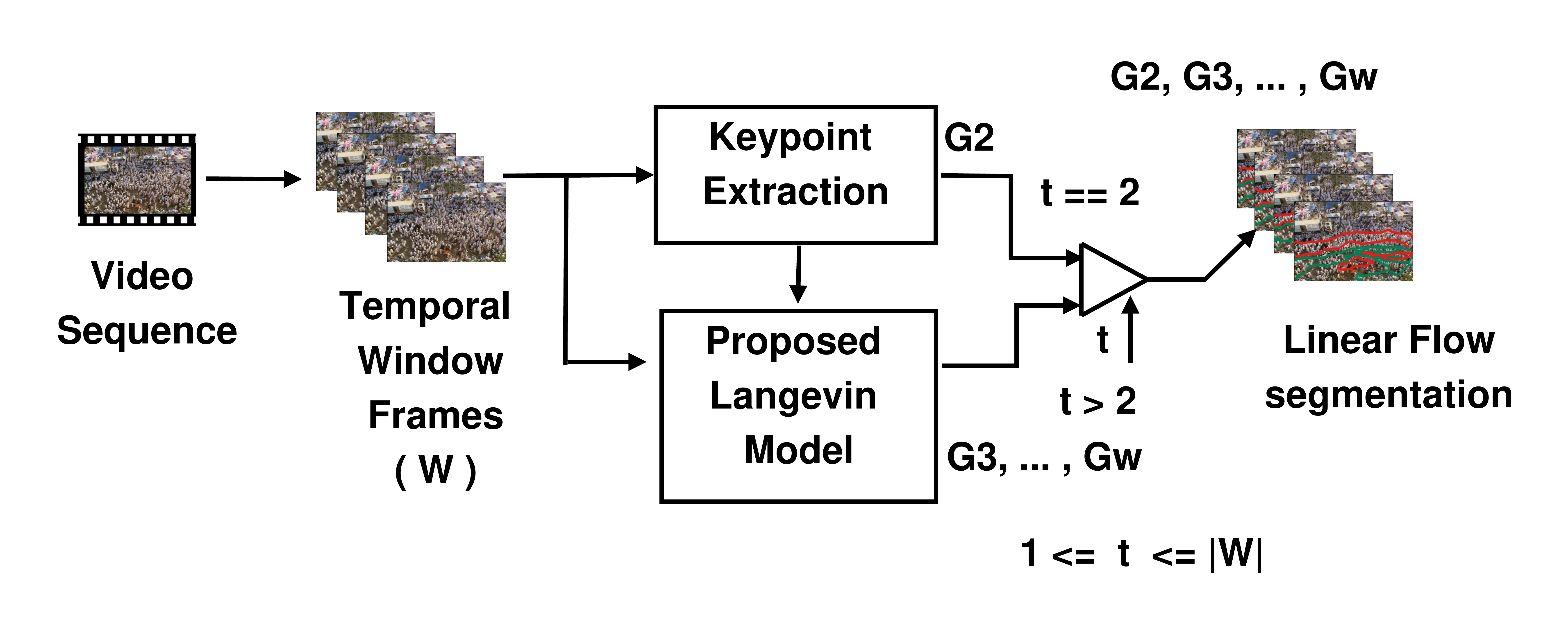}
    %{physicsFormulation6.pdf}
    \caption{Block diagram representation of the proposed crowd flow segmentation scheme. Over a temporal window $W$, the first two frames are used for keypoint extraction to generate segmented map $G_{2}$ consisting of grouped keypoints and the subsequent frames are used for Langevin guided flow segmentation to generate $|W|-1$ segmented maps.}  
    \label{fig:BD}
\end{figure}
\subsection{Keypoint Extraction}

\par Dense optical flow using Farneback's method as described in \citep{farneback2003two} has been estimated on first two down-scaled frames of $W$. The flow vectors obtained are used to compute magnitude and orientation maps using equations (\ref{Eqn:magnitude}) and (\ref{Eqn:orientation}), respectively. The orientation map is quantized into eight bins using a magnitude threshold within $[0,2\pi]$. The envelope joining all the bin peaks forms a quantization curve.
\begin{equation}
v = \sqrt{|v_{x}|^2+|v_{y}|^2}
\label{Eqn:magnitude}
\end{equation}
\begin{equation}
\theta=\arctan {(|v_y|/|v_x|)}
\label{Eqn:orientation}
\end{equation}
Using a standard peak detection algorithm upon this quantization curve, the peaks are detected. Keypoints corresponding to these peaks are retained and others are discarded. In the next step, grouping of retained keypoints is performed. Mainly, two factors are considered for grouping: orientation and spatial connectivity. The keypoints surrounding the considered keypoint are in a group if their quantized orientations are equal and they lie within a $3$x$3$ neighborhood of the considered keypoint. The last condition accounts for the spatial connectivity of the considered keypoint with its neighboring keypoints.
 %The keypoints are obtained corresponding to the highest peaks and are grouped based on spatial connectivity and orientations. 
These grouped keypoints are assumed as initial segregated structured flows and are fed to Langevin-based model for the temporal flow segmentation for the remaining frames within the window. The entire process is illustrated in Fig.\ref{fig:KE}. 

\begin{figure}
%\vspace{-10pt}
	\centering	
    \includegraphics[scale=0.38,width=0.64\textwidth]{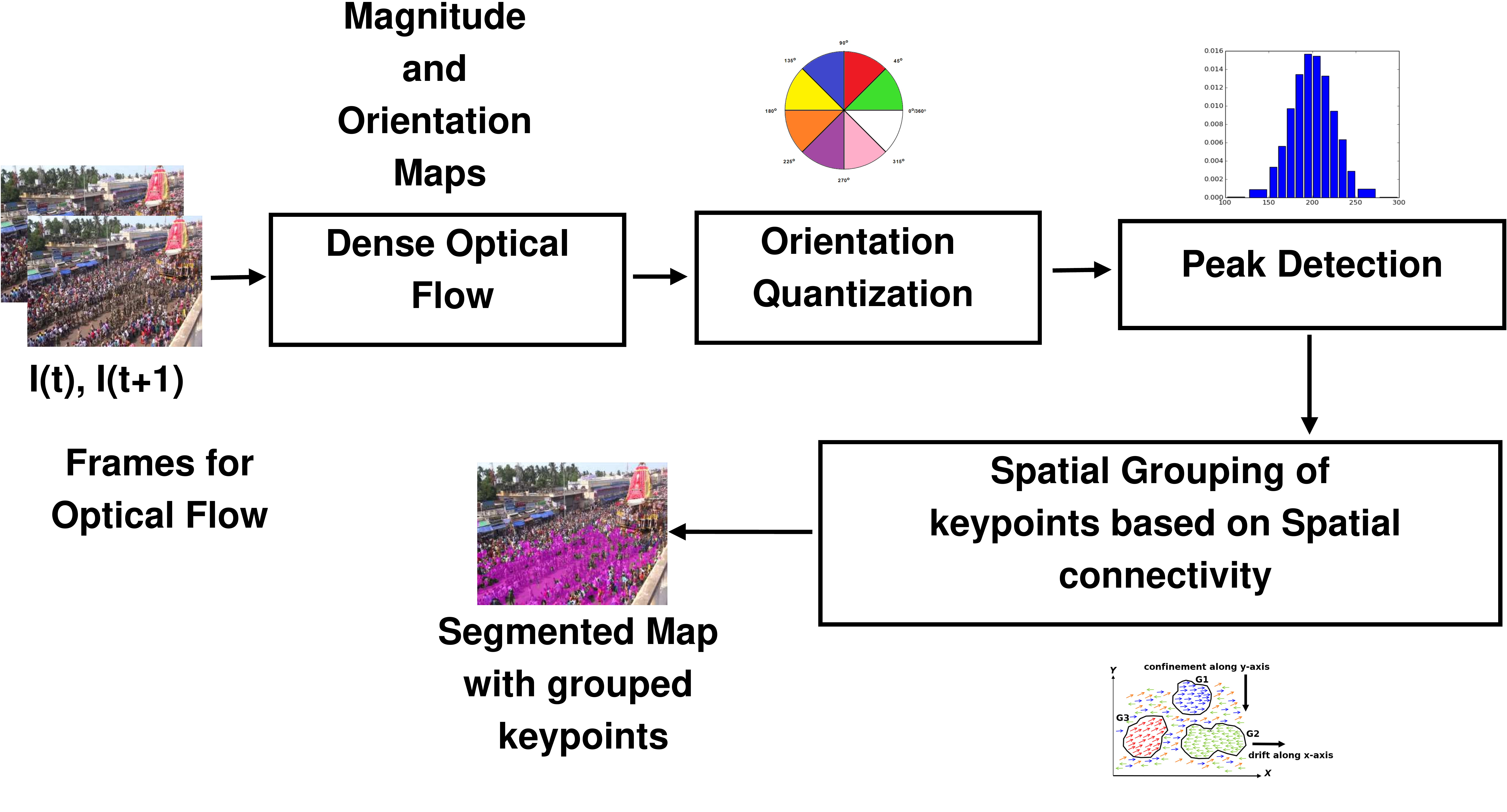}
    %{physicsFormulation6.pdf}
    \caption{Keypoint extraction performed over the first two frames of the window $W$.}
    \label{fig:KE}
\end{figure}
%\subsection{Langevin Model Guided for Flow Segmentation }
\subsection{Langevin Equation-guided Flow Segmentation}
\subsubsection{Formulation of Langevin Equation-based Force Model}

The dense crowd can be considered analogous to particles moving in the fluid. After careful visual observation of several real-live crowd movement videos, we have realized that, structured crowd usually move together in groups of similar orientations as shown in Fig.\ref{fig:CG}. In such cases, it is possible to approximate the motion with the help of Langevin theory. We have assumed that the force is acting on these groups instead of individual particle. 
Thus, the resultant force can be reconstructed as combination  of various forces acting upon and within the group as represented in (\ref{Eqn:ForceInertia}).
\begin{equation}
{F_{inertial}=~F_{external}+F_{drift/confine}+F_{disturbance}}
\label{Eqn:ForceInertia}
\end{equation}
The resultant force as mentioned in (\ref{Eqn:ForceInertia}) is constituted with external forces arising due to the motion of the  group in the surrounding ($F_{external}$), the drift forces that may cause the particle to drift along with the group in a particular direction or the confinement forces that confines the particles to stay within the group i.e. ($F_{drift/confine}$), and disturbances due to the noise in the group ($F_{disturbance}$). 

For example, if we assume the movement of particles of the group along $x$-axis as depicted in Fig.~\ref{fig:CG}, inertial force on a particle along $x$-axis can be represented as (\ref{Eqn:MassForceInertia}),
\begin{equation}
 m\frac{d^2x}{dt^2}\
  =-\gamma_{x}\frac{dx}{dt}+F+D_{x}\xi_{x}(t)
  \label{Eqn:MassForceInertia}
\end{equation}
where $m$ is the mass of the particle, $\gamma_{x}$ represents the resistive force due to particles and surrounding groups, $F$ is the constant drift force, $D_{x}$ is the strength of the noise, and $\xi_{x}$ represents the random force due to random density fluctuations in the considered group.
Simplifying (\ref{Eqn:MassForceInertia}), we obtain
\begin{equation}
 m\frac{dv_{x}}{dt} =-\gamma_{x}v_{x}+F+D_{x}\xi_{x}(t).
  \label{Eqn:VelForceInertiaX}
\end{equation}

The first term in the right hand side of (\ref{Eqn:VelForceInertiaX}) is the resistive or opposing force experienced by the group because of the surrounding particles, while the second term is the drifting force responsible for causing the motion of the particle in the group along $x$-axis. The third term is the force resulted due to the noise and internal disturbances within the group.
Similarly, for the force acting on the particles in the group along $y$-axis is represented as in (\ref{Eqn:VelForceInertiaY}),
\begin{equation}
 m\frac{dv_{y}}{dt}\
  =-\gamma_{y}v_{y}-\frac{\partial U(y)}{\partial y}+D_{y}\xi_{y}(t)
    \label{Eqn:VelForceInertiaY}
\end{equation}

where $m$ is the mass of the particle, $\gamma_{y}$ represents the resistive force due to motion of the surrounding group, $U$ is the confinement force, $\frac{\partial U(y)}{\partial y}$ is the rate of change of confinement force along $y$-axis along unit length, $D_{y}$ is the strength of the noise, and $\xi_{y}$ represents the represents the random force due to random density fluctuations within the considered group.\\
\subsubsection{Implementation of Langevin-based Force Model for Flow Segmentation}

\par The numerical solutions of (\ref{Eqn:VelForceInertiaX}) and (\ref{Eqn:VelForceInertiaY}) give the predicted velocities of the particles in motion along $x$-axis and $y$-axis as shown in~(\ref{Eqn:VelPredX})~and~(\ref{Eqn:VelPredY}), respectively. Further integrating ~(\ref{Eqn:VelPredX})~and~(\ref{Eqn:VelPredY}), we get the new or predicted positions of the particles.
\begin{equation}
v_{x, new}=v_{x, old}-\gamma_{x}v_{x,old}{\Delta}t+F{\Delta}t+D_{x}\xi_{x}{\Delta}t
\label{Eqn:VelPredX}
\end{equation}
\begin{equation}
v_{y, new}=v_{y, old}-\gamma_{y}v_{y,old}{\Delta}t-\frac{\partial U(y)}{\partial y}{\Delta}t+D_{y}\xi_{y}{\Delta}t
\label{Eqn:VelPredY}
\end{equation}

Equations~(\ref{Eqn:PosPredX}) and (\ref{Eqn:PosPredY}) represent the predicted position of the particle with respect to its intial position $(x_{old},y_{old})$,

%\begin{equation}
%%\begin{split}
%x_{new} = x_{old}-\int (\gamma_{x}v_{x,new}{\Delta}t)\text{d}t \\
% 		 +\int(F{\Delta}t)\text{d}t+\int(D_{x}\xi_{x}{\Delta}t)\text{d}t
%\label{Eqn:PosPredX}
%%\end{split}
%\end{equation}
%\begin{equation}
%%\begin{split}
%y_{new} = y_{old}-\int(\gamma_{y}v_{y,new}{\Delta}t)\text{d}t \\
%		-\int(\frac{\partial U(y)}{\partial y}{\Delta}t)\text{d}t 
%		+ (\int(D_{y}\xi_{y}{\Delta}t)\text{d}t)
%\label{Eqn:PosPredY}
%%\end{split}
%\end{equation}

\begin{equation}
%\begin{split}
x_{new} = x_{old}+v_{x,new}\text{d}t
\label{Eqn:PosPredX}
%\end{split}
\end{equation}
\begin{equation}
%\begin{split}
y_{new} = y_{old}+v_{y,new}\text{d}t
\label{Eqn:PosPredY}
%\end{split}
\end{equation}

where $\Delta$t is the increment in time. In the above equations, the mass of each particle is set to $1$ for consistency.

The segmentation map in the previous section, consists of groups with similar orientations. For each particle i.e. keypoint in the group, equations (\ref{Eqn:VelForceInertiaX} and \ref{Eqn:VelForceInertiaY}) and equations (\ref{Eqn:PosPredX} and \ref{Eqn:PosPredY}) are used to predict the velocity and position of the particle, repsectively. For these groups, drift is the force that controls the overall group movement along $x$-axis. It is basically a group force which is computed as the cumulative sum of acceleration of the particles along $x$-axis as mentioned in (\ref{Eqn:driftforce}).
\begin{equation}
F_{drift,g}=m\sum_{i,g}\frac{dv_{x_(i,g)}}{dt} 
\label{Eqn:driftforce}
\end{equation}

Similarly, the confinement force confines the group in the $y$- axis which can be estimated as cumulative sum of acceleration of the particles along $y$-axis as in (\ref{Eqn:confinementforce}),
\begin{equation}
U_{confinement,g}=m\sum_{i,g}\frac{dv_{y_(i,g)}}{dt}
\label{Eqn:confinementforce}
\end{equation}

where $v_{x(i,g)}$ and $v_{y(i,g)}$ represent the velocities for the $i^{th}$ keypoint in the $g^{th}$ group along x-axis and y-axis, respectively. As mentioned earlier, mass is set to $1$ for consistency.

\par The predicted velocities, $\overrightarrow{v}_{x,new}$, $\overrightarrow{v}_{y,new}$ are further used  to compute magnitude and orientation maps, which are used to estimate the flow in the remaining frames of the current window avoiding optical flow computation in every consecutive frames. Finally, temporal segmentation maps are obtained representing the dominant flows in the window. The process of flow segmentation is presented in Algorithm \ref{algo:method}.
%$1$.  

\begin{algorithm}
\scriptsize

\caption{Crowd flow segmentation using Langevin theory}
\label{algo:method}
\textbf{Input:} $F (f_{1}, f_{2}, f_{3}, ...~, f_{T})$ = Video sequence with $T$ number of frames, $|W|$ = Size of Window, $(\gamma_{x}, \gamma_{y}, \xi_{x}, \xi_{y}, D_{x}, D_{y})$ = Parameters of the Proposed Model, $m_{t}$ = Magnitude threshold, $b$ = Quantization bins. \\
\textbf{Output:} $G$ = Linear flow segmented groups, where $|S|$= $|W|-1$
\begin{algorithmic}[1]
\State Initialize $m$~=~$\frac{T}{|W|}$.  
\For {i = 1 to m}
\State $W_{i}={f_{p+1}, f_{p+2}, ... ,f_{p+|W|}}$, where $p = i*|W|$
\State Calculate ${v_{x}, v_{y}}$ using Farneback method$(f_{p+1}, f_{p+2})$.
\State Calculate $ M and~\theta$ using (\ref{Eqn:magnitude}) and (\ref{Eqn:orientation}).
\State Compute Q by quantizing $\theta$ into $b$ bins in the range of $0$-$2\pi$ over the magnitude threshold $m_{t}$.
\State Extract keypoints $K$ and group them into groups based on spatial connectivity and orientation, in order to form $G_{2}$ segmented map .
	\For {j = 3 to $|W|$}
		\State Using (\ref{Eqn:VelPredX}-\ref{Eqn:VelPredY}) and (\ref{Eqn:PosPredX}-\ref{Eqn:PosPredY}), estimate the new positions of the groups.
	\EndFor
\EndFor
\end{algorithmic}
\end{algorithm}

\section{Results and Discussions}
\label{sections:results}

In this section, we first discuss about the datasets that have been used for evaluation of the proposed method, followed by experiments based on forces and force-parameters of the Langevin-guided segmentation force model. The segmentation results and computational results are presented in Sections \ref{section:CP} and \ref{section:SR}, respectively.

\subsection{Datasets}
\label{section:DS}
Two video datasets have been used for testing the proposed flow segmentation method. One of them is publicly available dataset containing three different videos. The other one (our dataset) contains ten hours of video recording of Cart Festival (Sri Jagannath Ratha Yatra) at Puri (Odisha, India). The details are presented in Table \ref{table:Datasets}.
% Table
\begin{table}[h]
\tiny
\centering
\caption{Datasets used for evaluation of the proposed method}
%\scriptsize
\label{table:Datasets}
%\begin{tabular}{|l|c|c|c|}
\begin{tabular}{lccc}
\hline
\multicolumn{1}{l}{\#Dataset} & \multicolumn{1}{c}{Crowd Density} & \multicolumn{1}{c}{Types of Motion} & \multicolumn{1}{c}{\begin{tabular}[c]{@{}c@{}}Significant Crowd \\ behavior\end{tabular}} \\ \hline
Marathon-I & Sparse & \begin{tabular}[c]{@{}c@{}}Linear, unidirectional\\  crowd movements\end{tabular} & \begin{tabular}[c]{@{}c@{}}People running \\ in one direction\end{tabular} \\
Marathon-II & Dense & \begin{tabular}[c]{@{}c@{}}Linear, unidirectional\\  crowd movements\end{tabular} & \begin{tabular}[c]{@{}c@{}}People running\\ in one direction\end{tabular} \\
Fair & Semi-Dense & \begin{tabular}[c]{@{}c@{}}Linear, bilinear, mixing \\ crowd movements\end{tabular} & \begin{tabular}[c]{@{}c@{}}People moving in \\ two different directions\end{tabular} \\
Rath Yatra & Semi-Dense & \begin{tabular}[c]{@{}c@{}}Linear, mixing \\ crowd movements\end{tabular} & \begin{tabular}[c]{@{}c@{}}People pulling \\ the cart in one direction\end{tabular} \\ \hline
\end{tabular}%
\end{table}

\subsection{Estimation of the Parameters}
\label{section:PS}
The proposed Langevin theory-based model aims to describe the random movement of structured groups in dense crowds. The parameters of force equations are mentioned in (\ref{Eqn:VelForceInertiaX}) and (\ref{Eqn:VelForceInertiaY}). 

%-------------------------New Figures for rx,ry-----------------------
	\begin{figure}[h]            %{0.40\textwidth}
%\begin{figure}[H]
	\centering	
	%\hspace{-1 cm}
 \includegraphics[scale=0.58,width=0.68\textwidth]{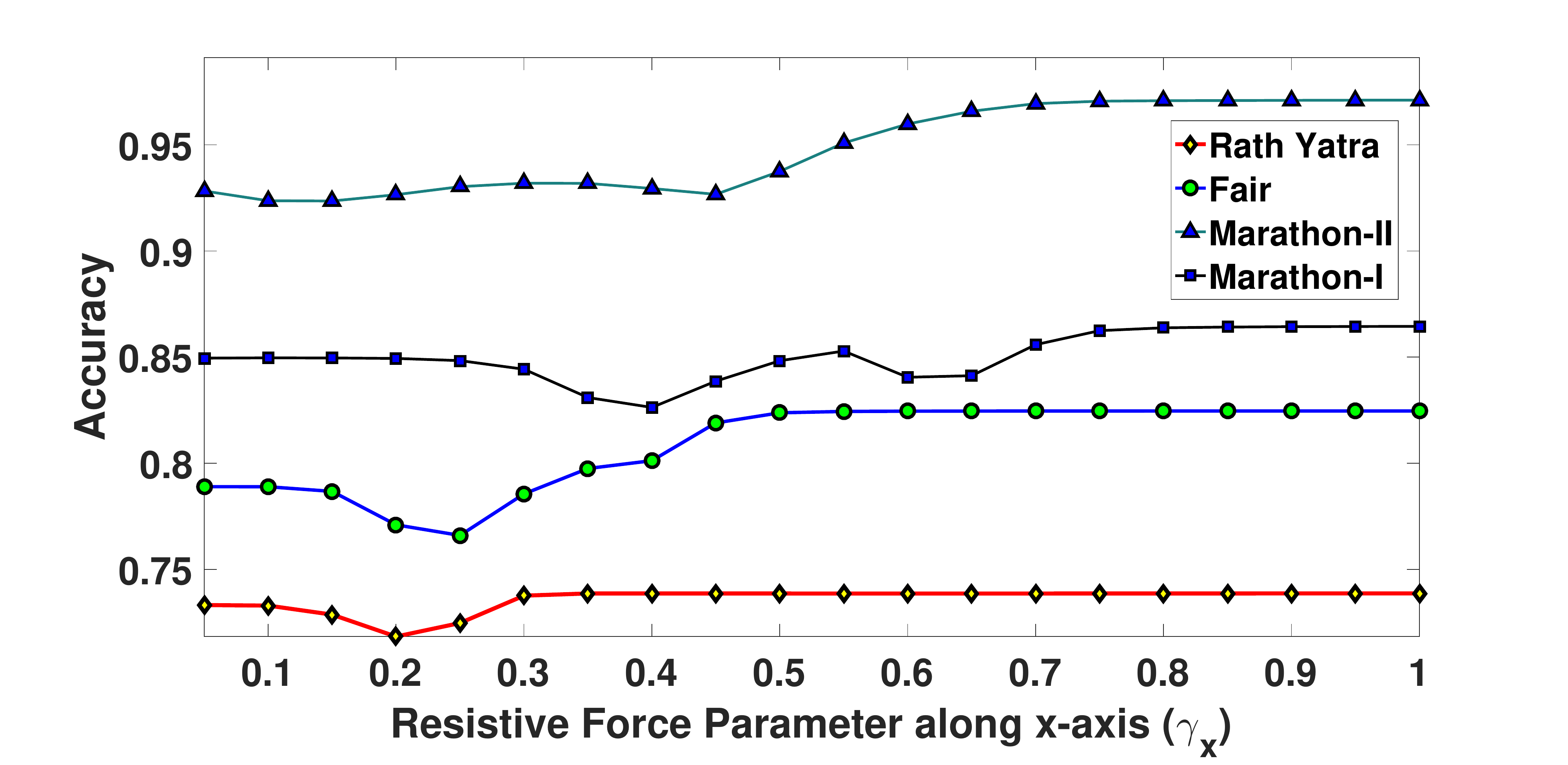}

    \caption{Graph showing how accuracy varies with respect to Resistive Force Parameter ($\gamma_{x}$) along $x$-axis.}
    \label{fig:rx}
 \end{figure}
 	\begin{figure}[h]            %{0.40\textwidth}
%\begin{figure}[H]
	\centering	
	%\hspace{-1 cm}
 \includegraphics[scale=0.58,width=0.68\textwidth]{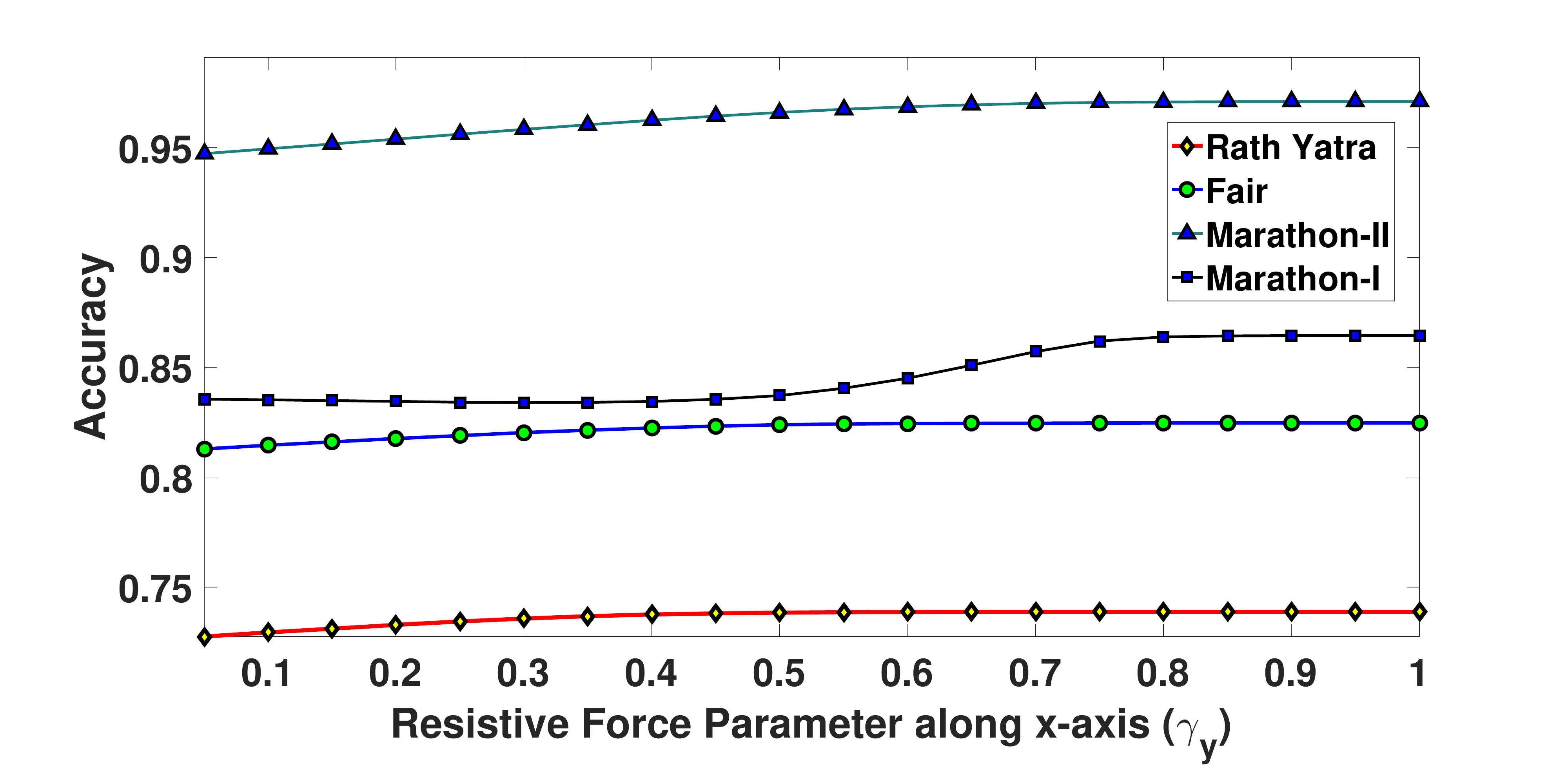}

    \caption{Graph showing how accuracy varies with respect to Resistive Force Parameter ($\gamma_{y}$) along $y$-axis.}
    \label{fig:ry}
 \end{figure}

%-------------------------------------------------------------------------

\par $\gamma_{x}$ and $\gamma_{y}$ are resistive force parameters, which are integral components of resistive forces acting upon the structured groups. Fig.\ref{fig:rx} and Fig.\ref{fig:ry} depict how the overall segmentation accuracy varies with $\gamma_{x}$ and $\gamma_{y}$. It has been observed from the graphs that the segmentation is not stable as the accuracy varies for initial values of $\gamma_{x}$ and $\gamma_{y}$. However, beyond certain values of $\gamma_{x}$ and $\gamma_{y}$, the accuracy does not change noticeably indicating a saturation in the segmentation process. It has been found that for Marathon-I and Marathon-II videos, when $\gamma_{x}$ is chosen to be $0.8$, the accuracy stabilizes. However, for Fair and Rath Yatra videos, accuracy stabilizes when the value of $\gamma_x$ is set to $0.6$ and $0.4$, respectively. Similarly, it has been observed that when $\gamma_{y}$ is outside the range $[0.6, 0.8]$, accuracy is consistent  across all videos. We therefore argue, more dense the crowd, more is the value of the resistive force. On the basis of above experiments, both $\gamma_{x}$ and $\gamma_{y}$ have been fixed to $0.8$.

\par The random fluctuating force consists of the parameters $\xi_{x}D_{x}$ and $\xi_{y}D_{y}$. These parameters are responsible for creating the disturbances within the group. 

In the graphs shown in Fig.\ref{fig:exdx} and Fig.\ref{fig:eydy}, it can be observed that when $\xi_{x}D_{x}$ remains within $[0.1, 0.7]$, segmentation output stabilizes. However, when its value is above $0.7$, segmentation accuracy drops. Similarly, when $\xi_{y}D_{y}$ remains within $[0.05, 0.6]$ range, accuracy does not change much. Beyond this, accuracy reduces sharply. Therefore, $\xi_{x}D_{x}$ and $\xi_{y}D_{y}$ have been fixed to $0.1$ and $0.5$, respectively.
%rx=0.8;  ry=0.8; ex=0.2; ey=0.72; Dx=0.05; Dy=0.8;
% DxEx
\begin{figure*}[htp]     %!h
   \centering
 %   \vspace{-20pt}
  %   \vspace{- 1 cm}
  \begin{subfigure}[b]{0.4\textwidth}
\includegraphics[width=5.0 cm,height=3.0 cm]{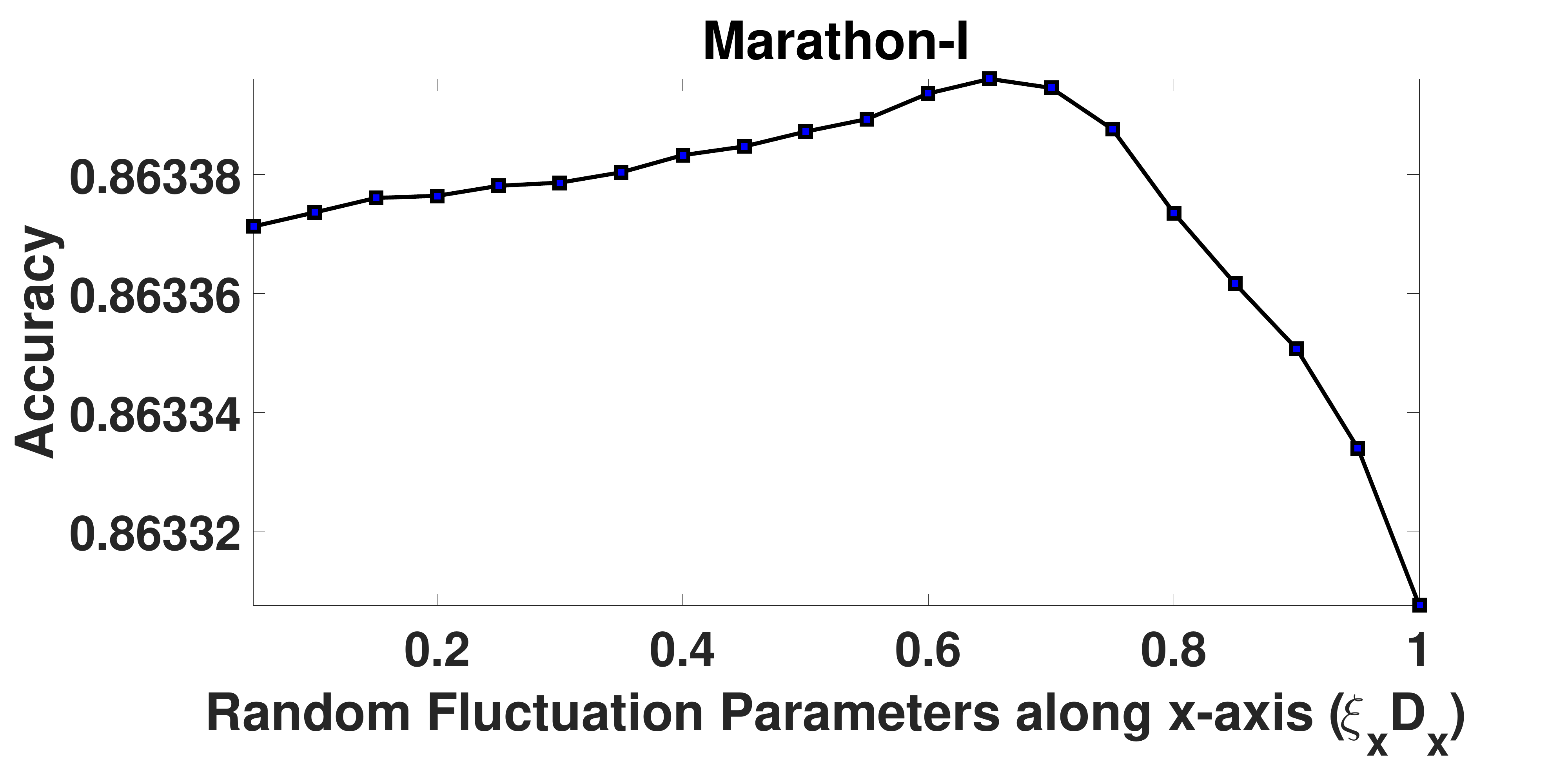}
        \caption{}
        \label{fig:m1_exdx}
    \end{subfigure}
    \hspace{0.5cm}
  %  \hspace{1.5 cm}
        \begin{subfigure}[b]{0.4\textwidth}
\includegraphics[width=4.8 cm,height=3.0 cm]{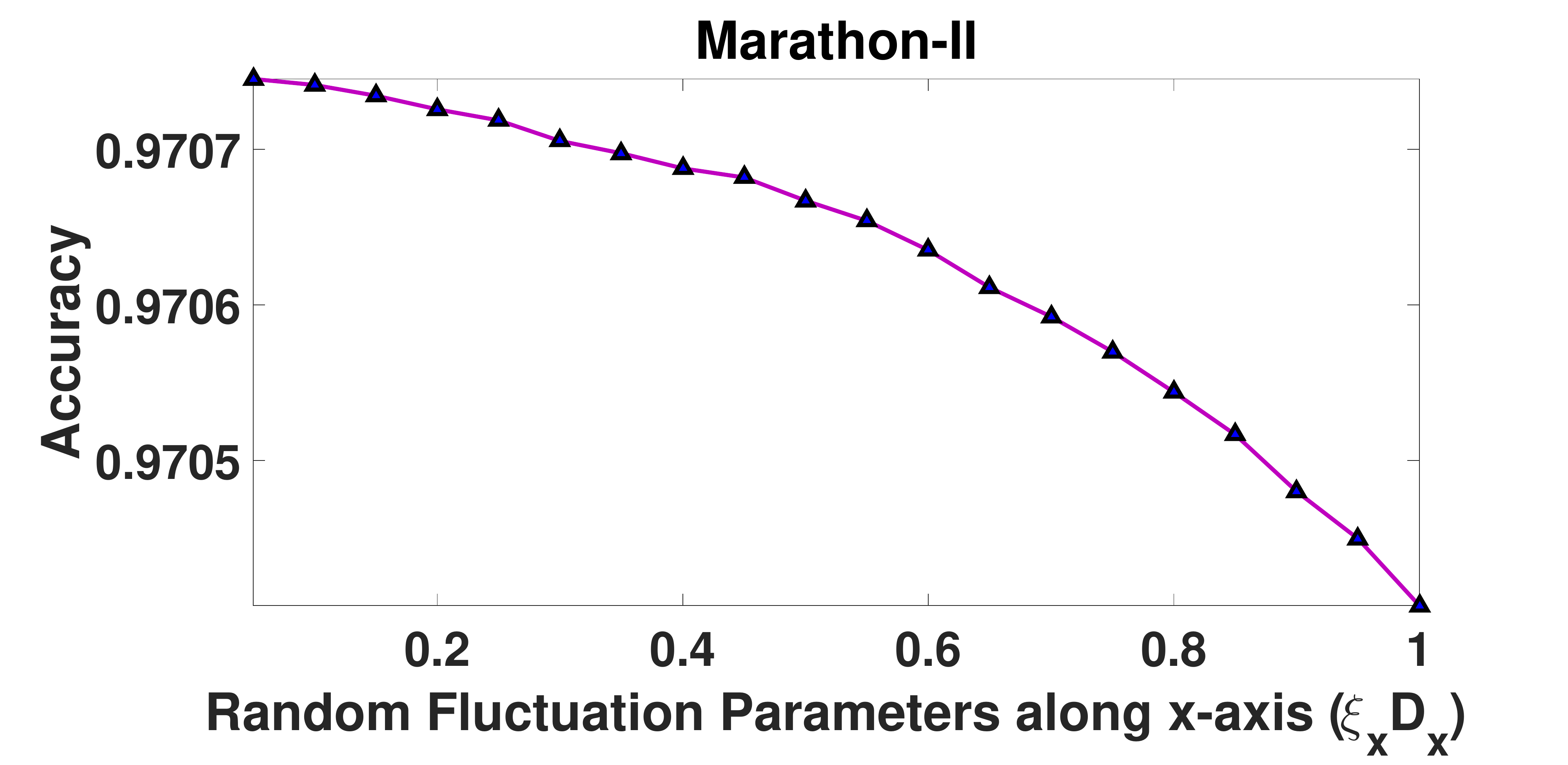}
        \caption{}
        \label{fig:m2_exdx}
    \end{subfigure}
    \hspace{0.4cm}
        \begin{subfigure}[b]{0.4\textwidth}
\includegraphics[width=4.8 cm,height=3.0 cm]{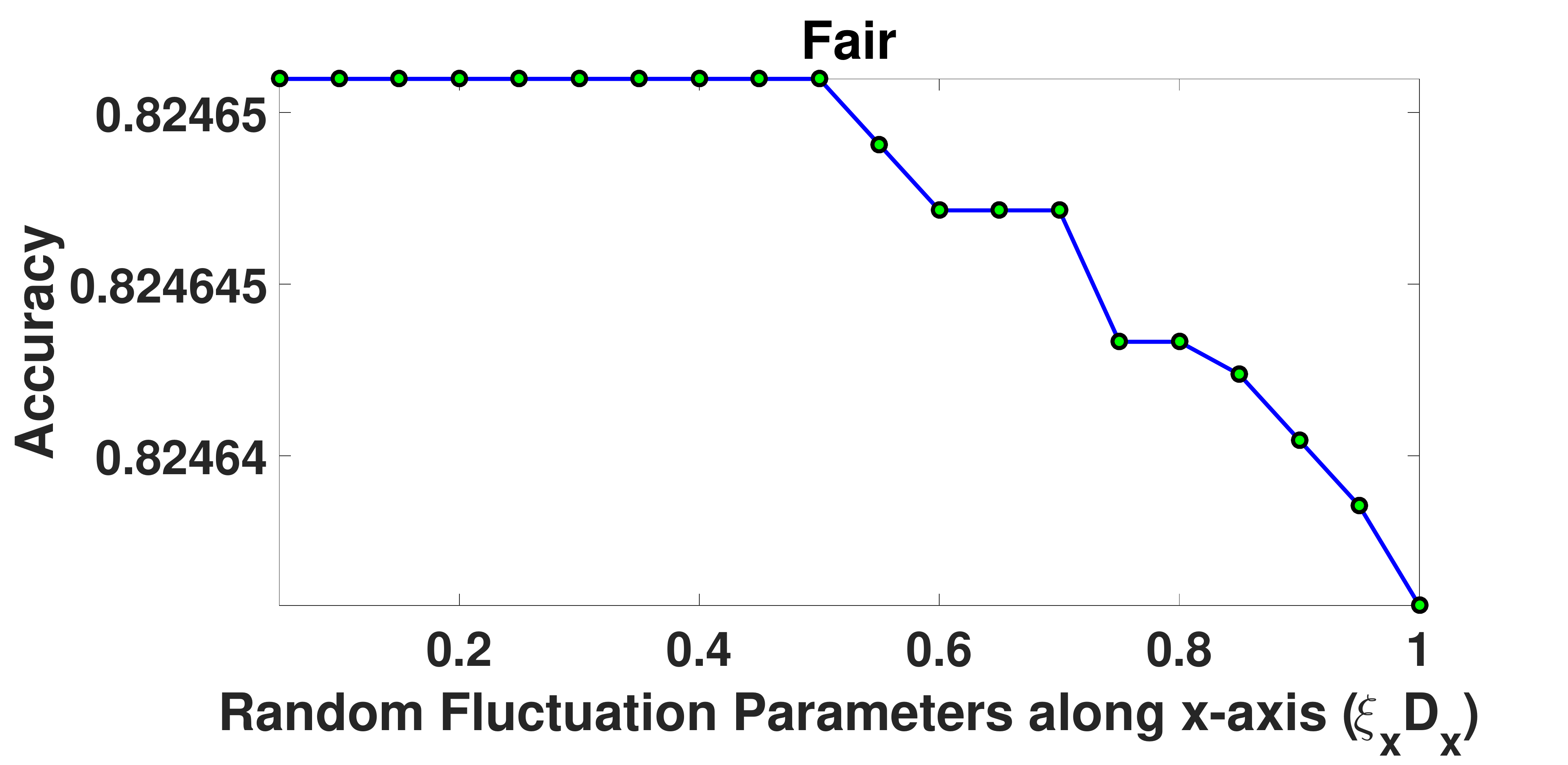}
        \caption{}
        \label{fig:fair_exdx}
    \end{subfigure}
 %   \hspace{1.5 cm}
 \hspace{0.4cm}
            \begin{subfigure}[b]{0.4\textwidth}
\includegraphics[width=4.8 cm,height=3.0 cm]{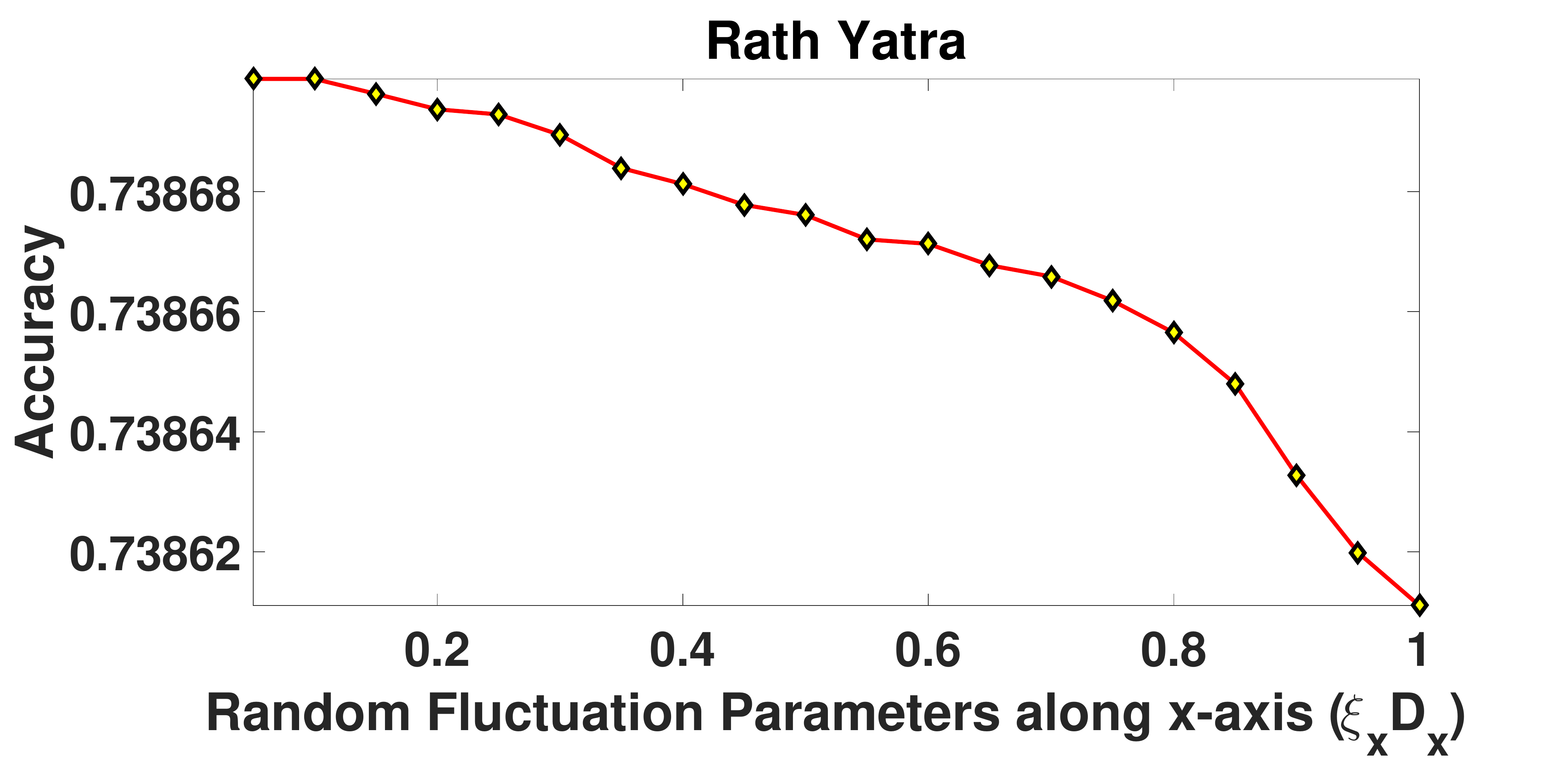}
        \caption{}
        \label{fig:RY_exdx}
    \end{subfigure}

    \caption{(a-d) Graphs showing how accuracy varies with respect to  random fluctuation force parameters along $x$-axis ($\xi_{x}D_{x}$) for different videos.}
    \label{fig:exdx}
    
\end{figure*}

% DyEy
\begin{figure*}[htp]     %!h
\centering
  %  \vspace{-1.0 cm}
  \begin{subfigure}[b]{0.4\textwidth}
\includegraphics[width=4.8 cm,height=3.0 cm]{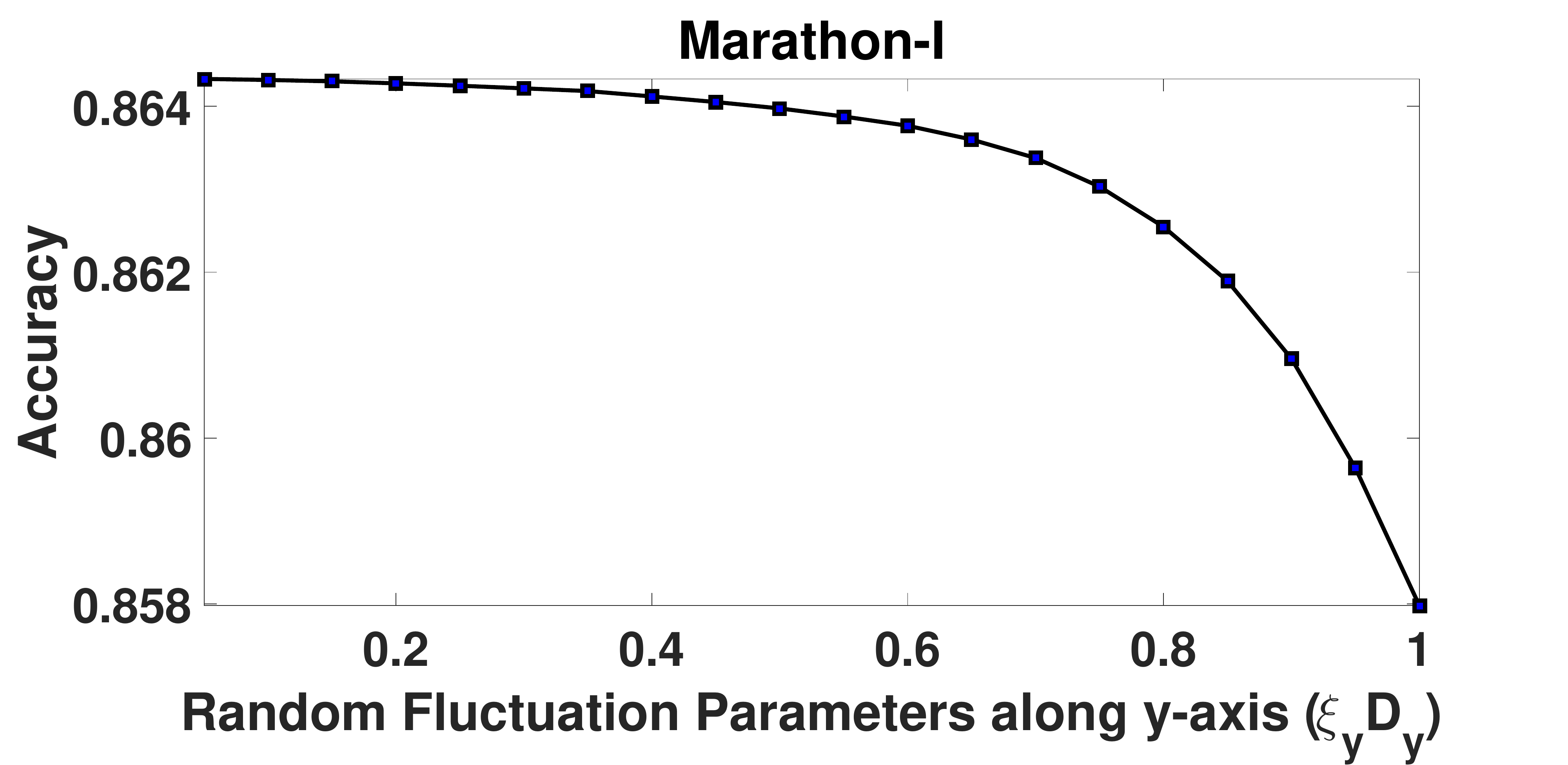}
        \caption{}
        \label{fig:m1_eydy}
    \end{subfigure}
   % \hspace{1.5 cm}
   \hspace{0.4cm}
        \begin{subfigure}[b]{0.4\textwidth} \hspace{-20pt}
\includegraphics[width=4.8 cm,height=3.0 cm]{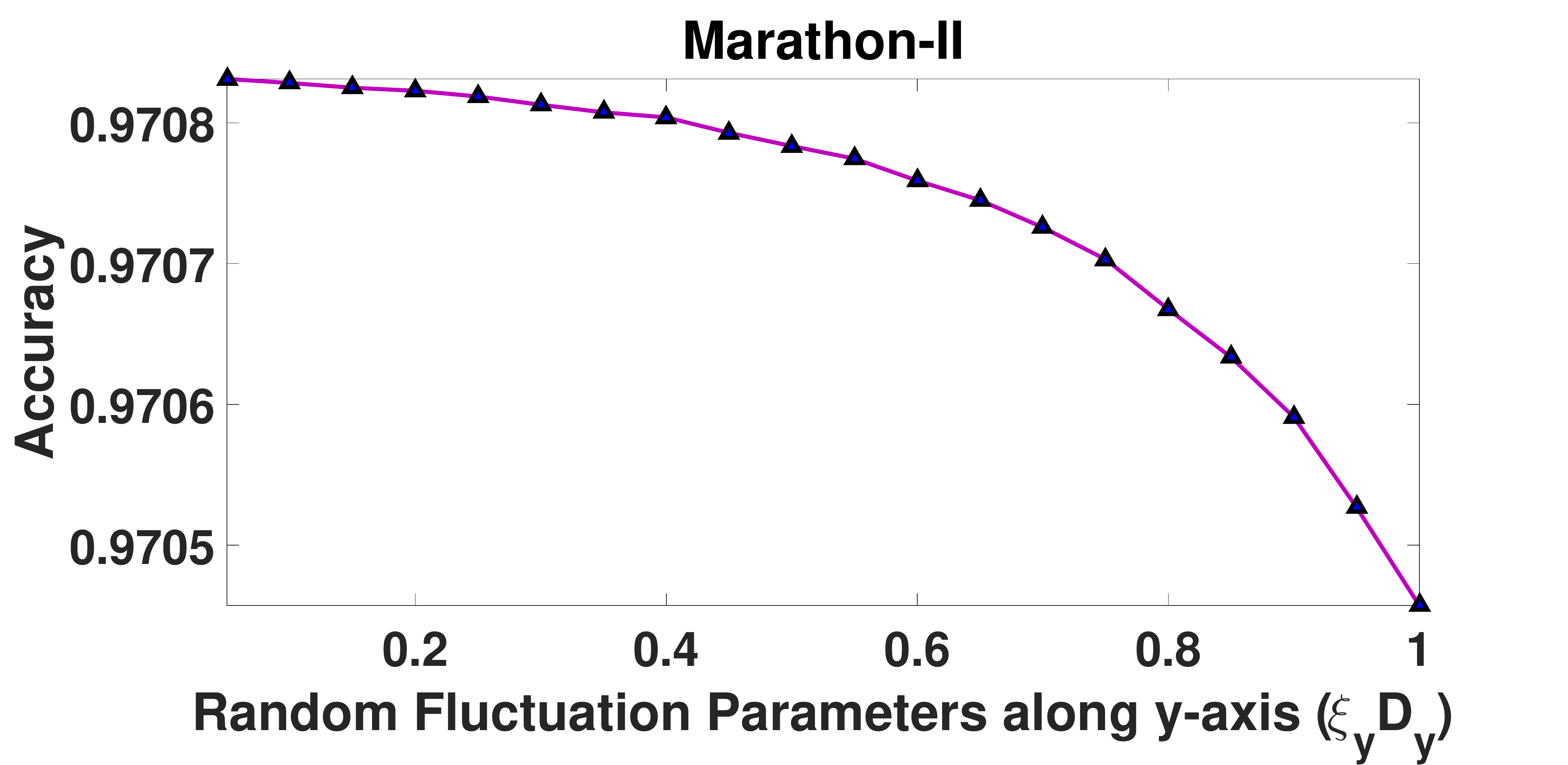}
        \caption{}
        \label{fig:m2_eydy}
    \end{subfigure}
    \hspace{0.4cm}
        \begin{subfigure}[b]{0.4\textwidth}
\includegraphics[width=4.8 cm,height=3.0 cm]{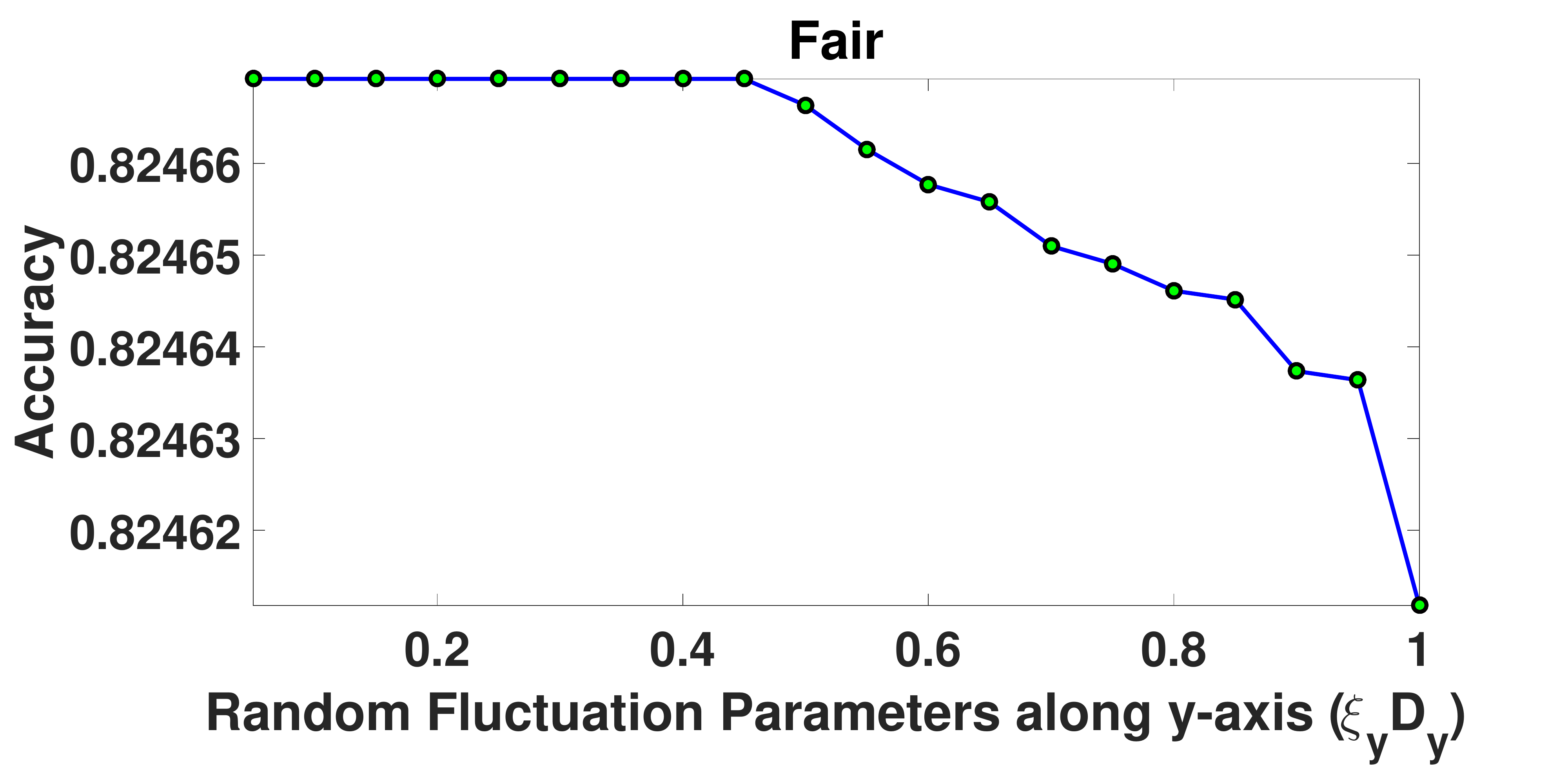}
        \caption{}
        \label{fig:fair_eydy}
    \end{subfigure}
    \hspace{0.1cm}
   %\hspace{1.0 cm}
   %\hspace{0.68 cm}
            \begin{subfigure}[b]{0.4\textwidth}
\includegraphics[width=4.8 cm,height=3.0 cm]{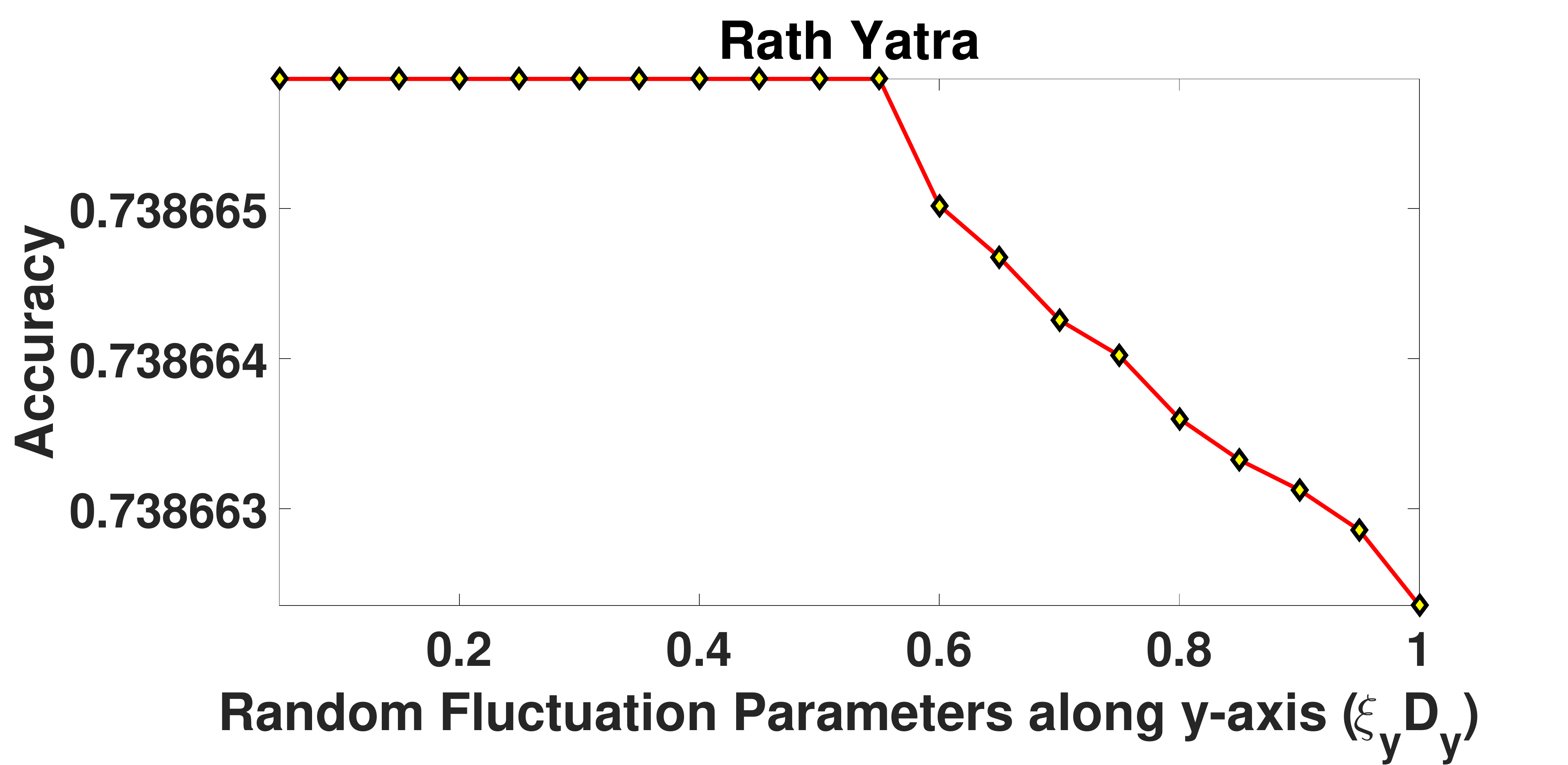}
        \caption{}
        \label{fig:RY_eydy}
    \end{subfigure}

    \caption{(a-d) Graphs showing how accuracy varies with respect to  random fluctuation force parameters along $y$-axis ($\xi_{y}D_{y}$) for different videos.}
    \label{fig:eydy}
    
\end{figure*}

\subsection{Ablation Experiment on the Proposed Force Model }
\label{section:AE} 
This sub-section discusses the results obtained from the ablation experiments performed on the proposed force model presented in (\ref{Eqn:ForceInertia}).
For these experiments, seven combinations of forces: $F_{external}$, $F_{drift/confine}$, $F_{disturbance}$, $F_{external}$ + $F_{drift/confine}$, $F_{external}$ + $F_{drift/confine}$ + $F_{disturbance}$, $F_{external}$ + $F_{disturbance}$ and $F_{drift/confine}$ + $F_{disturbance}$ have been formed in order to understand the importance of each force in the proposed force model. In order to understand the effect of the resistive forces and drift force, Marathon-II video has been chosen for experimentation. The experimental results shown in Fig.\ref{fig:AE1} reveal that the combination of all forces i.e. $F_{external}$+$F_{drift/confine}$+$F_{disturbance}$ has the highest accuracy among all other combinations and it is closer to ground truth. Furthermore, it can also be seen that the accuracy related to $F_{disturbance}$ is less which indicates that the video has less random behavior. Similarly, in order to demonstrate the effect of random force, Rath Yatra video is chosen. It can be seen in the Fig.\ref{fig:AE2} that the combination of all forces has the highest accuracy but most importantly, accuracy  due to $F_{disturbance}$ is overshadowing other forces at certain instances which clearly indicates the randomness in the video.

\begin{figure}
%\scriptsize
%\vspace{-10pt}
	\centering	
 \includegraphics[scale=0.60,width=0.75\textwidth]{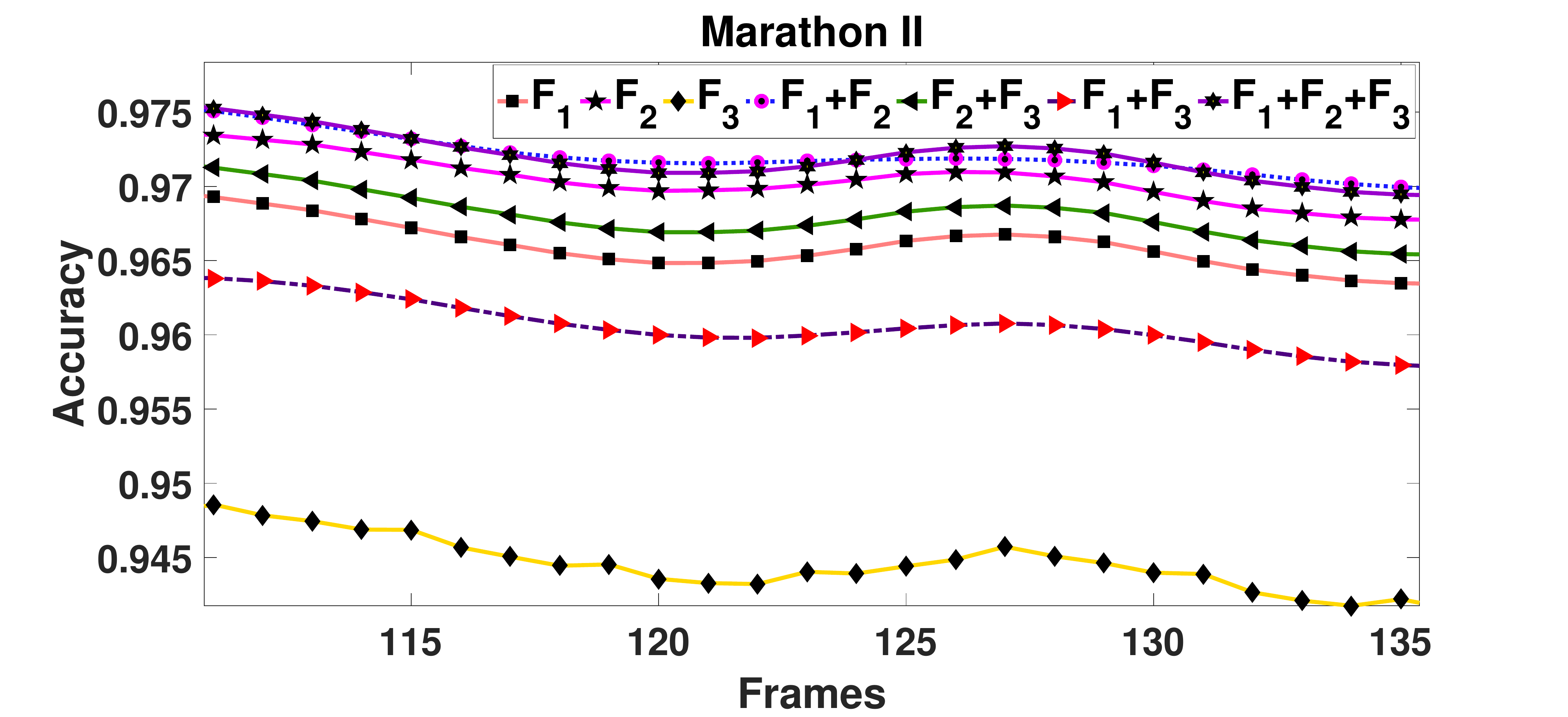}
    \caption{Ablation experiment results conducted on the force model proposed in (\ref{Eqn:ForceInertia}) performed on Marathon-II video to demonstrate the effect of $F_{external}$ and $F_{drift/confine}$. Here $F_{external}$, $F_{drift/confine}$, and $F_{disturbance}$ are considered as $F_{1}$, $F_{2}$, and $F_{3}$, respectively. (Best viewed in color)}
    \label{fig:AE1}
    %\vspace{-2 cm}
\end{figure}

\begin{figure}

%\vspace{-2 cm}
	\centering	
 \includegraphics[scale=0.60,width=0.75\textwidth]{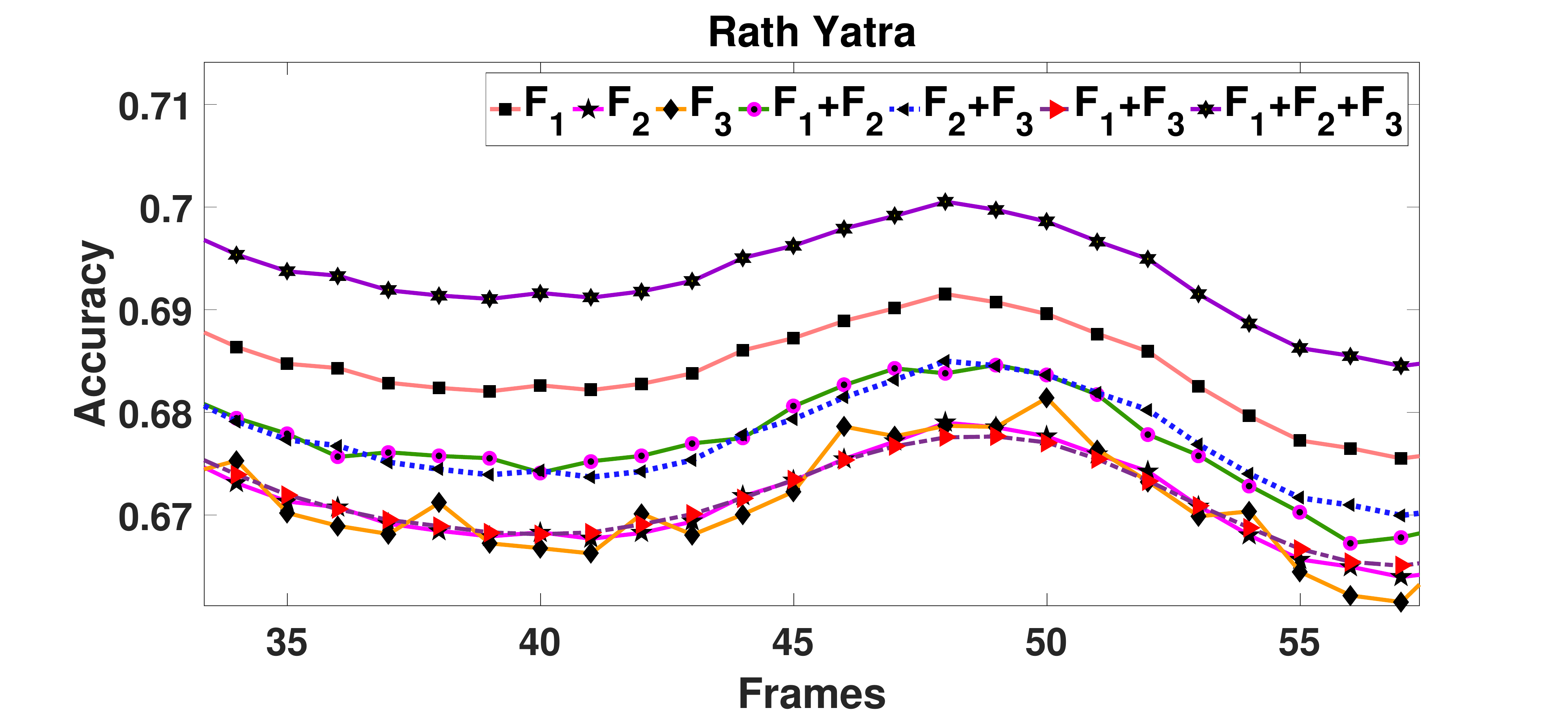}
    \caption{Ablation experiment results conducted on the force model proposed in (\ref{Eqn:ForceInertia}) performed on Rath Yatra video to demonstrate the effect of $F_{disturbance}$. Here, $F_{external}$, $F_{drift/confine}$, and $F_{disturbance}$ are considered as $F_{1}$, $F_{2}$, and $F_{3}$, respectively. (Best viewed in color)}
    \label{fig:AE2}
\end{figure}

\subsection{Flow Segmentation Results and Comparisons} 
\label{section:SR}
For comparisons, ground truths have been obtained by marking the dominant flows in the videos. The accuracy is calculated using (\ref{Eqn:accuracy}),
\begin{equation}
\label{Eqn:accuracy}
Accuracy=\frac{Area(S_{G} \cap G_{T})}{Area(G_{T})}
\end{equation}
where $S_{G}$ is the segmented image and $G_{T}$ is the ground truth image.

\par Proposed method generates $(W-1)$ segmented maps within a window. The first segmented map is the output obtained after grouping of optical flow keypoints based on spatial connectivity over a magnitude threshold of $0.4$. The other $(W-2)$ segmented maps are obtained using Langevin theory-based model.
The method has been compared with methods proposed in \citep{ali2007lagrangian}, \citep{santoro2010crowd}, \citep{zhou2014CC},and \citep{ullah2017density}, respectively. The outputs of \citep{ali2007lagrangian} and \citep{ullah2017density} is a segmentation mask which can be used directly in equation (\ref{Eqn:accuracy}). However, the method in \citep{santoro2010crowd} and \citep{zhou2014CC} produces outputs as clustered, tracked keypoints which are edge-grown, followed by morphological opening to get the segmentation mask. This mask is used for comparison with ground truth image.

\begin{figure}[htp]     %!h
    \centering
   % \vspace{-20pt}
   \captionsetup[subfigure]{labelformat=empty}
\begin{subfigure}[t]{0.20\textwidth}
\includegraphics[scale=0.1,width=0.8\textwidth]{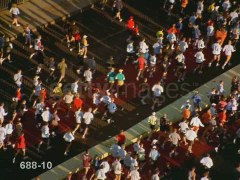}
        \caption{1}
        \label{fig:Original68816}
    \end{subfigure}
       \begin{subfigure}[t]{0.20\textwidth}
\includegraphics[scale=1.0,width=0.8\textwidth]{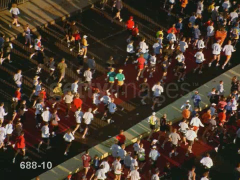}
        \caption{2}
        \label{fig:Original68817}
    \end{subfigure}
       \begin{subfigure}[t]{0.20\textwidth}
        \includegraphics[scale=0.1,width=0.8\textwidth]{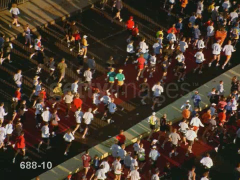}
        \caption{3}
        \label{fig:Original68818}
        \end{subfigure}
        \begin{subfigure}[t]{0.20\textwidth}
        \includegraphics[scale=0.1,width=0.8\textwidth]{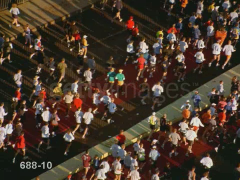}
        \caption{4}
        \label{fig:Original68819}
    \end{subfigure}
          \begin{subfigure}[t]{0.20\textwidth}
        \includegraphics[scale=0.1,width=0.8\textwidth]{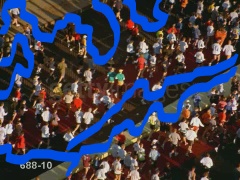}
        \caption{5}
        \label{fig:GT68816}
    \end{subfigure}
       \begin{subfigure}[t]{0.20\textwidth}
        \includegraphics[scale=0.1,width=0.8\textwidth]{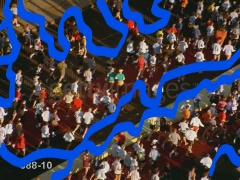}
        \caption{6}
        \label{fig:GT68817}
    \end{subfigure}
       \begin{subfigure}[t]{0.20\textwidth}
        \includegraphics[scale=0.1,width=0.8\textwidth]{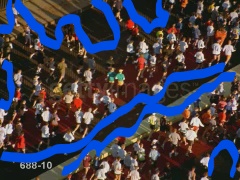}
        \caption{7}
        \label{fig:GT68818}
        \end{subfigure}
        \begin{subfigure}[t]{0.20\textwidth}
        \includegraphics[scale=0.1,width=0.8\textwidth]{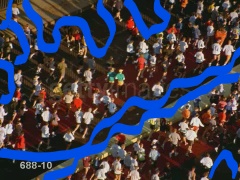}
        \caption{8}
        \label{fig:GT68819}
    \end{subfigure}
        \begin{subfigure}[t]{0.20\textwidth}
        \includegraphics[scale=0.1,width=0.8\textwidth]{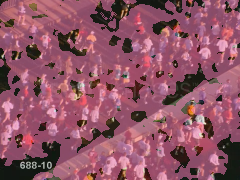}
        \caption{9}
        \label{fig:CP68816}
    \end{subfigure}
       \begin{subfigure}[t]{0.20\textwidth}
        \includegraphics[scale=0.1,width=0.8\textwidth]{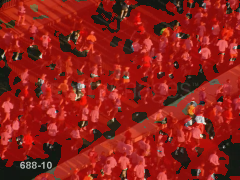}
        \caption{10}
        \label{fig:CP68817}
    \end{subfigure}
       \begin{subfigure}[t]{0.20\textwidth}
        \includegraphics[scale=0.1,width=0.8\textwidth]{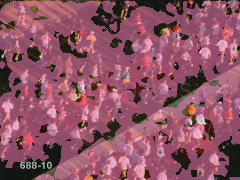}
        \caption{11}
        \label{fig:CP68818}
        \end{subfigure}
        \begin{subfigure}[t]{0.20\textwidth}
        \includegraphics[scale=0.1,width=0.8\textwidth]{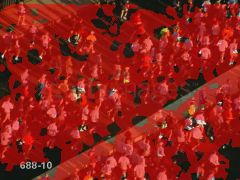}
        \caption{12}
        \label{fig:CP68819}
    \end{subfigure}
                   \begin{subfigure}[t]{0.20\textwidth}
        \includegraphics[scale=1.0,width=0.8\textwidth]{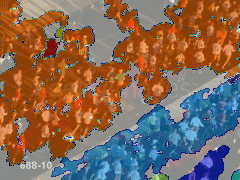}
        \caption{13}
        \label{fig:LD68816}
    \end{subfigure}
       \begin{subfigure}[t]{0.20\textwidth}
        \includegraphics[scale=0.1,width=0.8\textwidth]{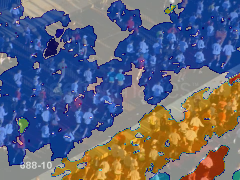}
        \caption{14}
        \label{fig:LD68817}
    \end{subfigure}
       \begin{subfigure}[t]{0.20\textwidth}
        \includegraphics[scale=0.1,width=0.8\textwidth]{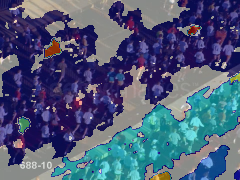}
        \caption{15}
        \label{fig:LD68818}
        \end{subfigure}
        \begin{subfigure}[t]{0.20\textwidth}
        \includegraphics[scale=0.1,width=0.8\textwidth]{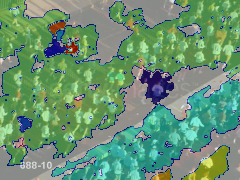}
        \caption{16}
        \label{fig:LD68819}
        \end{subfigure}
           \begin{subfigure}[t]{0.20\textwidth}
        \includegraphics[scale=0.1,width=0.8\textwidth]{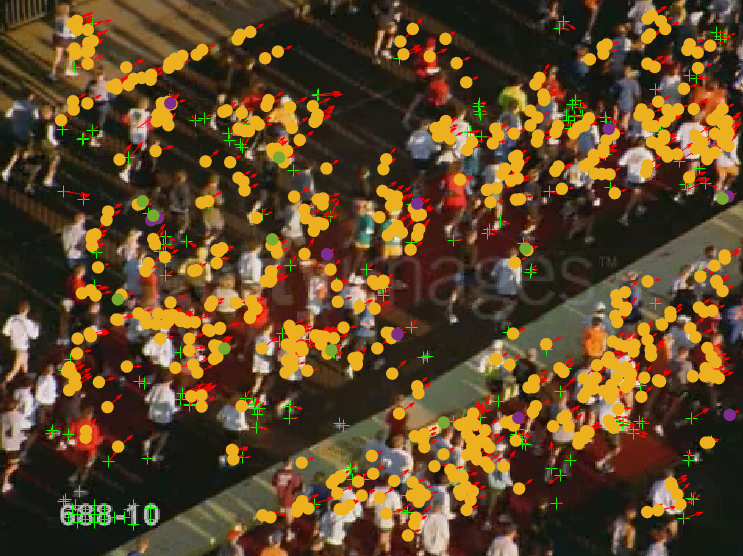}
        \caption{17}
        \label{fig:Santoro68816}
        \end{subfigure}
               \begin{subfigure}[t]{0.20\textwidth}
        \includegraphics[scale=0.1,width=0.8\textwidth]{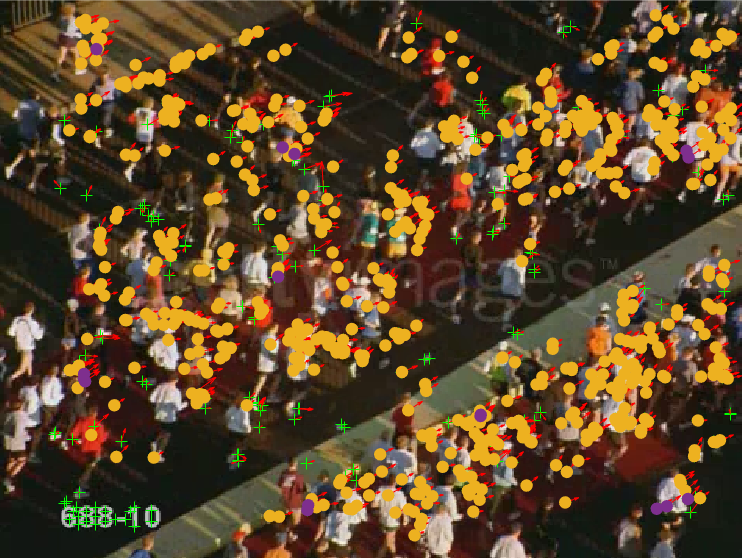}
        \caption{18}
        \label{fig:Santoro68817}
        \end{subfigure}
               \begin{subfigure}[t]{0.20\textwidth}
        \includegraphics[scale=0.1,width=0.8\textwidth]{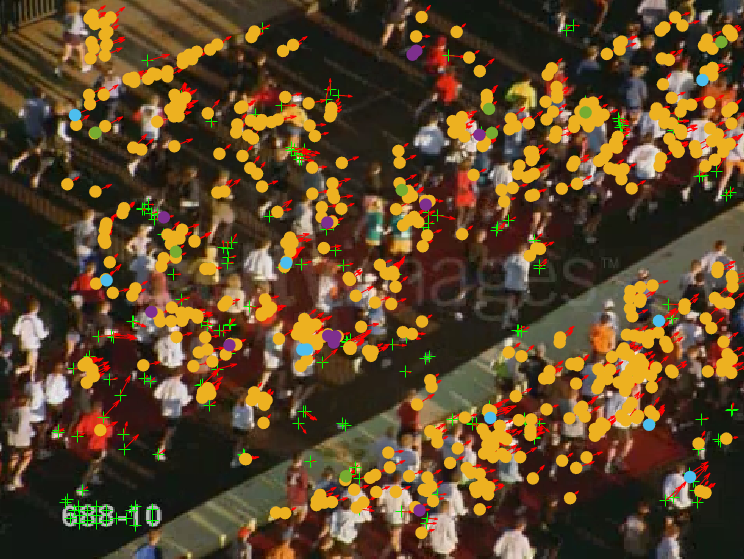}
        \caption{19}
        \label{fig:Santoro68818}
        \end{subfigure}
               \begin{subfigure}[t]{0.20\textwidth}
        \includegraphics[scale=0.1,width=0.8\textwidth]{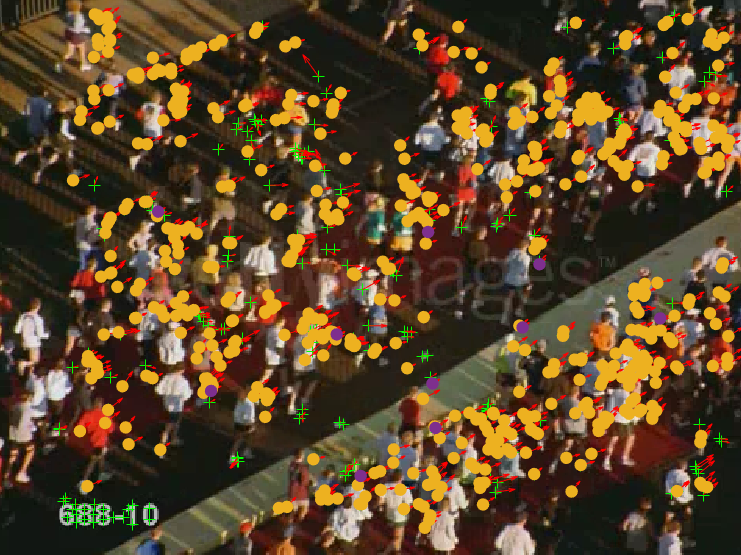}
        \caption{20}
        \label{fig:Santoro68819}
        \end{subfigure}
        \begin{subfigure}[t]{0.20\textwidth}
        \includegraphics[scale=0.1,width=0.8\textwidth]{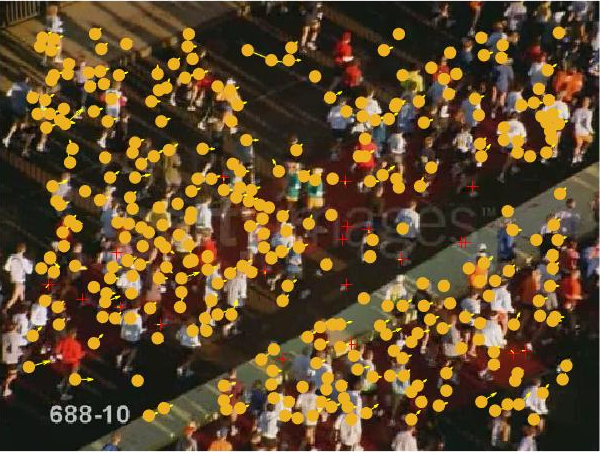}
        \caption{21}
        \label{fig:CC68816}
        \end{subfigure}
               \begin{subfigure}[t]{0.20\textwidth}
        \includegraphics[scale=0.1,width=0.8\textwidth]{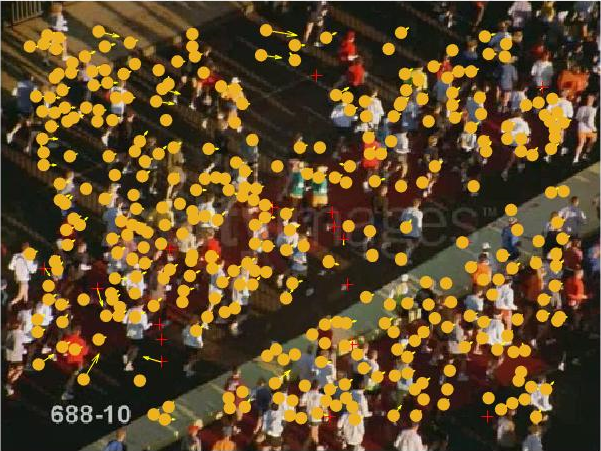}
        \caption{22}
        \label{fig:CC68817}
        \end{subfigure}
               \begin{subfigure}[t]{0.20\textwidth}
        \includegraphics[scale=0.1,width=0.8\textwidth]{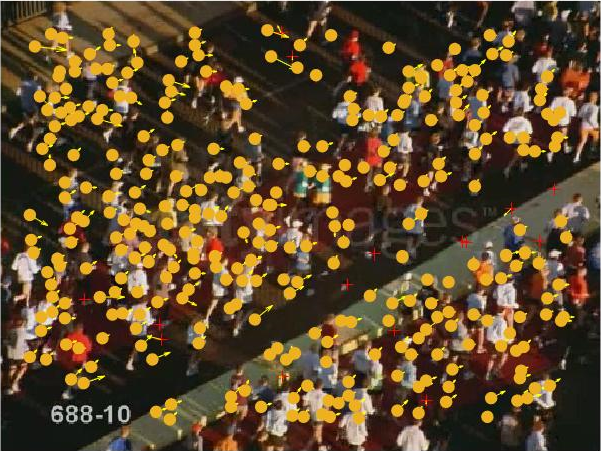}
        \caption{23}
        \label{fig:CC68818}
        \end{subfigure}
               \begin{subfigure}[t]{0.20\textwidth}
        \includegraphics[scale=0.1,width=0.8\textwidth]{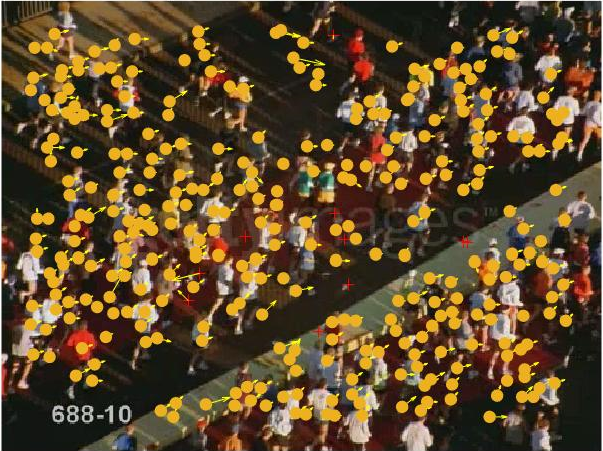}
        \caption{24}
        \label{fig:CC68819}
        \end{subfigure}
        \begin{subfigure}[t]{0.20\textwidth}
        \includegraphics[scale=0.1,width=0.8\textwidth]{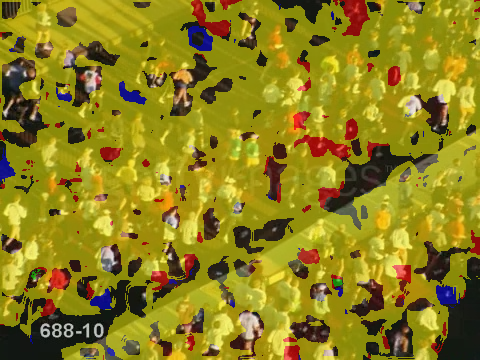}
        \caption{25}
        \label{fig:DIHM68816}
        \end{subfigure}
               \begin{subfigure}[t]{0.20\textwidth}
        \includegraphics[scale=0.1,width=0.8\textwidth]{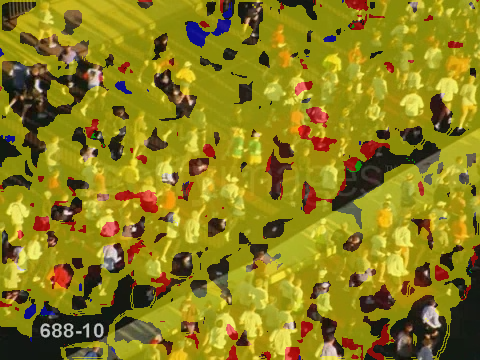}
        \caption{26}
        \label{fig:DIHM68817}
        \end{subfigure}
               \begin{subfigure}[t]{0.20\textwidth}
        \includegraphics[scale=0.1,width=0.8\textwidth]{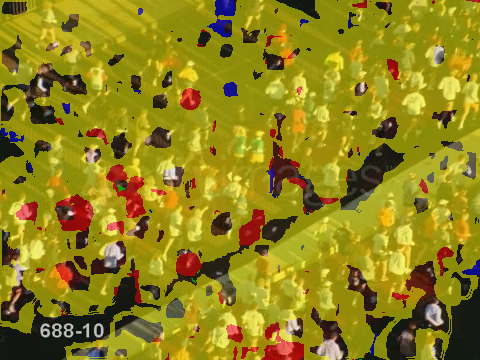}
        \caption{27}
        \label{fig:DIHM68818}
        \end{subfigure}
               \begin{subfigure}[t]{0.20\textwidth}
        \includegraphics[scale=0.1,width=0.8\textwidth]{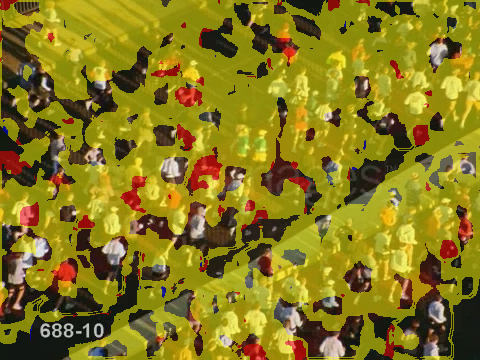}
        \caption{28}
        \label{fig:DIHM68819}
        \end{subfigure}
     \caption{($1$-$4$) Original recorded Frames~(16-19) of the Marathon-I video,  ($5$-$8$) Ground Truth Frames, ($9$-$12$) represent segmented outputs obtained using~proposed method, ($13$-$16$) represent outputs of segmentation method~\citep{ali2007lagrangian}, ($17$-$20$) represent outputs of segmentation method~\citep{santoro2010crowd}, ($21$-$24$) represent outputs of segmentation method~\citep{zhou2014CC} and ($25$-$28$) represent outputs of segmentation using~\citep{ullah2017density}, respectively. (Best viewed in color)}

\label{fig:688}
\end{figure}
%Marathon II
\begin{figure}[htp]     %!h
    \centering
    \captionsetup[subfigure]{labelformat=empty}
 %   \vspace{-20pt}
\begin{subfigure}[t]{0.20\textwidth}
        \includegraphics[scale=0.1,width=0.8\textwidth]{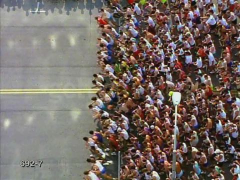}
        \caption{1}
        \label{fig:Original69251}
    \end{subfigure}
       \begin{subfigure}[t]{0.20\textwidth}
        \includegraphics[scale=0.1,width=0.8\textwidth]{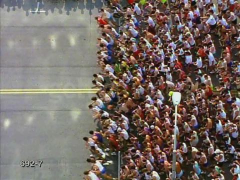}
        \caption{2}
        \label{fig:Original69252}
    \end{subfigure}
       \begin{subfigure}[t]{0.20\textwidth}
        \includegraphics[scale=0.1,width=0.8\textwidth]{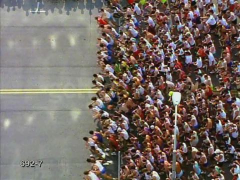}
        \caption{3}
        \label{fig:Original69253}
        \end{subfigure}
        \begin{subfigure}[t]{0.20\textwidth}
        \includegraphics[scale=0.1,width=0.8\textwidth]{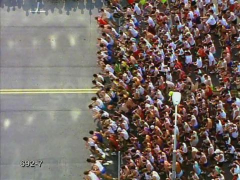}
        \caption{4}
        \label{fig:Original69254}
    \end{subfigure}
          \begin{subfigure}[t]{0.20\textwidth}
        \includegraphics[scale=0.1,width=0.8\textwidth]{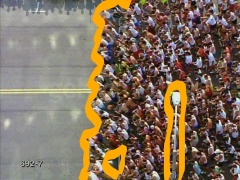}
        \caption{5}
        \label{fig:GT69251}
    \end{subfigure}
       \begin{subfigure}[t]{0.20\textwidth}
        \includegraphics[scale=0.1,width=0.8\textwidth]{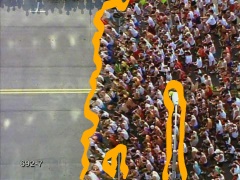}
        \caption{6}
        \label{fig:GT69252}
    \end{subfigure}
       \begin{subfigure}[t]{0.20\textwidth}
        \includegraphics[scale=0.1,width=0.8\textwidth]{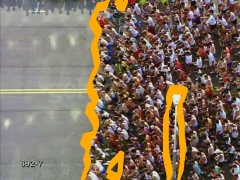}
        \caption{7}
        \label{fig:GT69253}
        \end{subfigure}
        \begin{subfigure}[t]{0.20\textwidth}
        \includegraphics[scale=0.1,width=0.8\textwidth]{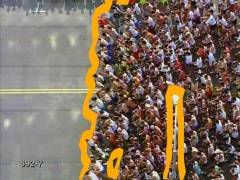}
        \caption{8}
        \label{fig:GT69254}
    \end{subfigure}
        \begin{subfigure}[t]{0.20\textwidth}
        \includegraphics[scale=0.1,width=0.8\textwidth]{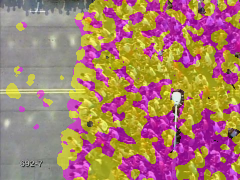}
        \caption{9}
        \label{fig:CP69251}
    \end{subfigure}
       \begin{subfigure}[t]{0.20\textwidth}
        \includegraphics[scale=0.1,width=0.8\textwidth]{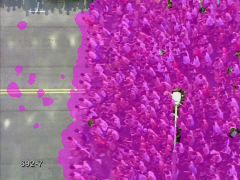}
        \caption{10}
        \label{fig:CP69252}
    \end{subfigure}
       \begin{subfigure}[t]{0.20\textwidth}
        \includegraphics[scale=0.1,width=0.8\textwidth]{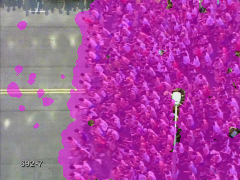}   
        \caption{11}
        \label{fig:CP69253}
        \end{subfigure}
        \begin{subfigure}[t]{0.20\textwidth}
        \includegraphics[scale=0.1,width=0.8\textwidth]{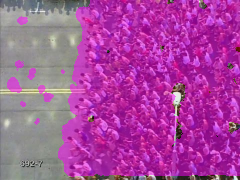}
        \caption{12}
        \label{fig:CP69254}
    \end{subfigure}
               \begin{subfigure}[t]{0.20\textwidth}
        \includegraphics[scale=0.1,width=0.8\textwidth]{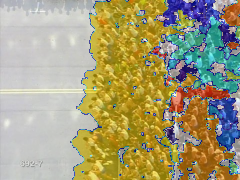}
        \caption{13}
        \label{fig:LD69251}
    \end{subfigure}
       \begin{subfigure}[t]{0.20\textwidth}
        \includegraphics[scale=0.1,width=0.8\textwidth]{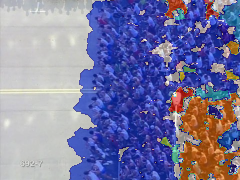}
        \caption{14}
        \label{fig:LD69252}
    \end{subfigure}
       \begin{subfigure}[t]{0.20\textwidth}
        \includegraphics[scale=0.1,width=0.8\textwidth]{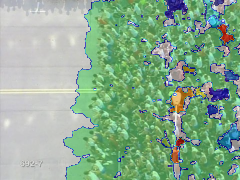}
        \caption{15}
        \label{fig:LD69253}
        \end{subfigure}
        \begin{subfigure}[t]{0.20\textwidth}
        \includegraphics[scale=0.1,width=0.8\textwidth]{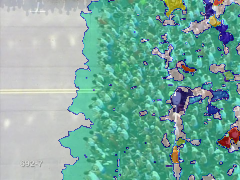}
        \caption{16}
        \label{fig:LD69254}
    \end{subfigure}
                  \begin{subfigure}[t]{0.20\textwidth}
        \includegraphics[scale=0.1,width=0.8\textwidth]{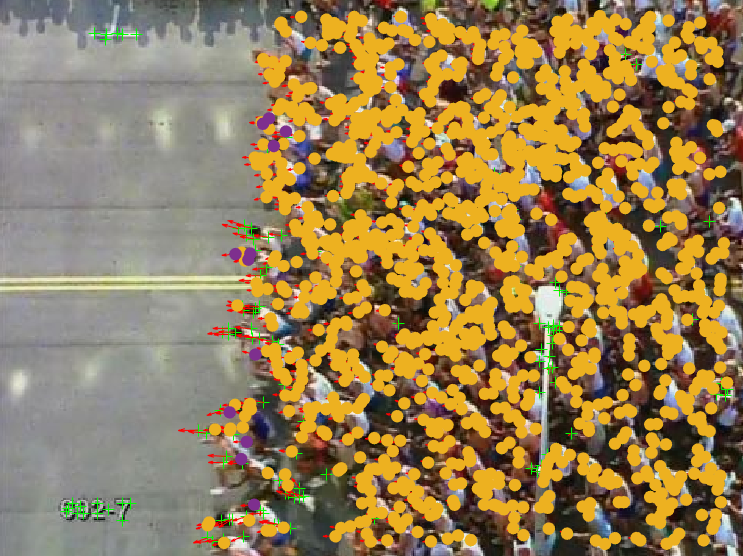}
        \caption{17}
        \label{fig:Santoro69251}
        \end{subfigure}
               \begin{subfigure}[t]{0.20\textwidth}
        \includegraphics[scale=0.1,width=0.8\textwidth]{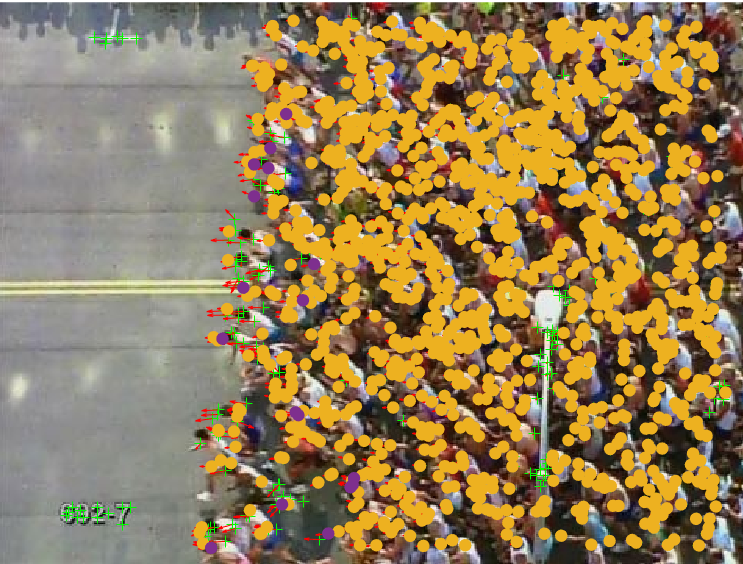}
        \caption{18}
        \label{fig:Santoro69252}
        \end{subfigure}
               \begin{subfigure}[t]{0.20\textwidth}
        \includegraphics[scale=0.1,width=0.8\textwidth]{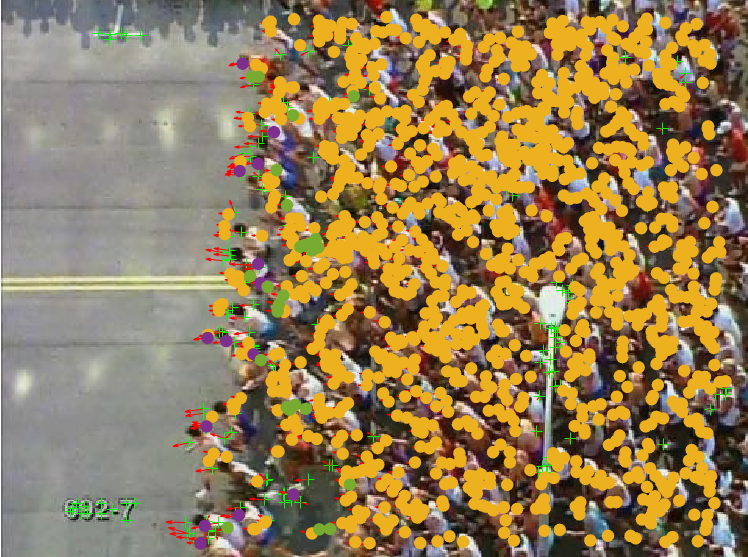}
        \caption{19}
        \label{fig:Santoro69253}
        \end{subfigure}
               \begin{subfigure}[t]{0.20\textwidth}
        \includegraphics[scale=0.1,width=0.8\textwidth]{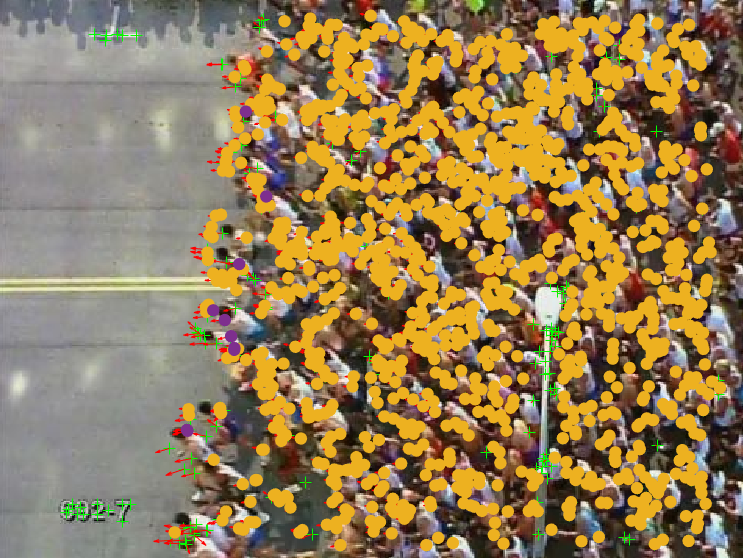}
        \caption{20}
        \label{fig:Santoro69254}
        \end{subfigure}
    \begin{subfigure}[t]{0.20\textwidth}
        \includegraphics[scale=0.1,width=0.8\textwidth]{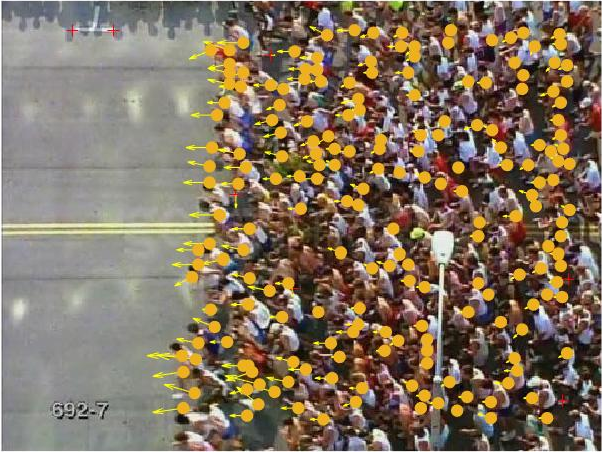}
        \caption{21}
        \label{fig:CC69251}
        \end{subfigure}
               \begin{subfigure}[t]{0.20\textwidth}
        \includegraphics[scale=0.1,width=0.8\textwidth]{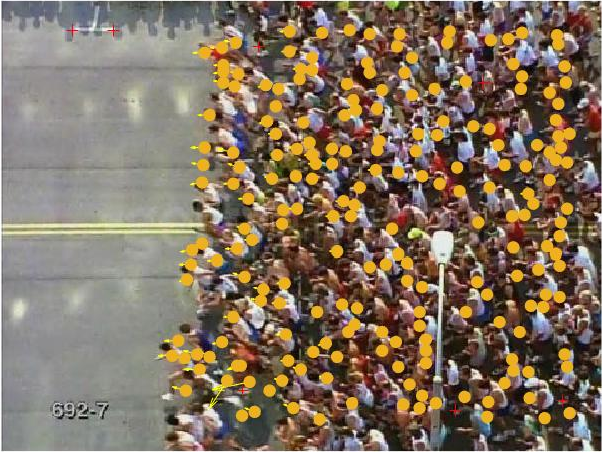}
        \caption{22}
        \label{fig:CC69252}
        \end{subfigure}
               \begin{subfigure}[t]{0.20\textwidth}
        \includegraphics[scale=0.1,width=0.8\textwidth]{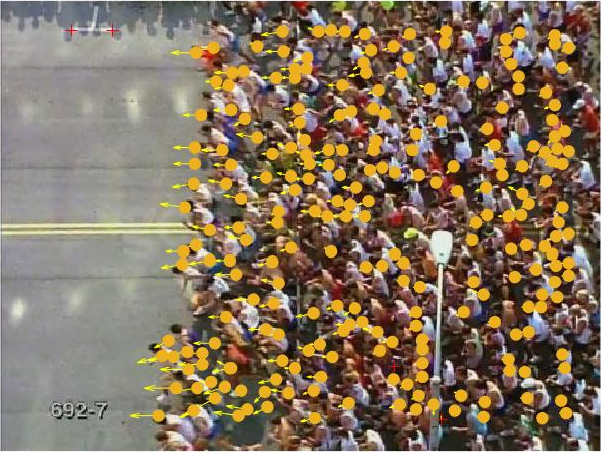}
        \caption{23}
        \label{fig:CC69253}
        \end{subfigure}
               \begin{subfigure}[t]{0.20\textwidth}
        \includegraphics[scale=0.1,width=0.8\textwidth]{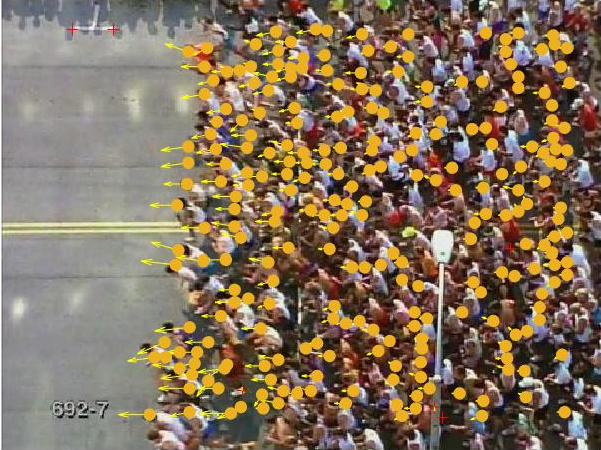}
        \caption{24}
        \label{fig:CC69254}
        \end{subfigure}
            \begin{subfigure}[t]{0.20\textwidth}
        \includegraphics[scale=0.1,width=0.8\textwidth]{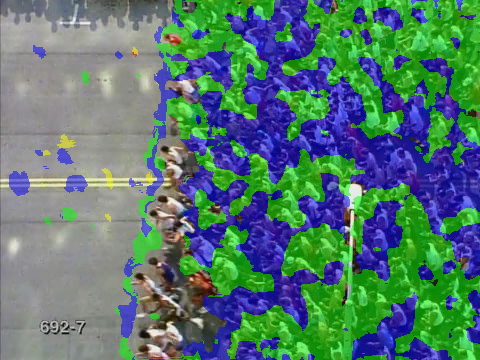}
        \caption{25}
        \label{fig:DIHM69251}
        \end{subfigure}
               \begin{subfigure}[t]{0.20\textwidth}
        \includegraphics[scale=0.1,width=0.8\textwidth]{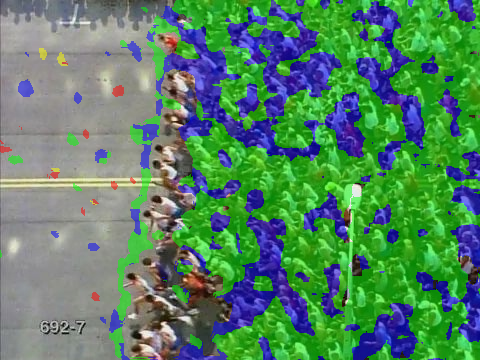}
        \caption{26}
        \label{fig:DIHM69252}
        \end{subfigure}
               \begin{subfigure}[t]{0.20\textwidth}
        \includegraphics[scale=0.1,width=0.8\textwidth]{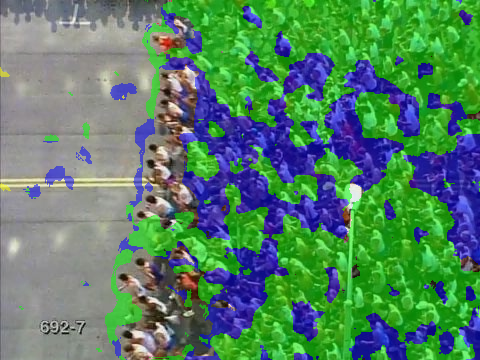}
        \caption{27}
        \label{fig:DIHM69253}
        \end{subfigure}
               \begin{subfigure}[t]{0.20\textwidth}
        \includegraphics[scale=0.1,width=0.8\textwidth]{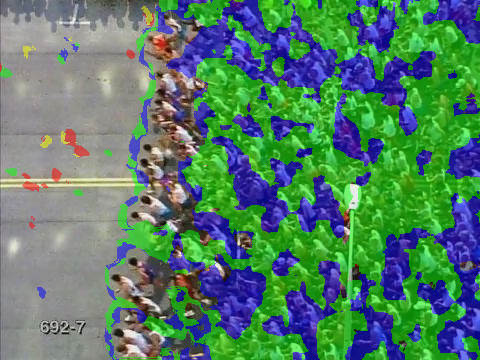}
        \caption{28}
        \label{fig:DIHM69254}
        \end{subfigure}
     \caption{($1$-$4$) Original recorded Frames~(51-54) of the Marathon-II video, ($5$-$8$) Ground Truth Frames, ($9$-$12$) represent segmented outputs obtained using~proposed method, ($13$-$16$) represent outputs of segmentation method~\citep{ali2007lagrangian}, ($17$-$20$) represent outputs of segmentation method~\citep{santoro2010crowd}, ($21$-$24$) represent outputs of segmentation method~\citep{zhou2014CC} and ($25$-$28$) represent outputs of segmentation using~\citep{ullah2017density}, respectively. (Best viewed in color)}

\label{fig:692}
\end{figure}
%------------------------------------------------------------------------
%Fair
\begin{figure}[t!]     %!h
    \centering
    \captionsetup[subfigure]{labelformat=empty}
  %  \vspace{-20pt}
\begin{subfigure}[t]{0.20\textwidth}
        \includegraphics[scale=0.1,width=0.8\textwidth]{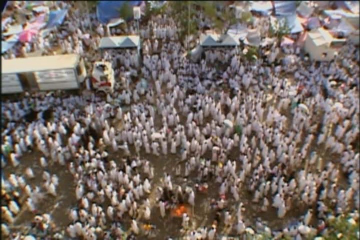}
        \caption{1}
        \label{fig:OriginalFair41}
    \end{subfigure}
       \begin{subfigure}[t]{0.20\textwidth}
        \includegraphics[scale=0.1,width=0.8\textwidth]{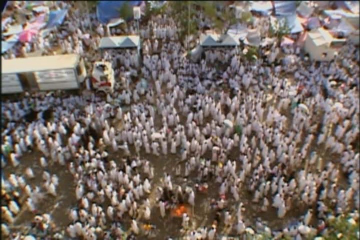}
        \caption{2}
        \label{fig:OriginalFair42}
    \end{subfigure}
       \begin{subfigure}[t]{0.20\textwidth}
        \includegraphics[scale=0.1,width=0.8\textwidth]{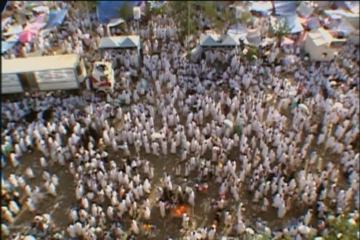}
        \caption{3}
        \label{fig:OriginalFair43}
        \end{subfigure}
        \begin{subfigure}[t]{0.20\textwidth}
        \includegraphics[scale=0.1,width=0.8\textwidth]{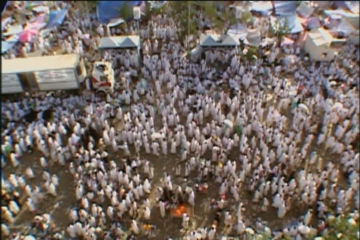}
        \caption{4}
        \label{fig:OriginalFair44}
    \end{subfigure}
          \begin{subfigure}[t]{0.20\textwidth}
        \includegraphics[scale=0.1,width=0.8\textwidth]{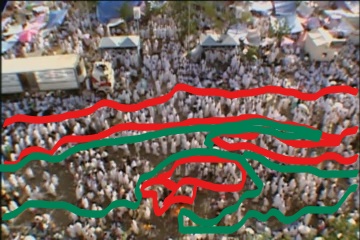}
        \caption{5}
        \label{fig:GTFair41}
    \end{subfigure}
       \begin{subfigure}[t]{0.20\textwidth}
        \includegraphics[scale=0.1,width=0.8\textwidth]{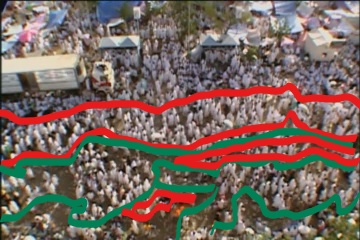}
        \caption{6}
        \label{fig:GTFair42}
    \end{subfigure}
       \begin{subfigure}[t]{0.20\textwidth}
        \includegraphics[scale=0.1,width=0.8\textwidth]{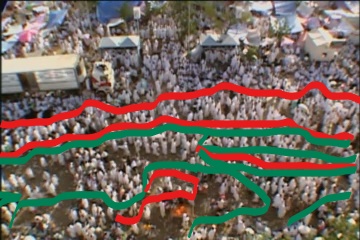}
        \caption{7}
        \label{fig:GTFair43}
        \end{subfigure}
        \begin{subfigure}[t]{0.20\textwidth}
        \includegraphics[scale=0.1,width=0.8\textwidth]{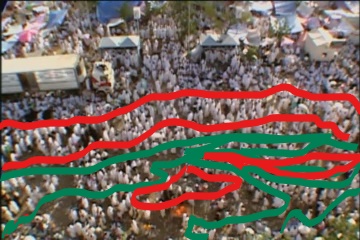}
        \caption{8}
        \label{fig:GTFair44}
    \end{subfigure}
        \begin{subfigure}[t]{0.20\textwidth}
        \includegraphics[scale=0.1,width=0.8\textwidth]{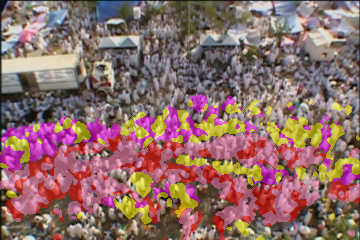}
        \caption{9}
        \label{fig:CPFair41}
    \end{subfigure}
       \begin{subfigure}[t]{0.20\textwidth}
        \includegraphics[scale=0.1,width=0.8\textwidth]{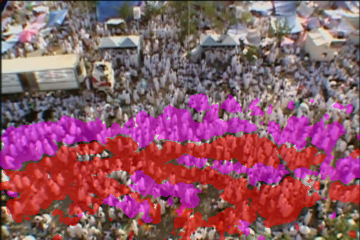}
        \caption{10}
        \label{fig:CPFair42}
    \end{subfigure}
       \begin{subfigure}[t]{0.20\textwidth}
        \includegraphics[scale=0.1,width=0.8\textwidth]{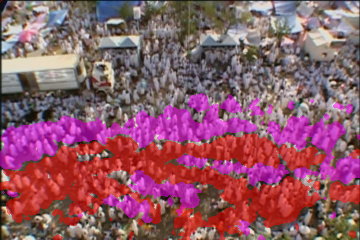} 
        \caption{11}
        \label{fig:CPFair43}
        \end{subfigure}
        \begin{subfigure}[t]{0.20\textwidth}
        \includegraphics[scale=0.1,width=0.8\textwidth]{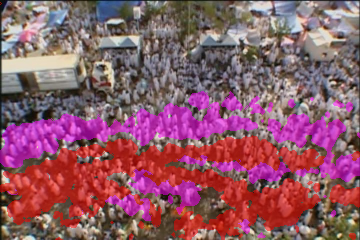}
        \caption{12}
        \label{fig:CPFair44}
    \end{subfigure}
           \begin{subfigure}[t]{0.20\textwidth}
        \includegraphics[scale=0.1,width=0.8\textwidth]{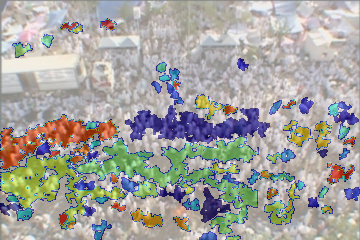}
        \caption{13}
        \label{fig:LDFair41}
    \end{subfigure}
       \begin{subfigure}[t]{0.20\textwidth}
        \includegraphics[scale=0.1,width=0.8\textwidth]{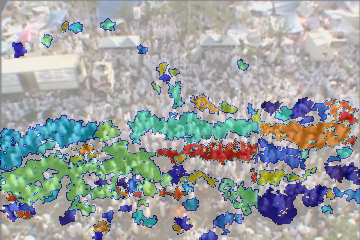}
        \caption{14}
        \label{fig:LDFair42}
    \end{subfigure}
       \begin{subfigure}[t]{0.20\textwidth}
        \includegraphics[scale=0.1,width=0.8\textwidth]{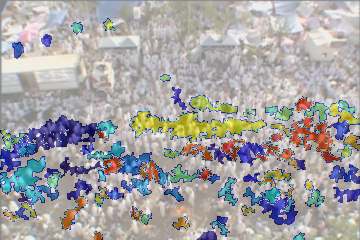}
        \caption{15}
        \label{fig:LDFair43}
        \end{subfigure}
        \begin{subfigure}[t]{0.20\textwidth}
        \includegraphics[scale=0.1,width=0.8\textwidth]{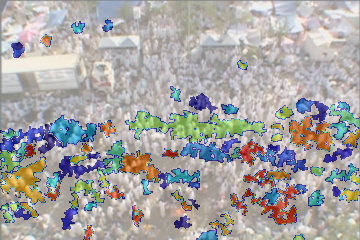}
        \caption{16}
        \label{fig:LDFair44}
    \end{subfigure}
                  \begin{subfigure}[t]{0.20\textwidth}
        \includegraphics[scale=0.1,width=0.8\textwidth]{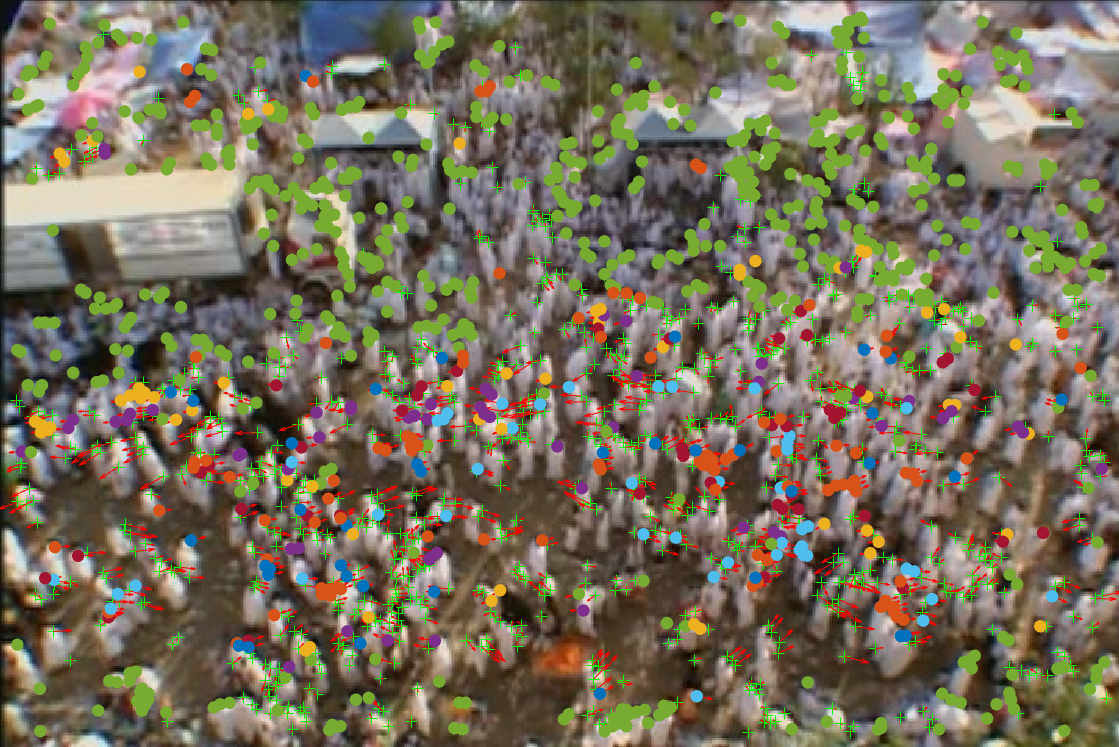}
        \caption{17}
        \label{fig:SantoroFair41}
        \end{subfigure}
               \begin{subfigure}[t]{0.20\textwidth}
        \includegraphics[scale=0.1,width=0.8\textwidth]{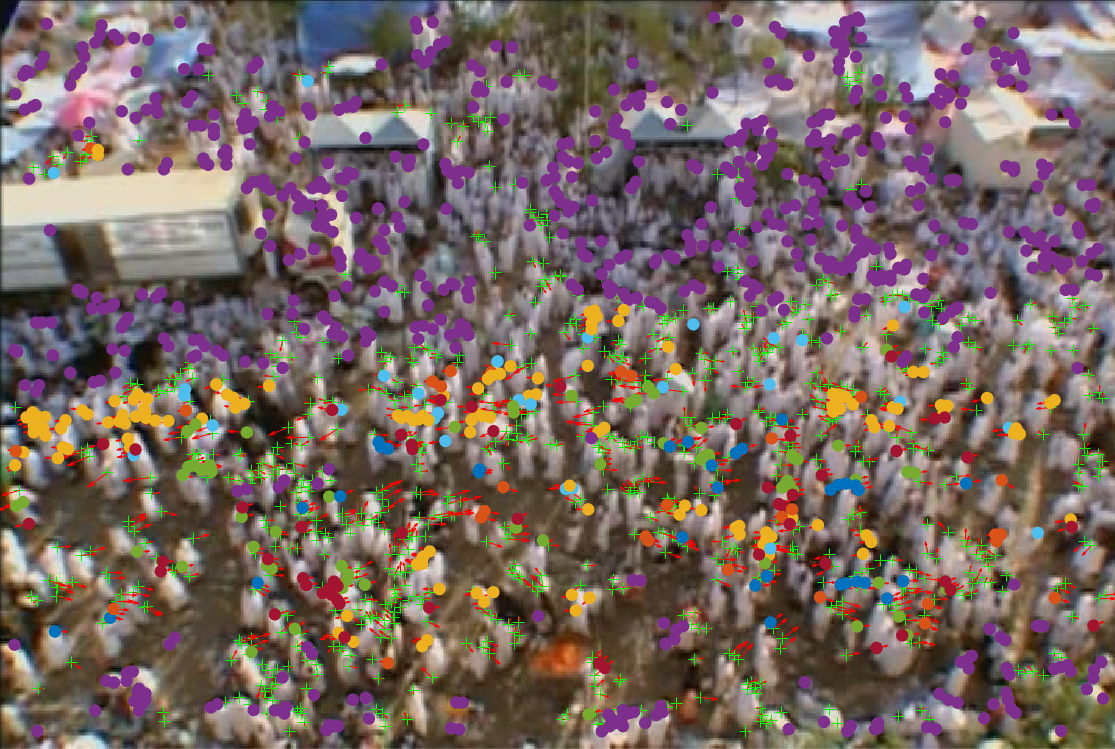}
        \caption{18}
        \label{fig:SantoroFair42}
        \end{subfigure}
               \begin{subfigure}[t]{0.20\textwidth}
        \includegraphics[scale=0.1,width=0.8\textwidth]{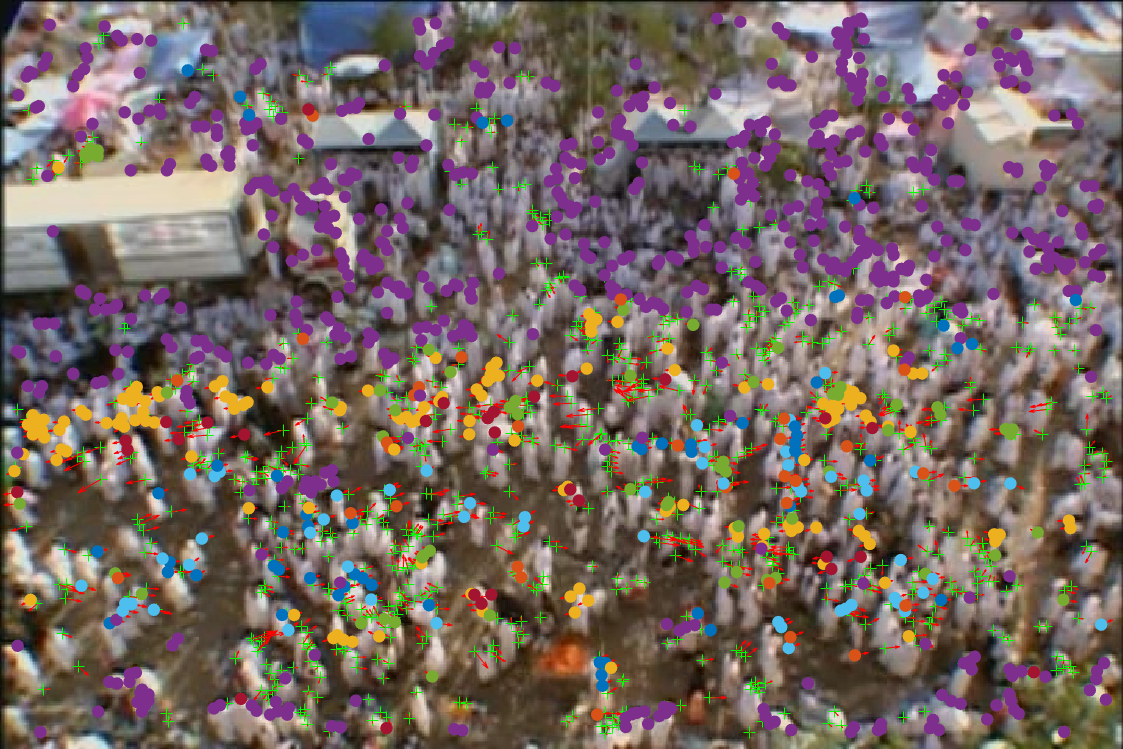}
        \caption{19}
        \label{fig:SantoroFair43}
        \end{subfigure}
               \begin{subfigure}[t]{0.20\textwidth}
        \includegraphics[scale=0.1,width=0.8\textwidth]{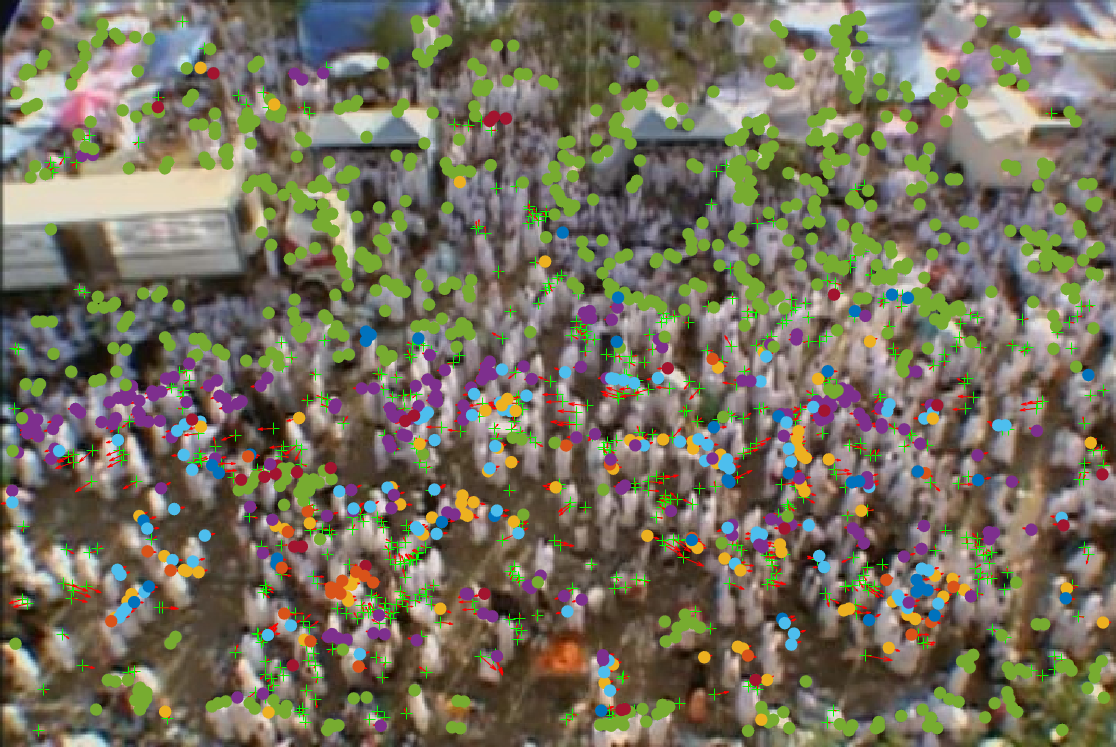}
        \caption{20}
        \label{fig:SantoroFair44}
        \end{subfigure}
     \begin{subfigure}[t]{0.20\textwidth}
        \includegraphics[scale=0.1,width=0.8\textwidth]{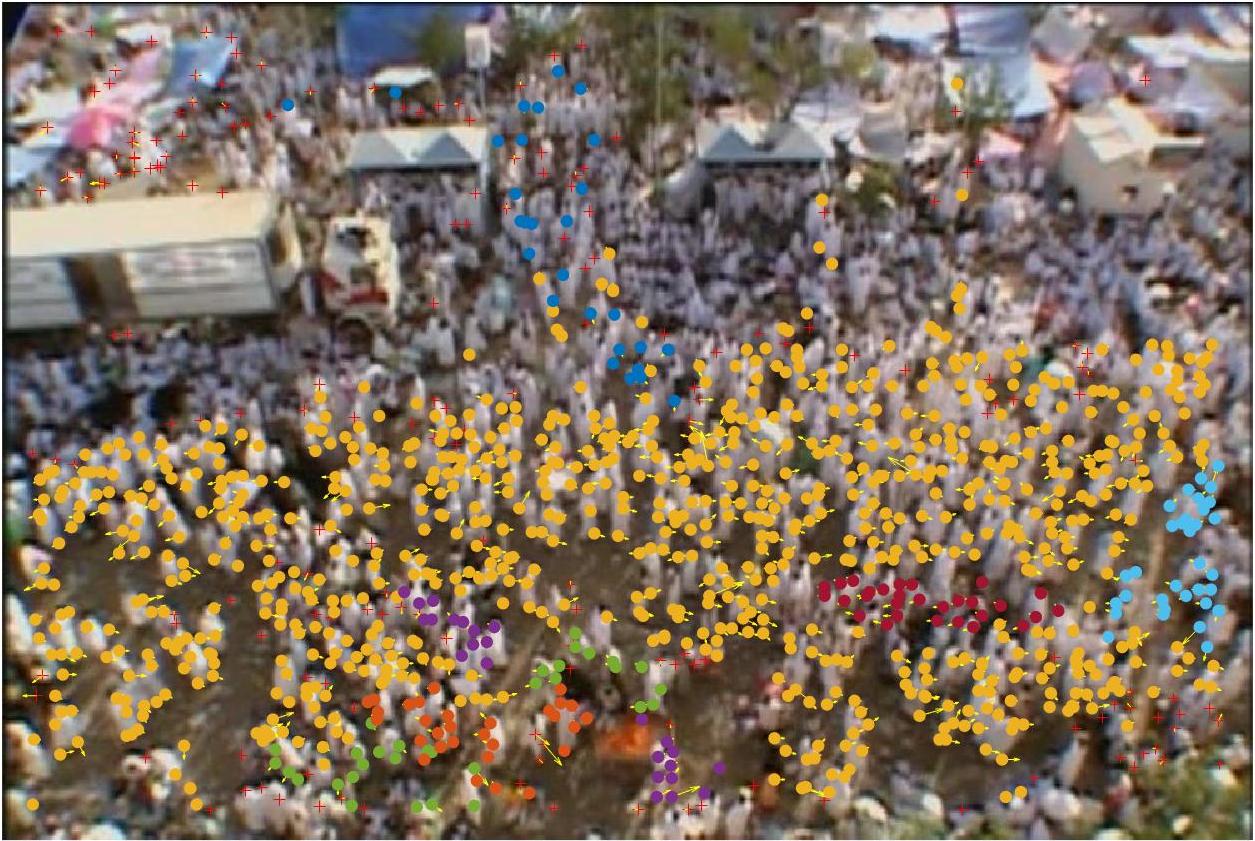}
        \caption{21}
        \label{fig:CCFair41}
        \end{subfigure}
               \begin{subfigure}[t]{0.20\textwidth}
        \includegraphics[scale=0.1,width=0.8\textwidth]{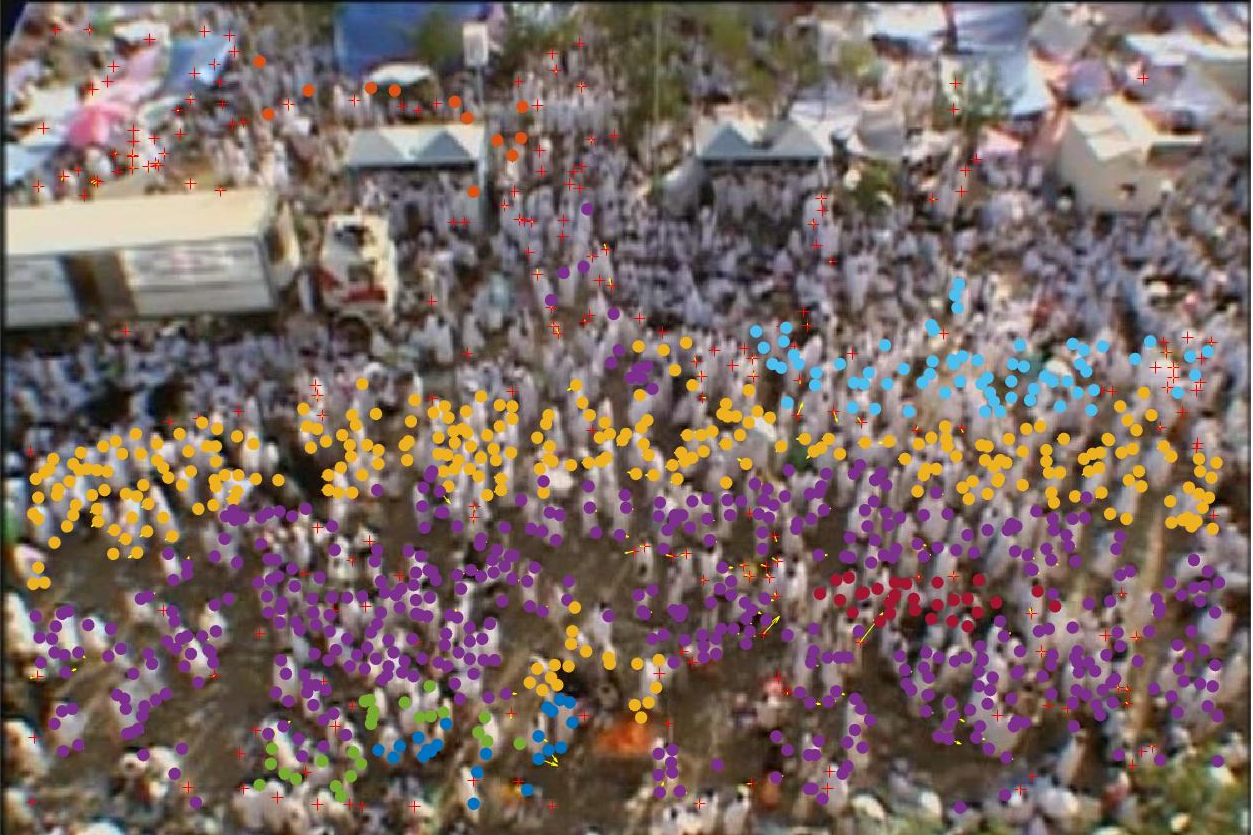}
        \caption{22}
        \label{fig:CCFair42}
        \end{subfigure}
               \begin{subfigure}[t]{0.20\textwidth}
        \includegraphics[scale=0.1,width=0.8\textwidth]{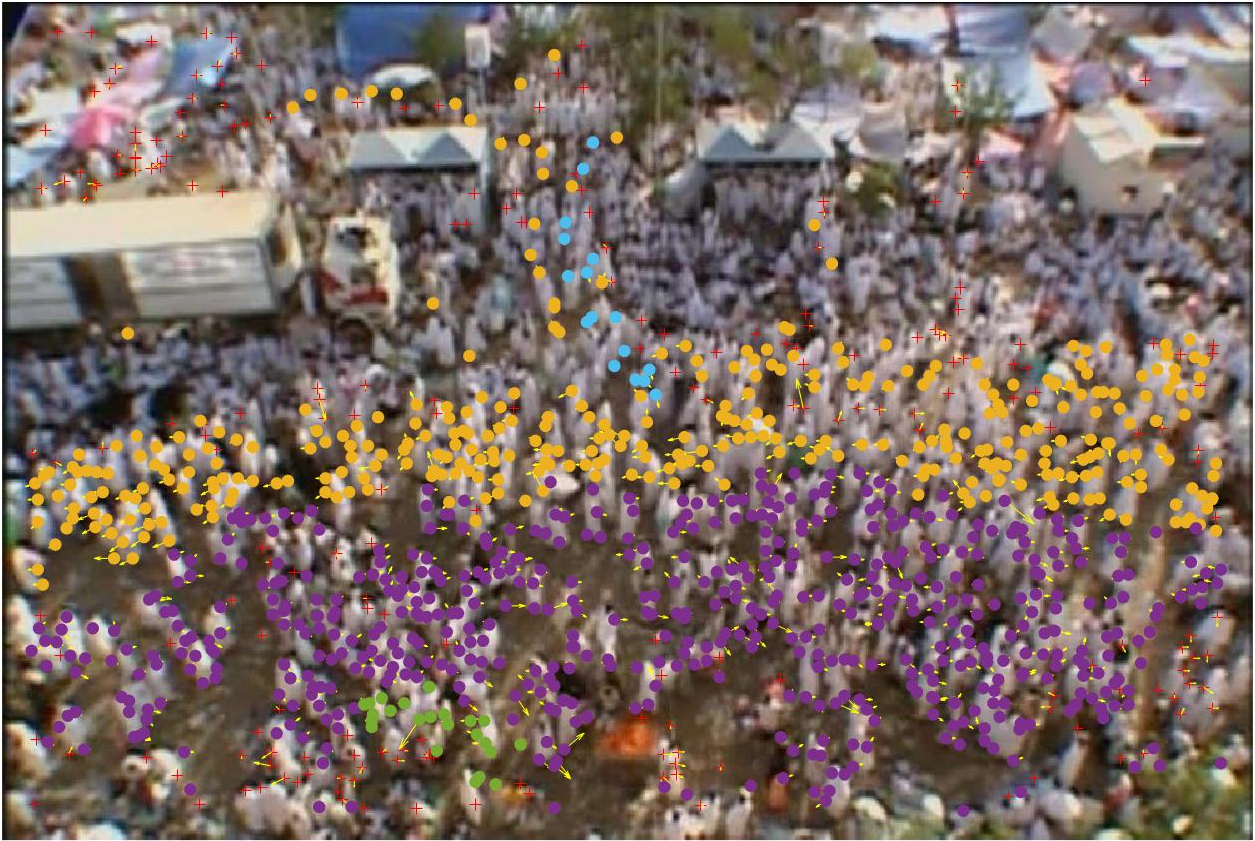}
        \caption{23}
        \label{fig:CCFair43}
        \end{subfigure}
               \begin{subfigure}[t]{0.20\textwidth}
        \includegraphics[scale=0.1,width=0.8\textwidth]{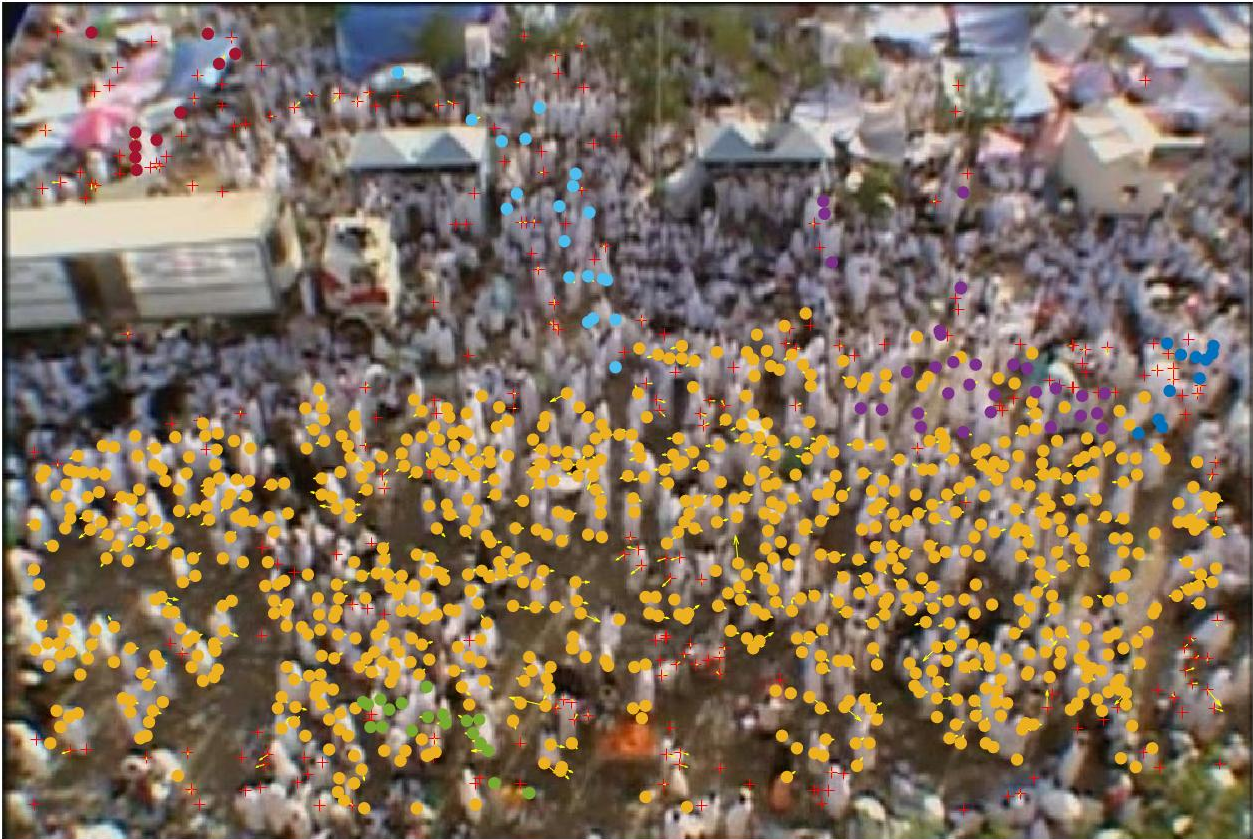}
        \caption{24}
        \label{fig:CCFair44}
        \end{subfigure}
            \begin{subfigure}[t]{0.20\textwidth}
        \includegraphics[scale=0.1,width=0.8\textwidth]{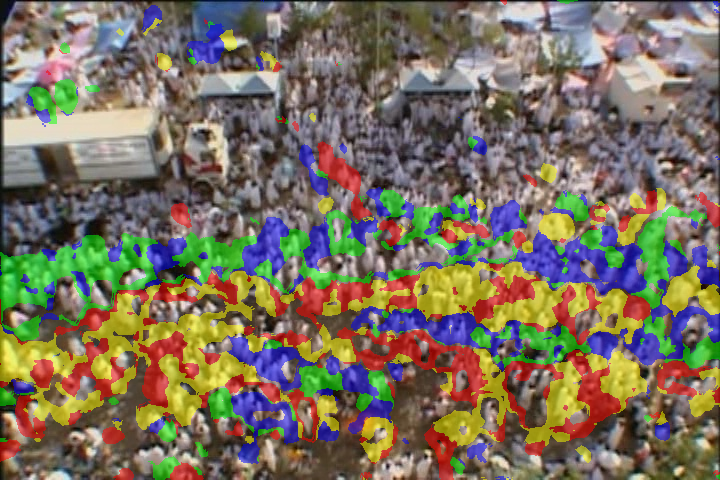}
        \caption{25}
        \label{fig:DIHMFair41}
        \end{subfigure}
               \begin{subfigure}[t]{0.20\textwidth}
        \includegraphics[scale=0.1,width=0.8\textwidth]{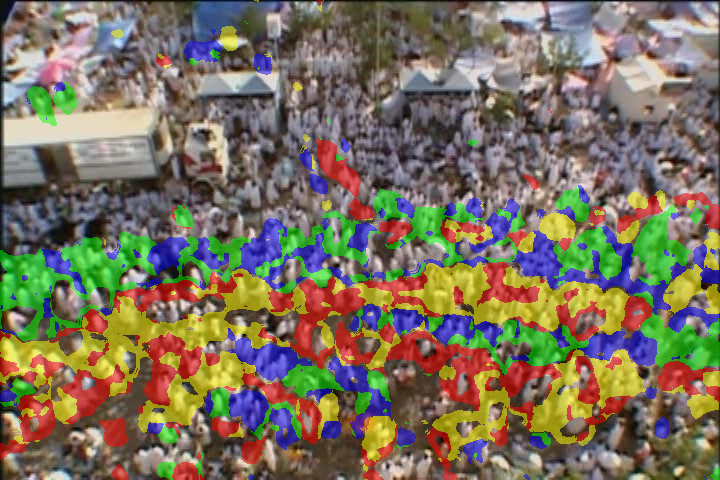}
        \caption{26}
        \label{fig:DIHMFair42}
        \end{subfigure}
               \begin{subfigure}[t]{0.20\textwidth}
        \includegraphics[scale=0.1,width=0.8\textwidth]{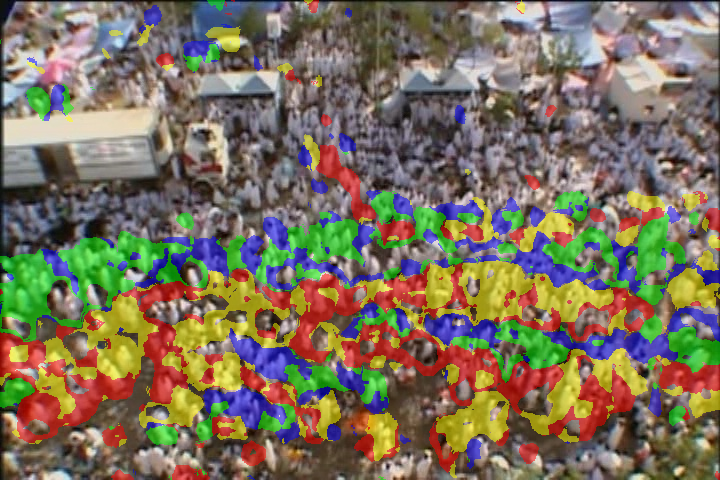}
        \caption{27}
        \label{fig:DIHMFair43}
        \end{subfigure}
               \begin{subfigure}[t]{0.20\textwidth}
        \includegraphics[scale=0.1,width=0.8\textwidth]{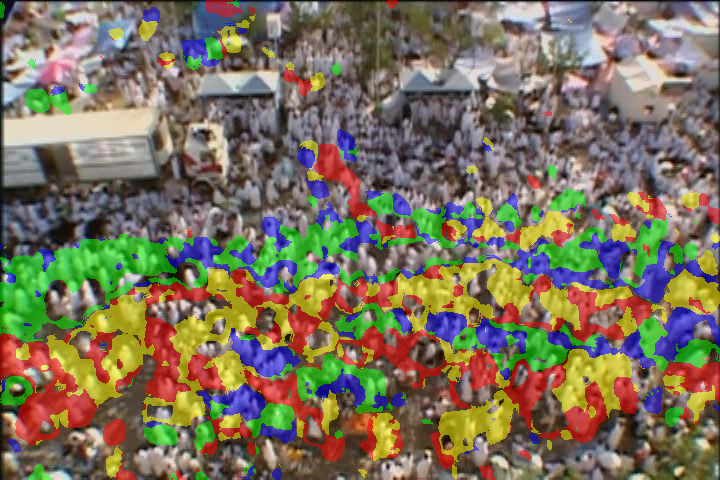}
        \caption{28}
        \label{fig:DIHMFair44}
        \end{subfigure}
     \caption{($1$-$4$) Original recorded Frames~(41-44) of the Fair video, ($5$-$8$) Ground Truth Frames, ($9$-$12$) represent segmented outputs obtained using~proposed method, ($13$-$16$) represent outputs of segmentation method~\citep{ali2007lagrangian}, ($17$-$20$) represent outputs of segmentation method~\citep{santoro2010crowd}, ($21$-$24$) represent outputs of segmentation method~\citep{zhou2014CC} and ($25$-$28$) represent outputs of segmentation using~\citep{ullah2017density}, respectively. (Best viewed in color)}

\label{fig:Fair}
\end{figure}
%---------------------------------------------------------------
%RY
\begin{figure}[htp]     %!h
   \centering
   \captionsetup[subfigure]{labelformat=empty}
 %   \vspace{-20pt}
\begin{subfigure}[t]{0.20\textwidth}
        \includegraphics[scale=0.1,width=0.8\textwidth]{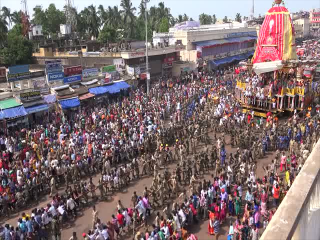}
        \caption{1}
        \label{fig:OriginalRY31}
    \end{subfigure}
       \begin{subfigure}[t]{0.20\textwidth}
        \includegraphics[scale=0.1,width=0.8\textwidth]{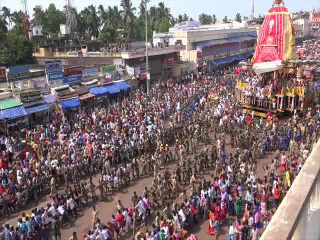}
        \caption{2}
        \label{fig:OriginalRY32}
    \end{subfigure}
       \begin{subfigure}[t]{0.20\textwidth}
        \includegraphics[scale=0.1,width=0.8\textwidth]{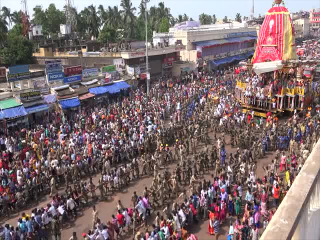}
        \caption{3}
        \label{fig:OriginalRY33}
        \end{subfigure}
        \begin{subfigure}[t]{0.20\textwidth}
        \includegraphics[scale=0.1,width=0.8\textwidth]{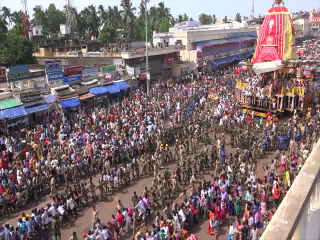}
        \caption{4}
        \label{fig:OriginalRY34}
    \end{subfigure}
          \begin{subfigure}[t]{0.20\textwidth}
        \includegraphics[scale=0.1,width=0.8\textwidth]{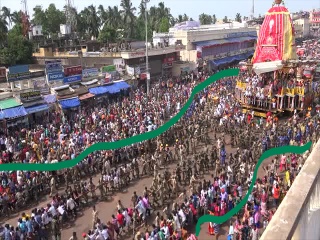}
        \caption{5}
        \label{fig:GTRY31}
    \end{subfigure}
       \begin{subfigure}[t]{0.20\textwidth}
        \includegraphics[scale=0.1,width=0.8\textwidth]{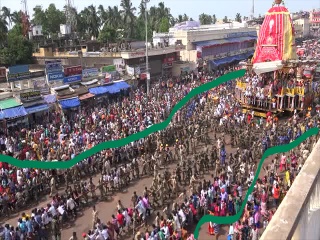}
        \caption{6}
        \label{fig:GTRY32}
    \end{subfigure}
       \begin{subfigure}[t]{0.20\textwidth}
        \includegraphics[scale=0.1,width=0.8\textwidth]{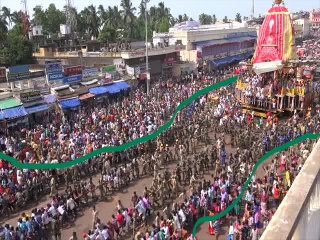}
        \caption{7}
        \label{fig:GTRY33}
        \end{subfigure}
        \begin{subfigure}[t]{0.20\textwidth}
        \includegraphics[scale=0.1,width=0.8\textwidth]{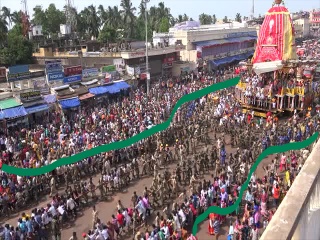}
        \caption{8}
        \label{fig:GTRY34}
    \end{subfigure}
        \begin{subfigure}[t]{0.20\textwidth}
        \includegraphics[scale=0.1,width=0.8\textwidth]{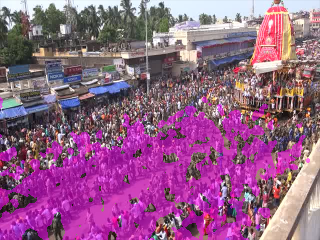}
        \caption{9}
        \label{fig:CPRY31}
    \end{subfigure}
       \begin{subfigure}[t]{0.20\textwidth}
        \includegraphics[scale=0.1,width=0.8\textwidth]{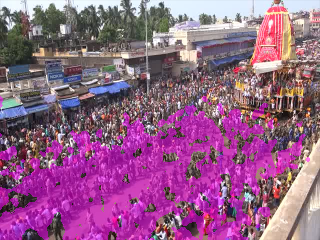}
        \caption{10}
        \label{fig:CPRY32}
    \end{subfigure}
       \begin{subfigure}[t]{0.20\textwidth}
        \includegraphics[scale=0.1,width=0.8\textwidth]{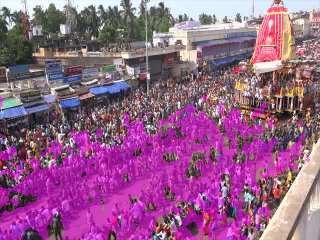}
        \caption{11}
        \label{fig:CPRY33}
        \end{subfigure}
        \begin{subfigure}[t]{0.20\textwidth}
        \includegraphics[scale=0.1,width=0.8\textwidth]{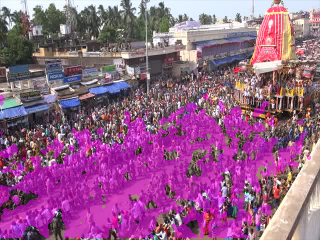}
        \caption{12}
        \label{fig:CPRY34}
         \end{subfigure}
        \begin{subfigure}[t]{0.20\textwidth}
        \includegraphics[scale=0.1,width=0.8\textwidth]{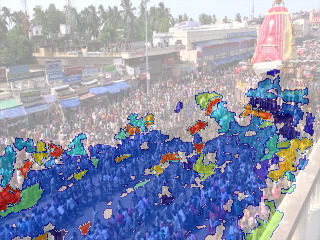}
        \caption{13}
        \label{fig:LDRY31}
    \end{subfigure}
       \begin{subfigure}[t]{0.20\textwidth}
        \includegraphics[scale=0.1,width=0.8\textwidth]{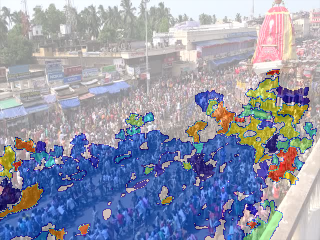}
        \caption{14}
        \label{fig:LDRY32}
    \end{subfigure}
       \begin{subfigure}[t]{0.20\textwidth}
        \includegraphics[scale=0.1,width=0.8\textwidth]{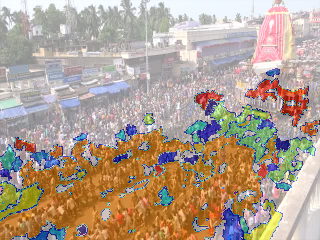}
        \caption{15}
        \label{fig:LDRY33}
        \end{subfigure}
        \begin{subfigure}[t]{0.20\textwidth}
        \includegraphics[scale=0.1,width=0.8\textwidth]{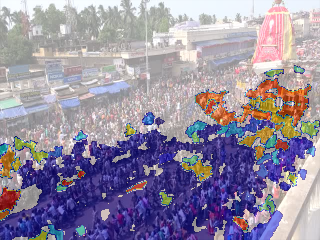}
        \caption{16}
        \label{fig:LDRY34}
    \end{subfigure}
              \begin{subfigure}[t]{0.20\textwidth}
        \includegraphics[scale=0.1,width=0.8\textwidth]{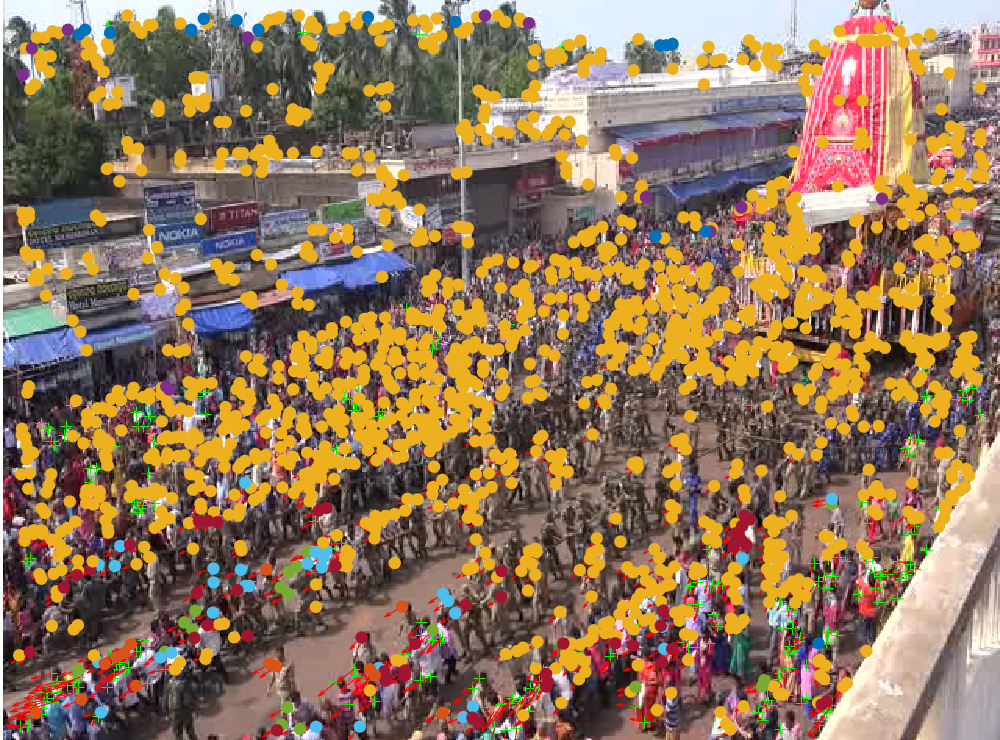}
        \caption{17}
        \label{fig:SantoroRY31}
        \end{subfigure}
               \begin{subfigure}[t]{0.20\textwidth}
        \includegraphics[scale=0.1,width=0.8\textwidth]{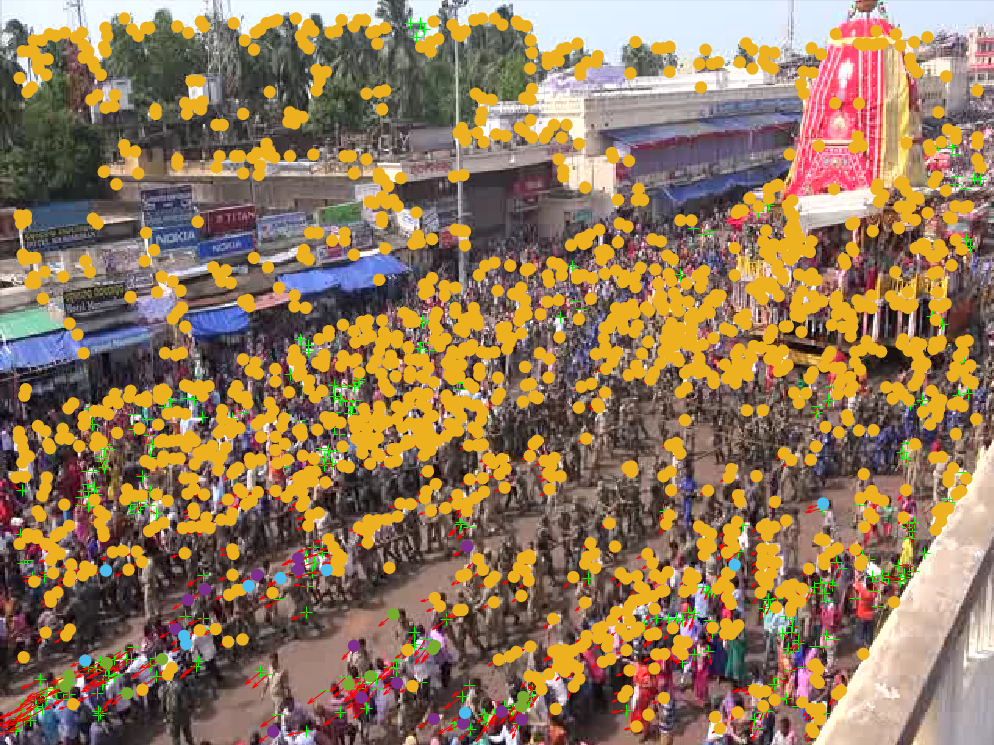}
        \caption{18}
        \label{fig:SantoroRY32}
        \end{subfigure}
               \begin{subfigure}[t]{0.20\textwidth}
        \includegraphics[scale=0.1,width=0.8\textwidth]{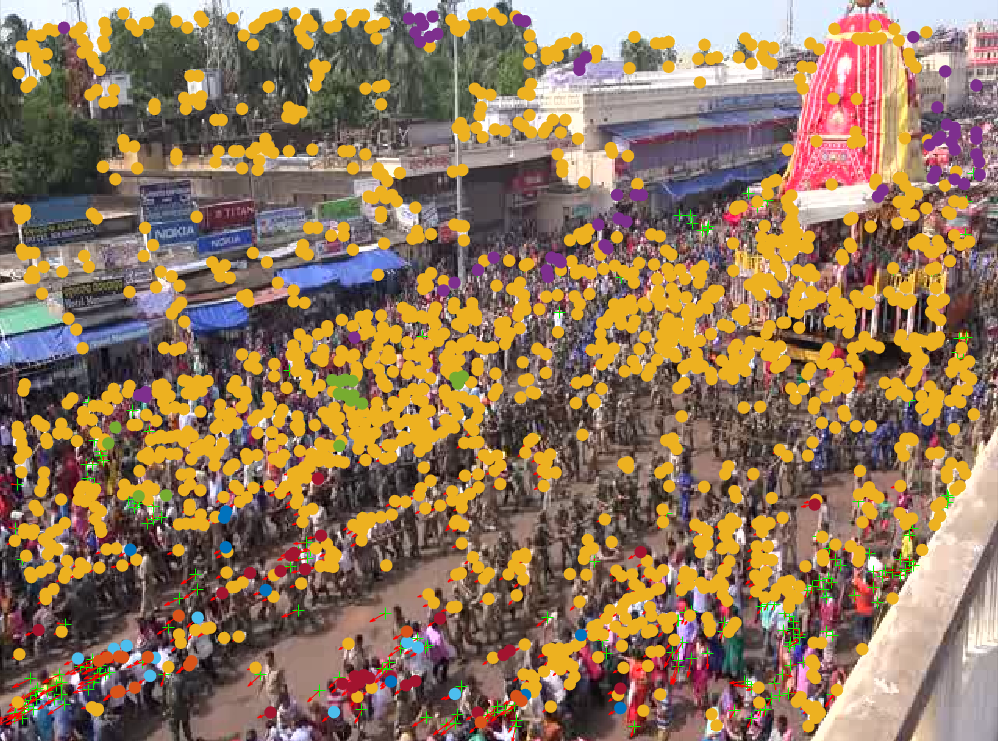}
        \caption{19}
        \label{fig:SantoroRY33}
        \end{subfigure}
               \begin{subfigure}[t]{0.20\textwidth}
        \includegraphics[scale=0.1,width=0.8\textwidth]{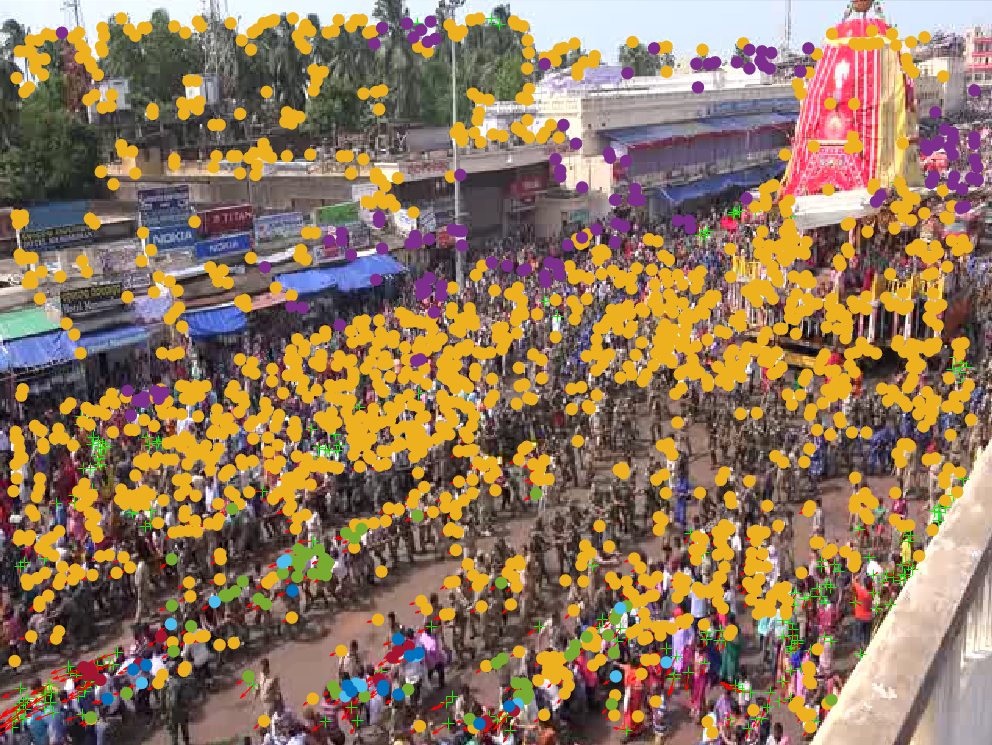}
        \caption{20}
        \label{fig:SantoroRY34}
        \end{subfigure}
   \begin{subfigure}[t]{0.20\textwidth}
        \includegraphics[scale=0.1,width=0.8\textwidth]{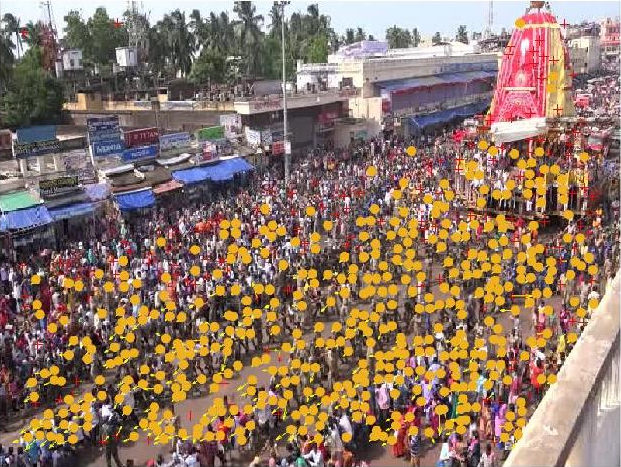}
        \caption{21}
        \label{fig:CCRY31}
        \end{subfigure}
               \begin{subfigure}[t]{0.20\textwidth}
        \includegraphics[scale=0.1,width=0.8\textwidth]{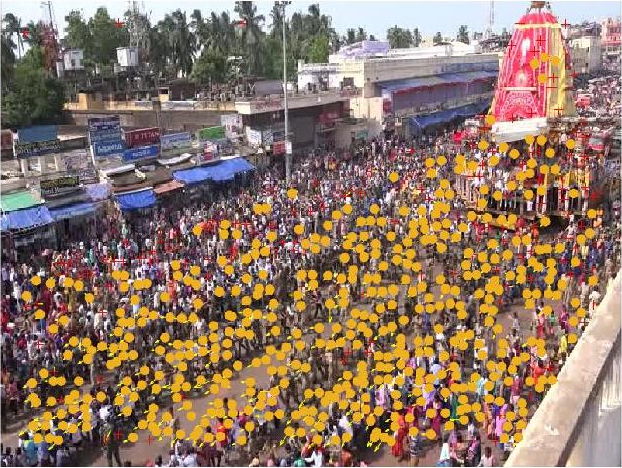}
        \caption{22}
        \label{fig:CCRY32}
        \end{subfigure}
               \begin{subfigure}[t]{0.20\textwidth}
        \includegraphics[scale=0.1,width=0.8\textwidth]{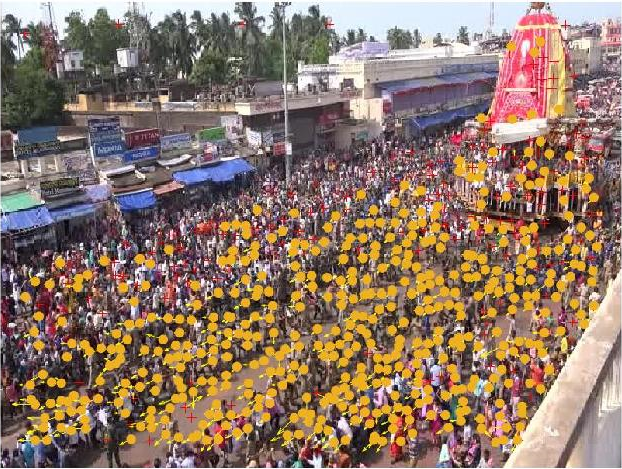}
        \caption{23}
        \label{fig:CCRY33}
        \end{subfigure}
               \begin{subfigure}[t]{0.20\textwidth}
        \includegraphics[scale=0.1,width=0.8\textwidth]{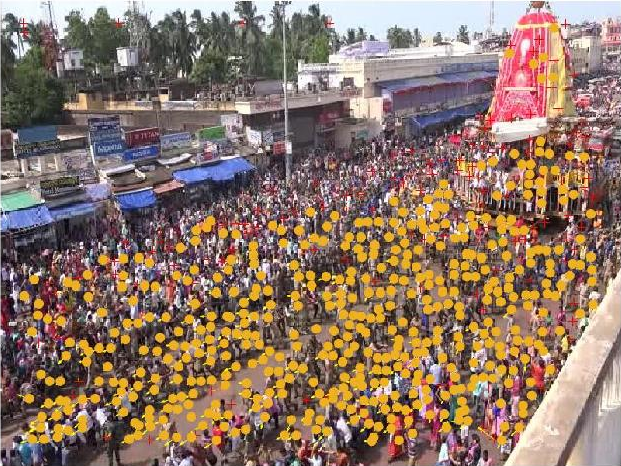}
        \caption{24}
        \label{fig:CCRY34}
        \end{subfigure}
           \begin{subfigure}[t]{0.20\textwidth}
        \includegraphics[scale=0.1,width=0.8\textwidth]{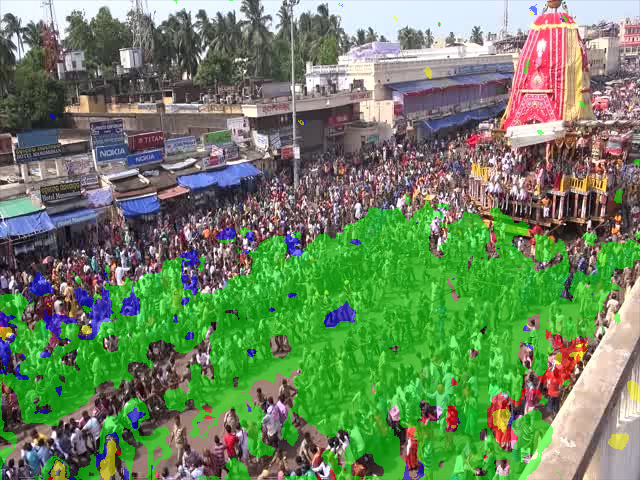}
        \caption{25}
        \label{fig:DIHMRY31}
        \end{subfigure}
               \begin{subfigure}[t]{0.20\textwidth}
        \includegraphics[scale=0.1,width=0.8\textwidth]{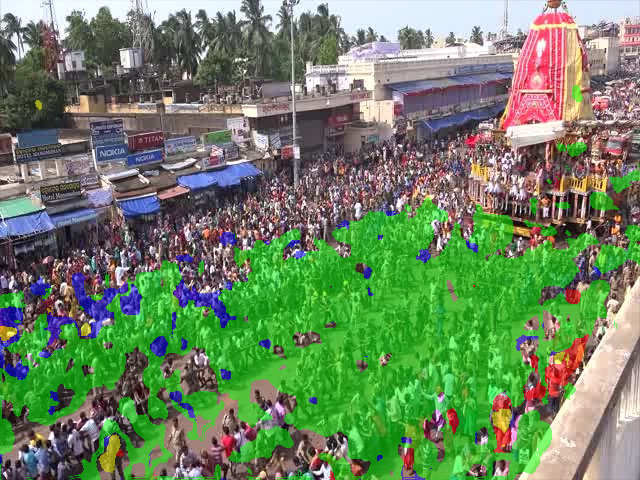}
        \caption{26}
        \label{fig:DIHMRY32}
        \end{subfigure}
               \begin{subfigure}[t]{0.20\textwidth}
        \includegraphics[scale=0.1,width=0.8\textwidth]{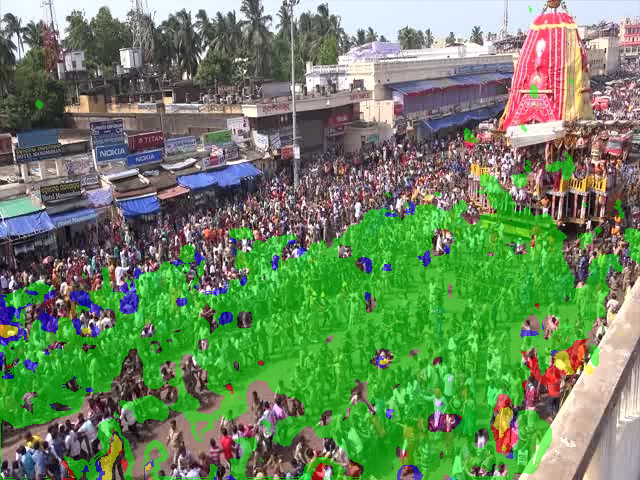}
        \caption{27}
        \label{fig:DIHMRY33}
        \end{subfigure}
               \begin{subfigure}[t]{0.20\textwidth}
        \includegraphics[scale=0.1,width=0.8\textwidth]{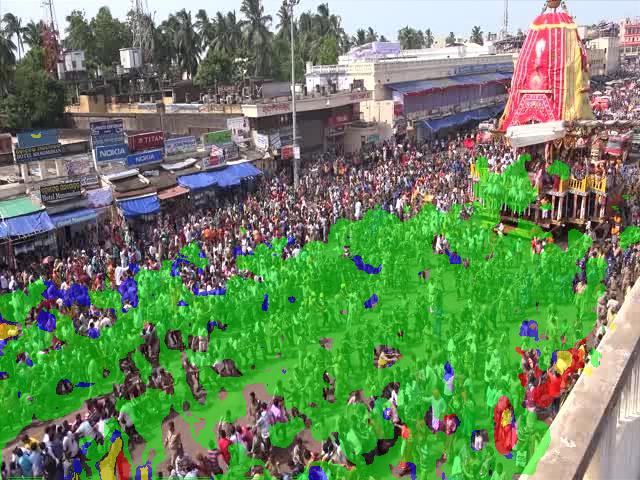}
        \caption{28}
        \label{fig:DIHMRY34}
        \end{subfigure}
        
    \caption{($1$-$4$) Original recorded Frames~(31-34) of the Rath Yatra video, ($5$-$8$) Ground Truth Frames, ($9$-$12$) represent segmented outputs obtained using~proposed method, ($13$-$16$) represent outputs of segmentation method~\citep{ali2007lagrangian}, ($17$-$20$) represent outputs of segmentation method~\citep{santoro2010crowd}, ($21$-$24$) represent outputs of segmentation method~\citep{zhou2014CC} and ($25$-$28$) represent outputs of segmentation using~\citep{ullah2017density}, respectively. (Best viewed in color)}
\label{fig:RY}
\end{figure}

\par Marathon-I video has a unidirectional linear motion flow. Even though the crowd is sparse, the proposed method is able to segment the unidirectional flow. The segmented flow using the proposed method is depicted in Fig.\ref{fig:688}.  
The accuracy plot is shown in Fig.\ref{fig:AccPlot688}. The plot consists of peaks at regular intervals which indicate the initialization of the window $W$ where accuracy is maximum. 
The average accuracy for Marathon-I has been found to be $89\%$, which is better than the methods proposed in \citep{ali2007lagrangian}, \citep{santoro2010crowd}, \citep{zhou2014CC}, and \citep{ullah2017density}. Marathon-II is a dense crowd video where people are running in one direction. This video can be considered as a perfect test video where the flow can be observed from the beginning. The images in Fig.\ref{fig:692} show how the proposed method is able to segment this increasing flow with an average accuracy of $97\%$.
 %accuracy plots
\begin{figure*}[t]     %!h
%\scriptsize
   \centering
  %  \vspace{-20pt}
\begin{subfigure}[b]{0.4\textwidth}
\includegraphics[width=6.0 cm,height=4.0 cm]{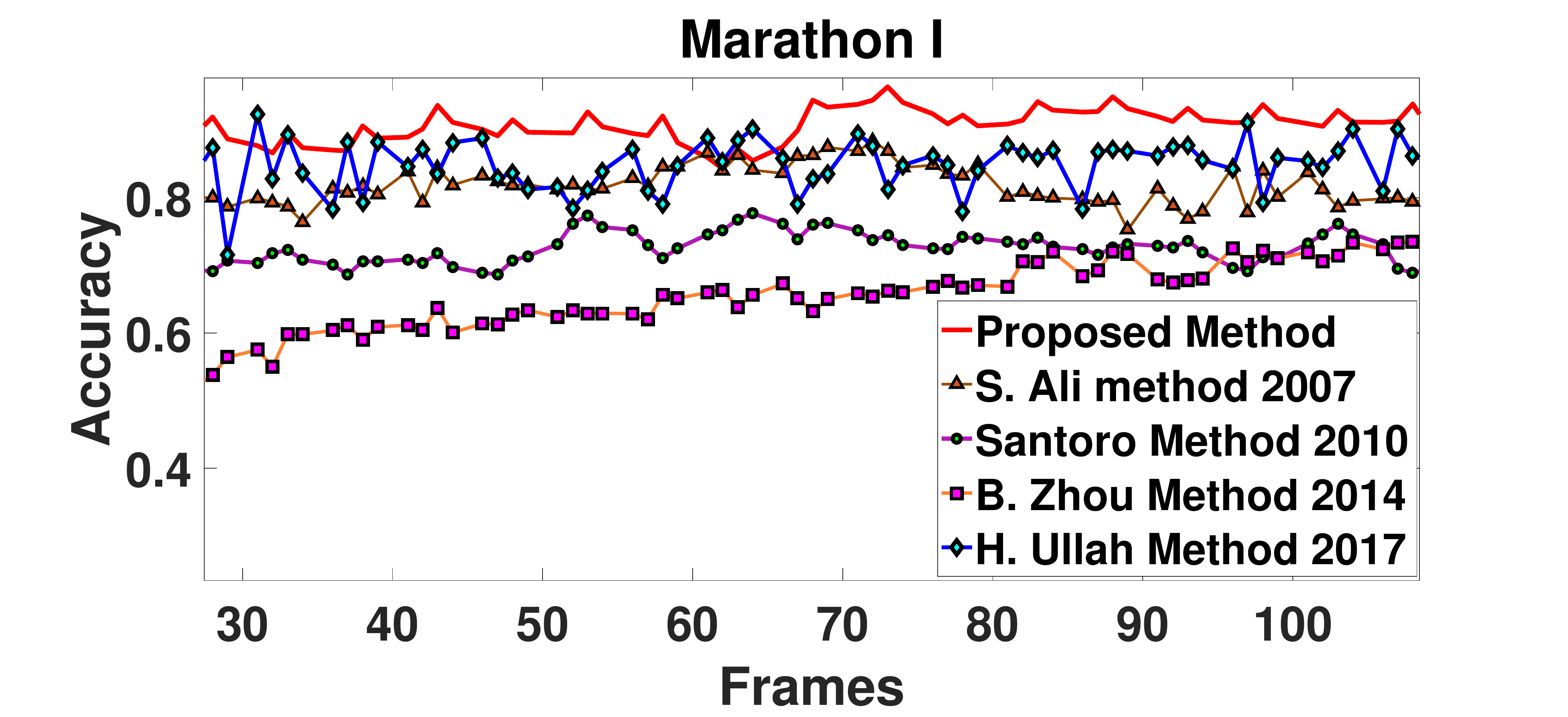} %accuracyComparision_FiveMethods_MarathonI
        \caption{}
        \label{fig:AccPlot688}
    \end{subfigure}
 %   \hspace{1 cm}
 \hspace{0.8cm}
       \begin{subfigure}[b]{0.4\textwidth}
\includegraphics[width=6.0 cm,height=4.0 cm]{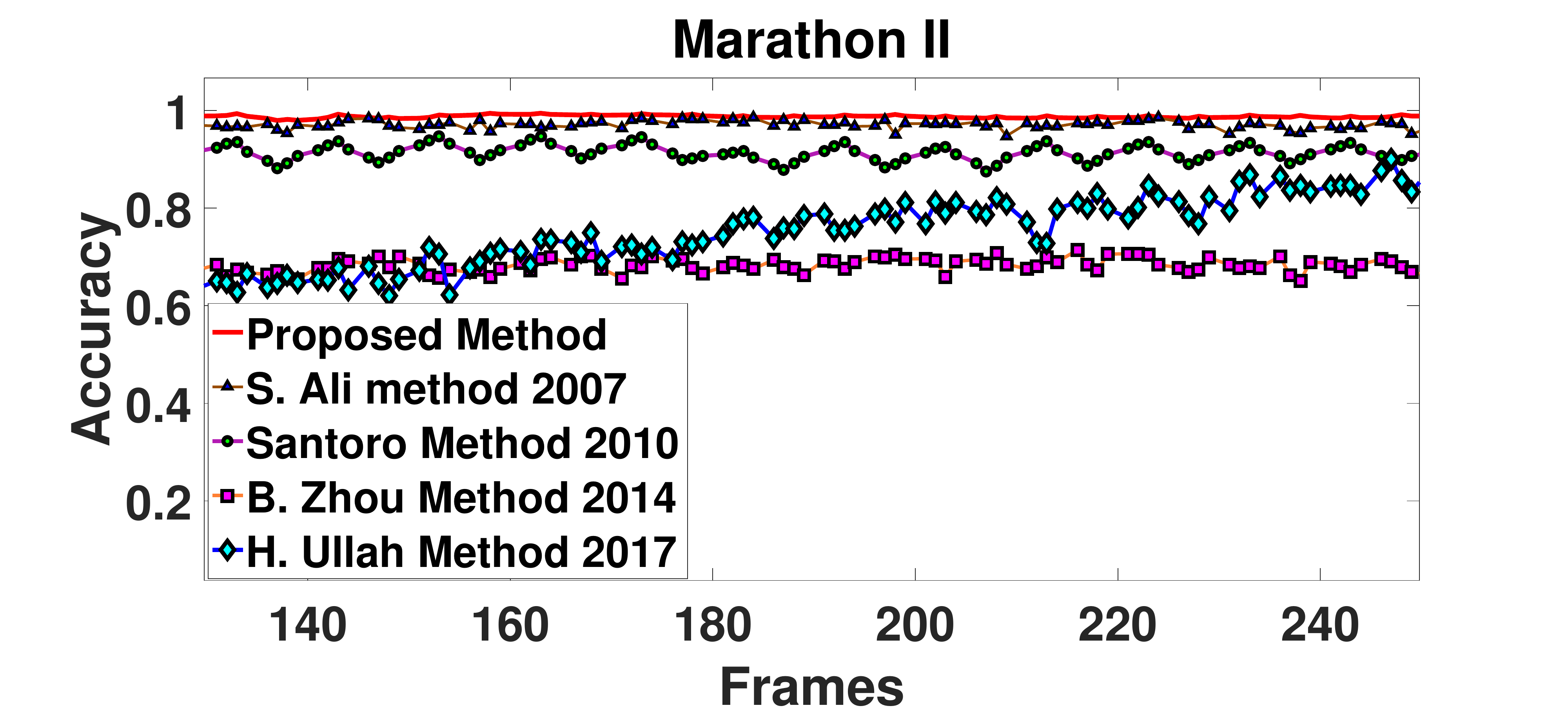} % accuracyComparision_FiveMethods_Fair
        \caption{}
        \label{fig:AccPlot692}
    \end{subfigure}
    \hspace{0.8cm}    %[scale=0.60,width=0.75\textwidth]
       \begin{subfigure}[b]{0.4\textwidth}
        \includegraphics[width=6.0 cm,height=4.0 cm]{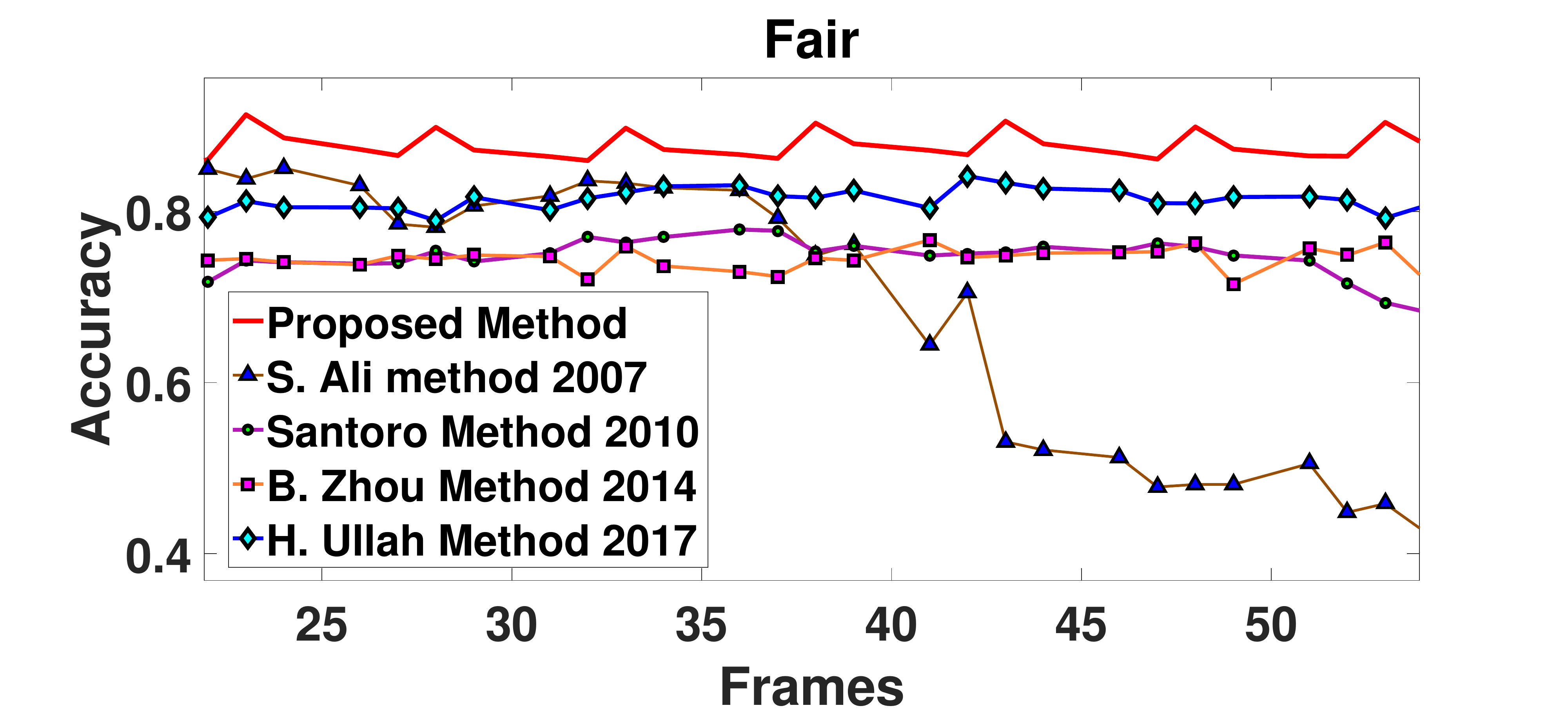}  %accuracyComparision_FiveMethods_Fair
        \caption{}
        \label{fig:AccPlotFair}
        \end{subfigure}
  %       \hspace{1 cm}
   \hspace{0.8cm}
       \begin{subfigure}[b]{0.4\textwidth}
        \includegraphics[width=6.0 cm,height=4.0 cm]{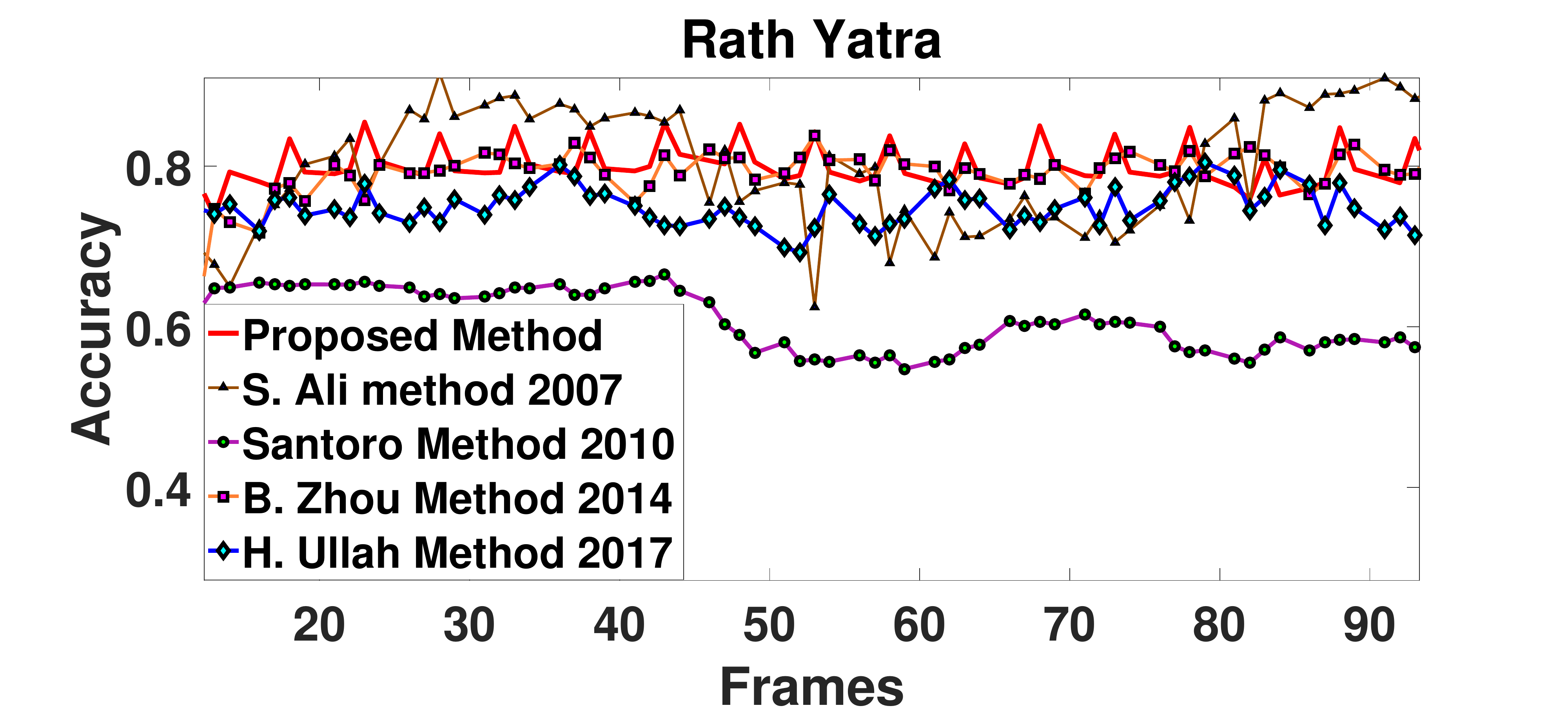} %accuracyComparision_FiveMethods_RY
        \caption{}
        \label{fig:AccPlotRY}
        \end{subfigure}
    \label{fig:accPlots}
    \caption{(a-d) Frame-wise accuracy plot of various videos for the proposed method, \citep{ali2007lagrangian}, \citep{santoro2010crowd}, \citep{zhou2014CC}, and \citep{ullah2017density}, respectively.(Best viewed in color)}
\end{figure*}
%accuracy plots window time execution
\begin{figure*}[htp]     %!h
 \centering
 %\footnotesize
%   \vspace{-40pt}
\begin{subfigure}[t]{0.4\textwidth}
\includegraphics[width=5.0 cm,height=4.0 cm]{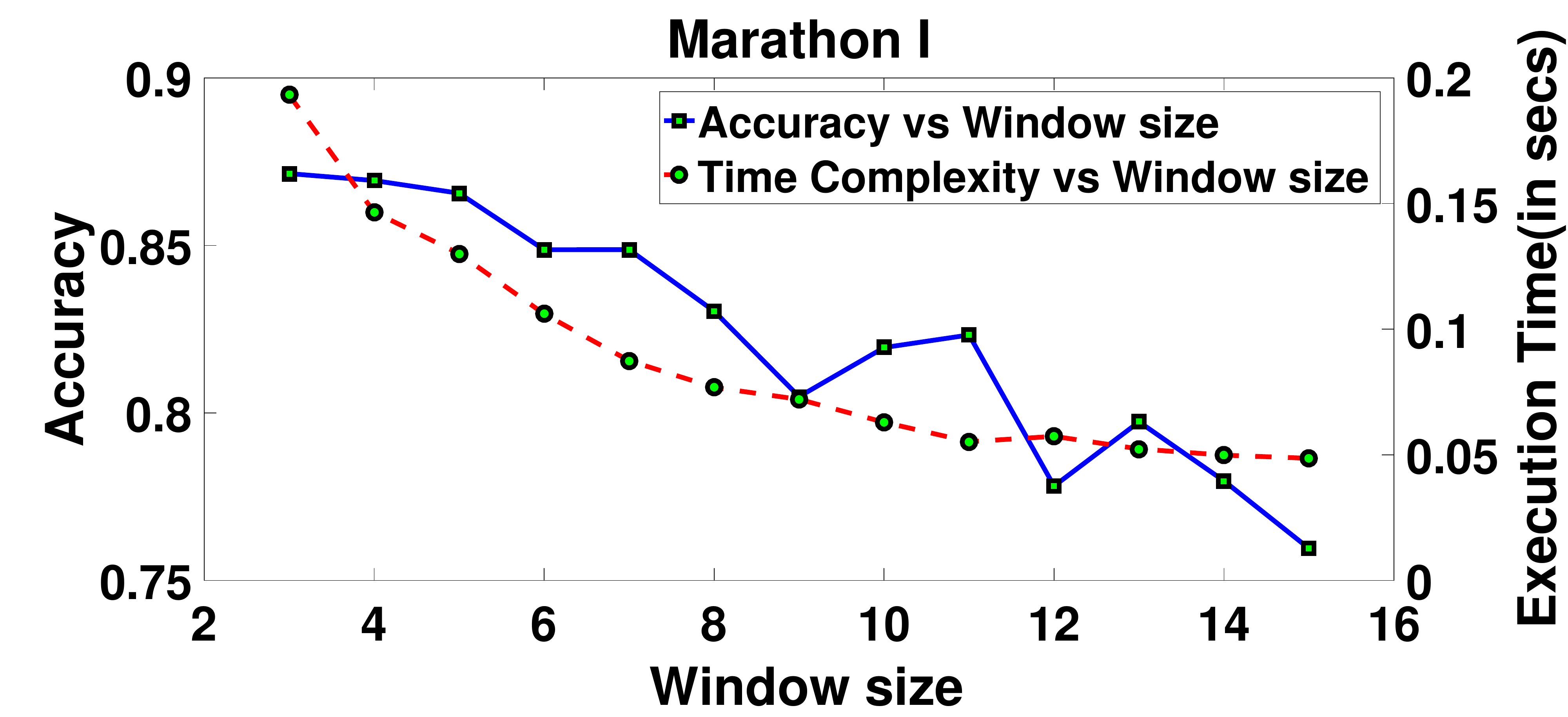}
        \caption{}
        \label{fig:AccWPlot688}
    \end{subfigure}
    \hspace{1.1 cm}
 %\hspace{0.6cm}
       \begin{subfigure}[t]{0.4\textwidth}
\includegraphics[width=5.0 cm,height=4.0 cm]{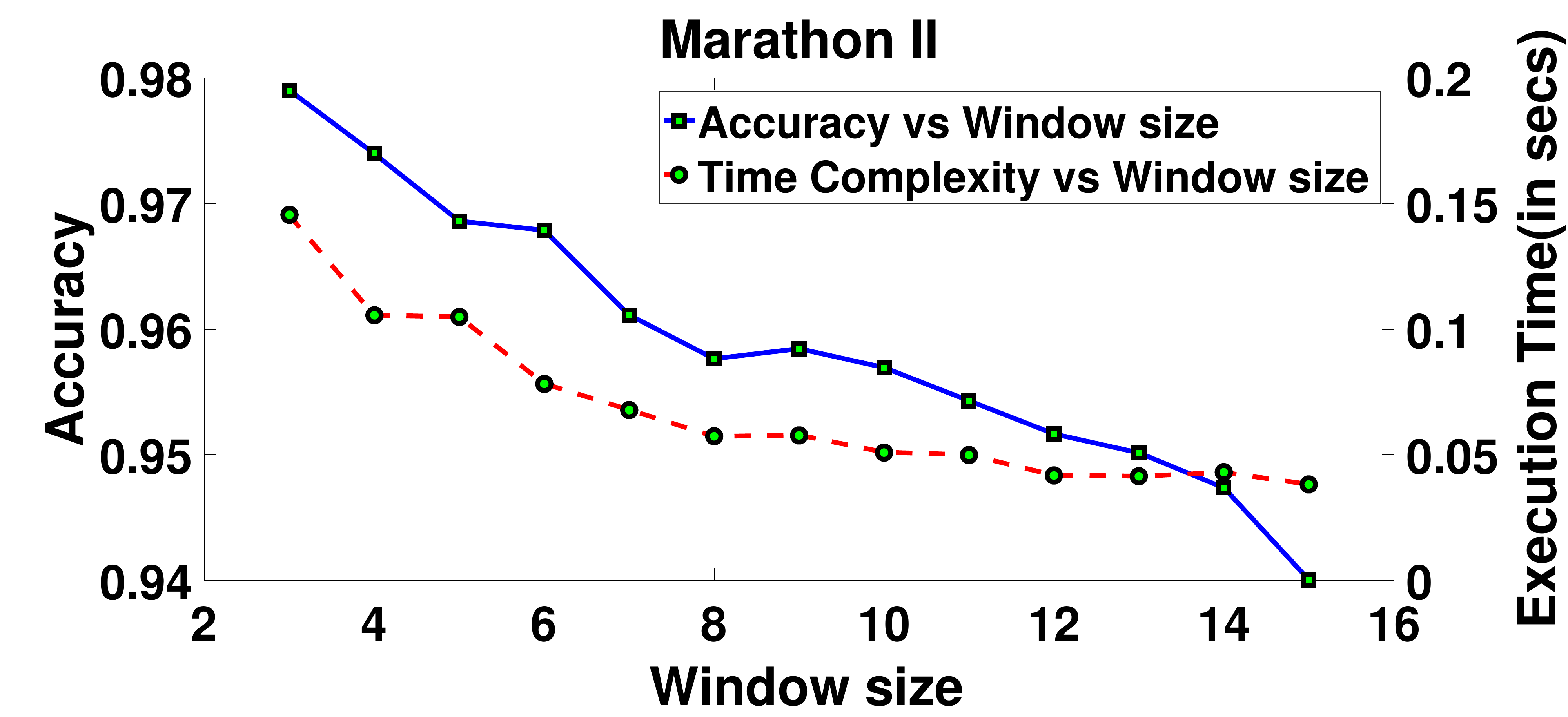}
        \caption{}
        \label{fig:AccWPlot692}
    \end{subfigure}
    %\hspace{0.6cm}
       \begin{subfigure}[t]{0.4\textwidth}
        \includegraphics[width=5.0 cm,height=4.0 cm]{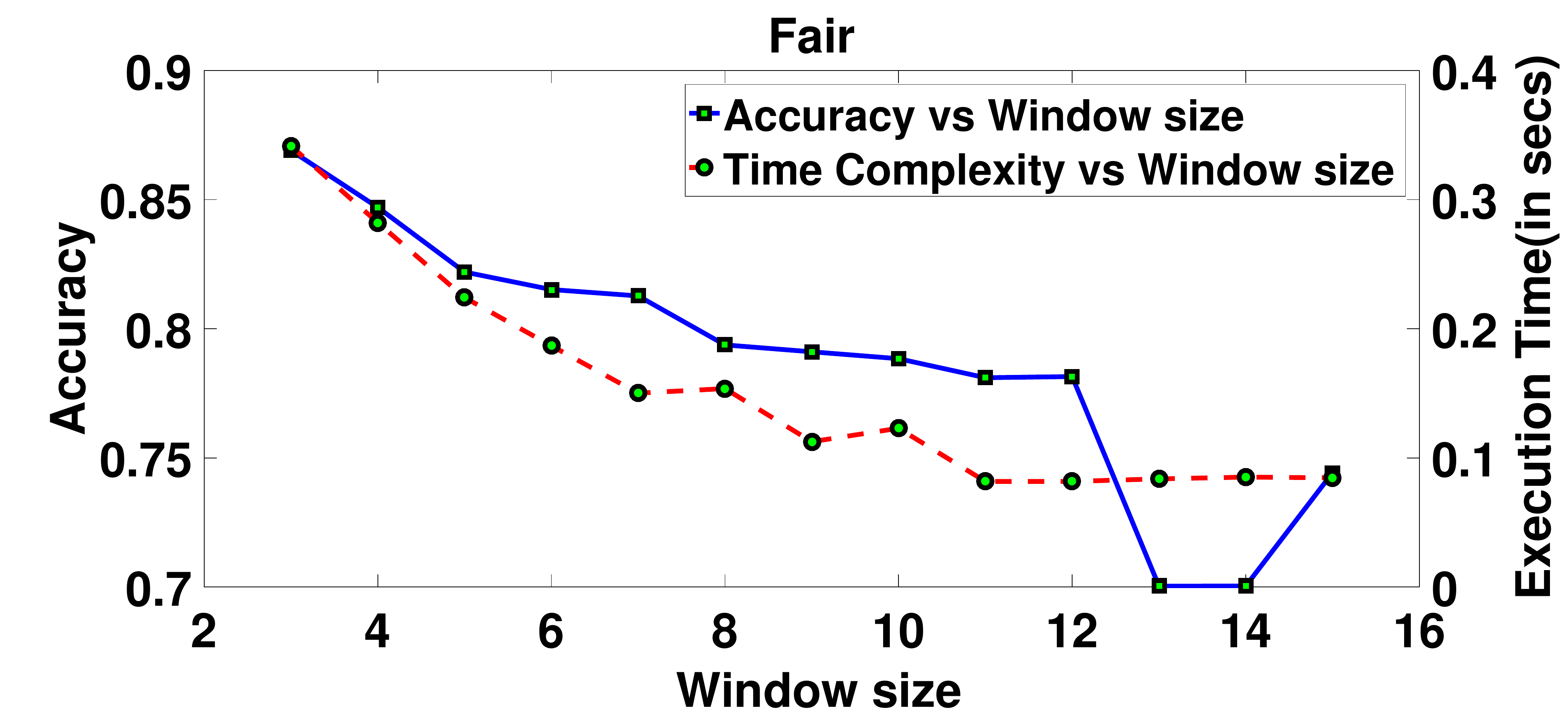}
        \caption{}
        \label{fig:AccWPlotFair}
        \end{subfigure}
        \hspace{1.1 cm}
 %\hspace{0.6cm}
       \begin{subfigure}[t]{0.4\textwidth}
        \includegraphics[width=5.0 cm,height=4.0 cm]{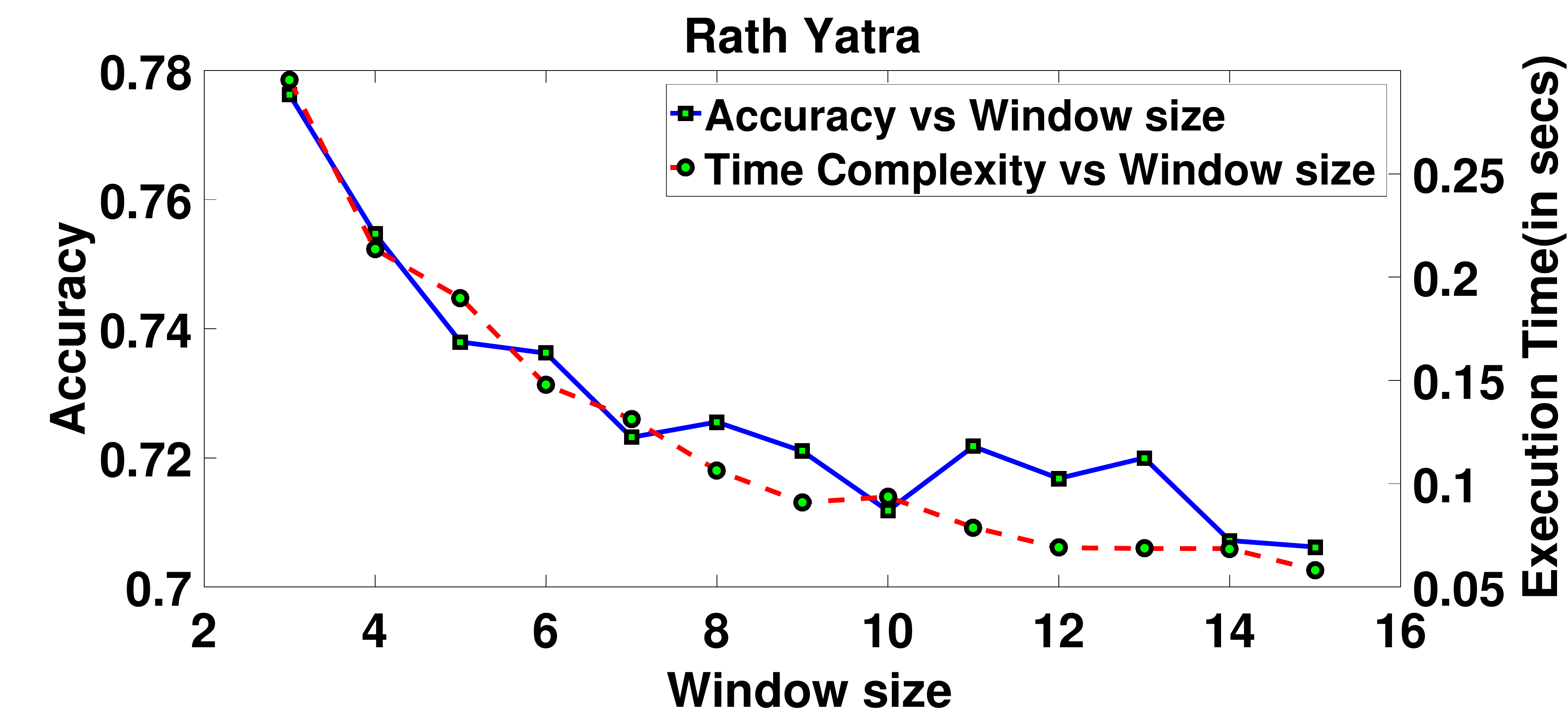}
        \caption{}
        \label{fig:AccWPlotRY}
        \end{subfigure}

    \caption{(a-d) Accuracy-Execution Time-Window plot for various videos. The red graphs indicate the execution time per frame using the proposed method. The blue graphs indicate the accuracy plot for the proposed method. (Best viewed in color)}
        \label{fig:accW}
\end{figure*}

 \par Fair video is a semi-dense sequence where people are moving in opposite directions. Our proposed method is able to handle this challenging situation and has segmented the bi-directional structured flows with an average accuracy of  $86\%$, which is better than \citep{ali2007lagrangian}, \citep{santoro2010crowd}, \citep{zhou2014CC} and \citep{ullah2017density}, respectively. The segmentation outputs and accuracy plots for this video are shown in Fig.\ref{fig:Fair} and Fig.\ref{fig:AccPlotFair}, respectively.

 \par Rath Yatra video is a sequence where people can be seen pulling the Cart (Rath) in one direction. The sequence consists of both  structured as well as unstructured flows. Pulling of the Rath is a structured flow, while the people moving around this structured flow in different directions can be considered as random. The proposed method is able to segment this structured flow with an average accuracy of $80\%$. Though the average accuracy is marginally lower than \citep{ali2007lagrangian}, however, the difference is not significant as can be seen in the Table \ref{table:Accuracy analysis}. This effect is because of more randomness in the crowd. The segmented maps and the accuracy plots are shown in Fig.\ref{fig:RY} and Fig.\ref{fig:AccPlotRY}, respectively.

\begin{table}[h]
\tiny
\centering
\caption{Comparison of the proposed method with state-of-the-art  in terms of accuracy.\label{table:Accuracy analysis}}
%\scriptsize
 
%\resizebox{\textwidth}{!}{%
%\begin{tabular}{|l|c|c|c|l|l|}
%\begin{tabular}{lccccc}
\begin{tabular}{lp{1.5cm}p{1.5cm}p{1.5cm}p{1.5cm}p{1.5cm}}
\hline
\multirow{2}{*}{\#Dataset} & \multicolumn{5}{c}{Average Accuracy (\%)} \\ \cline{2-6} 
 & Proposed Method & {}\citep{ali2007lagrangian}{}& {}\citep{santoro2010crowd}{}& {}\citep{zhou2014CC}{} & {}\citep{ullah2017density}{} \\  \hline
Marathon-I  & 89.68 & 80.46 & 71.75  & 63.41 & 84.98  \\  %\hline
Marathon-II & 97.53 & 95.14 & 89.63  & 66.74 & 80.36\\  %\hline
Fair        & 86.71 & 70.35 & 72.32  & 75.34 & 81.09 \\  %\hline
Rath Yatra  & 80.35 & 81.03 & 60.61  & 79.22 & 75.19  \\ \hline
\end{tabular}%

%}
%\label{table:Accuracy analysis} 
\end{table}
\subsection{Computational Performance} 
\label{section:CP}
We now present the computational overhead of the proposed method. The experiments have been conducted on a desktop computer powered by quad-core processor with $8$ GB of memory. 

The execution time of the proposed method has been compared with the execution time of popular existing state-of-the-art methods. It can be observed from Table \ref{table:Time analysis} that the proposed method is much faster than other methods. This is because the proposed method calculates optical flow at the start of the window and estimates the flow in the remaining frames of the window. As a result, a good amount of computation time is saved.
In another experiment related to execution time, it has been shown how the accuracy and execution time vary over varying window size. It has been shown in Fig.\ref{fig:accW} that as the window size increases, the accuracy also reduces. Therefore, selection of a reasonable window size is important. From all the graphs in Fig.\ref{fig:accW}, it may be observed that the accuracy is higher and the execution time is considerably lower when the window size is in between $4$ to $6$.  
\begin{table}[h]
\tiny
\centering
%\scriptsize
\caption{Comparison of the proposed method with popular existing methods in terms of execution time per frame
\label{table:Time analysis} }
%\resizebox{\textwidth}{!}{%
%\begin{tabular}{|l|c|c|c|l|l|}
%\begin{tabular}{lccccc}
\begin{tabular}{l p{1.5cm}p{1.5cm}p{1.5cm}p{1.5cm}p{1.5cm}}
\hline
\multirow{2}{*}{\#Dataset} & \multicolumn{5}{c}{Time taken per frame (in seconds) } \\ \cline{2-6} 
 & Proposed Method & {}\citep{ali2007lagrangian}{}& {}\citep{santoro2010crowd}{}& {}\citep{zhou2014CC}{} & {}\citep{ullah2017density}{} \\ \hline
Marathon-I  & 0.103 & 8.657 & 1.259  & 0.562 & 0.548  \\ %\hline
Marathon-II & 0.093 & 9.742 & 2.240  & 0.695 & 0.594\\ %\hline
Fair        & 0.159 & 12.229 & 1.811  & 0.756 & 0.862 \\ %\hline
Rath Yatra  & 0.139 & 10.682 & 1.834  & 0.702 & 0.744  \\ \hline
\end{tabular}%
%}
%\label{table:Accuracy analysis} 
\end{table}
\section{Conclusion}
\label{sections:Conclusions}
In this paper, crowd flow segmentation using Langevin equation has been proposed. The method is able to segment the linear flows successfully without the need of estimating the optical flow in every  frame. The solutions to the Langevin equations described are able to predict the velocity and position of the key points with noticeable accuracy. Computation time for dominant flow estimation can be substantially reduced using the proposed method. The proposed model can be extended to segment non-linear motion flows. The information obtained from the dominant flows can be used to train machine learning models. These trained models can be used for flow classification and prediction which are important part of intelligent crowd surveillance systems.

\section*{Acknowledgement}
\label{sections:ACK}
This research work is funded by Science and Engineering Research Board (SERB), Department of Science and Technology, Government of India through the grant YSS/2014/000046.

%\bibliographystyle{plainnat}
%\bibliography{bibref}

\end{document}